\documentclass[journal]{IEEEtran}
\usepackage[utf8]{inputenc}
\usepackage{amssymb}
\usepackage{graphicx}
\usepackage{booktabs}
\usepackage{multirow}
\usepackage{cite}
\usepackage{subfigure}
\usepackage{enumerate}
\usepackage{threeparttable}
\usepackage[switch]{lineno} 
\usepackage{lipsum} 
\usepackage{colortbl}
\usepackage{hhline}
\usepackage[fleqn]{amsmath}
\usepackage[fleqn]{mathtools}
\usepackage{bm}
\usepackage{makecell}
\usepackage{xcolor}
\usepackage{algorithm}
\usepackage{algorithmic}

\newcommand{\cb}[1]{{\color{blue} #1}}

\ifCLASSOPTIONcompsoc
  \usepackage[caption=false,font=normalsize,labelfont=sf,textfont=sf]{subfig}
\else
  \usepackage[caption=false,font=footnotesize]{subfig}
\fi
\begin{document}

\title{Few-Shot Hyperspectral Image Classification With Unknown Classes Using Multitask Deep Learning}

\author{Shengjie~Liu,~\IEEEmembership{Member,~IEEE,}
        Qian~Shi,~\IEEEmembership{Senior Member,~IEEE,}
        and~Liangpei~Zhang,~\IEEEmembership{Fellow,~IEEE}
\thanks{\cb{This is the preprint version. To read the final version please go to https://doi.org/10.1109/TGRS.2020.3018879}}
\thanks{Manuscript received May 11, 2020; accepted August 18, 2020. This work was supported in part by the National Natural Science Foundation of China under Grant 61976234, and in part by Guangdong Basic and Applied Basic Research Foundation under Grant 2019A1515011057. \textit{(Corresponding author: Qian Shi.)}} 
\thanks{Shengjie Liu is with the Department of Physics, The University of Hong Kong, Pokfulam, Hong Kong (e-mail: sjliu.me@gmail.com).}
\thanks{Qian Shi is with the Guangdong Provincial Key Laboratory of Urbanization and Geo-simulation, School of Geography and Planning, Sun Yat-sen University, Guangzhou 510275, China (e-mail: shixi5@mail.sysu.edu.cn).}
\thanks{Liangpei Zhang is with the State Key Laboratory of Information Engineering in
Surveying, Mapping and Remote Sensing and the Collaborative Innovation
Center of Geospatial Technology, Wuhan University, Wuhan 430079, China  (e-mail: zlp62@whu.edu.cn).}
}

\markboth{Accepted by IEEE TGRS}%
{Accepted by IEEE TGRS}

\maketitle

\begin{abstract}
Current hyperspectral image classification assumes that a predefined classification system is closed and complete, and there are no unknown or novel classes in the unseen data. However, this assumption may be too strict for the real world. 
Often, novel classes are overlooked when the classification
system is constructed. The closed nature forces a model to assign a label given a new sample and may lead to overestimation of known
land covers (e.g., crop area).
    To tackle this issue,
    we propose a multitask deep learning  method that simultaneously conducts classification and reconstruction in the open world (named MDL4OW) where unknown classes may exist. The reconstructed data are compared with the original data; those failing to be reconstructed are considered unknown, based on the assumption that they are not well represented in the latent features due to the lack of labels. 
    A threshold needs to be defined to separate the unknown and known classes; we propose two strategies based on the extreme value theory for few-shot and many-shot scenarios. 
    The proposed method was tested on  real-world hyperspectral images;  state-of-the-art results were achieved, e.g., improving the overall accuracy by 4.94\% for the Salinas data. By considering the existence of unknown classes in the open world, our method achieved more accurate hyperspectral image classification, especially under the few-shot context.
\end{abstract}

\begin{IEEEkeywords}
multitask learning, deep learning, convolutional neural network, hyperspectral image classification, open-set recognition, classification with unknown classes
\end{IEEEkeywords}

\IEEEpeerreviewmaketitle

\section{Introduction}
\label{sec:intro}

The world's urbanization has rapidly developed since the 1980s. With the growing population, the large number of human beings living on  this small planet becomes a crucial problem. Sufficient food supplies are needed to feed human beings, but sufficient land resources for daily life activities are also needed~\cite{gebbers2010precision}. Ongoing urbanization transforms natural environments into cities and increases the fraction of the impervious surface, eliminating rainwater infiltration and increasing the urban temperature. 
Planning land resource usage wisely is the key to maintaining the world's sustainability, and this requires timely and accurate land cover monitoring.  

Existing methods for land cover mapping include land survey and classification based on satellite and airborne remote sensing images. A land survey is impractical for large area monitoring as it is labor-intensive and time-consuming. Classifying each pixel of a satellite image is a more popular way. Two existing semi-automatic methods can be applied, i.e., semantic segmentation and pixel-wise classification. Semantic segmentation considers the entire image as a sample instance and directly outputs the segmentation result~\cite{audebert2016semantic}; it requires a large set of fully annotated images~\cite{kampffmeyer2016semantic}. This approach is not practical for detailed land cover mapping, as a) the land cover classification system is very complex, and b) the satellite data are quite diverse. Specifically, depending on the application, a remote sensing image can be either mapped into 3-4 classes (impervious, vegetation, water, and sometimes barren) or 10+ fine-grained classes  \cite{ridd1995exploring, phinn2002monitoring, yuan2005multi}. The land cover classification system is multilevel. 
    On the other hand, the sources of remote sensing images are diverse, making it difficult to reuse previous datasets. For optical remote sensing, the number of spectral bands ranges from one to hundreds. The diversity increases if non-optical data (e.g., PolSAR) are included. As a result, pixel-wise classification is more popular for land cover mapping.

    Remotely sensed hyperspectral images, due to their rich spectral information, are of great use for land cover classification. 
    The physical theory for hyperspectral image classification is that each class can be separated based on its spectral profiles. Due to its great potential,
    hyperspectral  image classification has gone through rapid development in the last two decades \cite{fauvel2012advances, bioucas2013hyperspectral, audebert2019deep}. 
    In the early 2000s, hyperspectral images were solely considered as sequential data, as shown in the classic 2-D storage format ({\#}sample$\times${\#}feature).  At this stage, support vector machines (SVMs) were the mainstream classifiers \cite{gualtieri1999support, demir2007hyperspectral, ghamisi2017advanced}. SVMs with kernels can project hyperspectral data to a high-dimensional hyperplane, where the training data should be highly separated based on their class belongings. 
    In the late 2000s, with the popularity of graph models (e.g., Markov random fields \cite{li2011spectral}), morphological profiles \cite{benediktsson2005classification}, and other spatial filters, including edge-preserving filters and Gabor filters \cite{kang2013spectral,shen2011three}, spectral-spatial classification of hyperspectral data became a popular topic  and influenced the last ten years  \cite{fauvel2012advances}. 
    At the same time, many advanced algorithms tackling the limited availability of training samples  were proposed, including active learning \cite{rajan2008active, shi2015spatial}, semi-supervised learning \cite{shi2013semisupervised, wang2017novel}, graph models \cite{ghamisi2013spectral}, sparse representation \cite{zhang2013nonlocal, wang2018locality}, and domain adaptation \cite{shi2015domain}.  
    Since the classification performance also depends on the dimension due to the curse of dimensionality, band selection  and dimension reduction techniques are widely used in hyperspectral image classification \cite{wang2016salient, yu2018class, luo2020dimensionality}.
    
    In the deep learning era, the novel convolutional neural networks (CNNs), recurrent neural networks and graph convolutional networks  started to dominate the classification of hyperspectral and remote sensing data \cite{lu2017remote, lee2017going,  mou2017deep, paoletti2018new,  hang2019cascaded, audebert2019deep,  wan2019multiscale, liu2020local}. 
    Manual feature engineering was replaced by automatic feature learning by deep networks. Now, hyperspectral image classification entered the stage with 99\% classification accuracy with sufficient training samples \cite{zhong2017spectral}.  Besides, many works have focused on improving the few-shot classification accuracy (classification with limited samples), including deep few-shot learning (DFSL) with neural networks and SVM\cite{liu2018deep},  spatial-spectral relation network (SS-RN) \cite{rao2019spatial},  and deep relation network \cite{gao2020deep}.
    However, the high accuracy is achieved under the assumption that there is no unknown class in the unseen data. If an instance does not belong to any of the known classes, this instance will not harm the classification accuracy from the current evaluation system, as such instance will not be considered. But the existence of unknown or novel classes will lead to false positives and thereby reduce the precision of a model. If many instances belong to the unknown class are classified as a known class, the number of the known classes will be  largely overestimated.      
    Unfortunately, in the real world, it is difficult, if not impossible, to define all potential classes. As mentioned, a sample collection based on a field survey can only cover a small portion of the study area due to budget limits. For example, \cite{zhou2019dcn} collected ground truth data along the major road network, and those buried far from the road network were inevitably neglected. Therefore, a machine learning model trained on the collected samples might face some instances (the unknown) that cannot be classified into one of the known classes.

\begin{figure}[!t]
\centering
\subfigure[Image]{\includegraphics[width=0.12\textwidth]{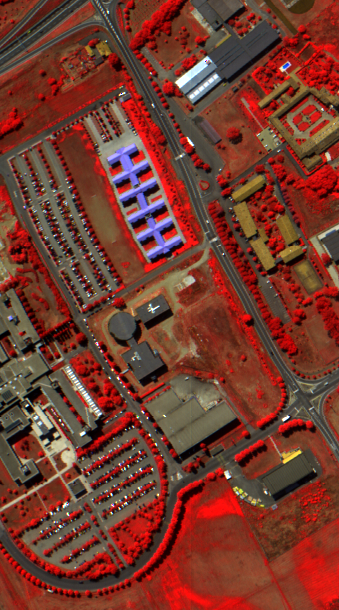}} 
\subfigure[Classification]{\includegraphics[width=0.12\textwidth]{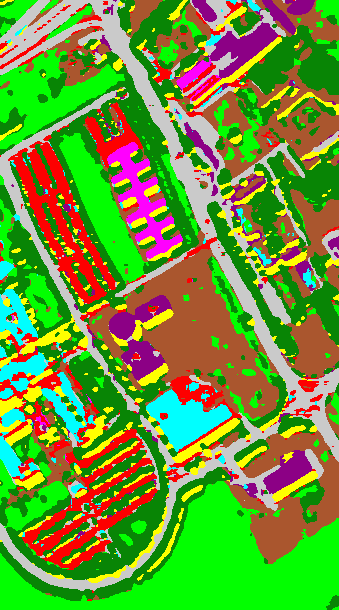}}
\subfigure[Ground truth]{\includegraphics[width=0.12\textwidth]{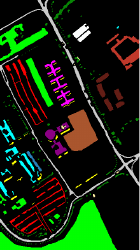}} 
\subfigure{\includegraphics[width=0.08\textwidth]{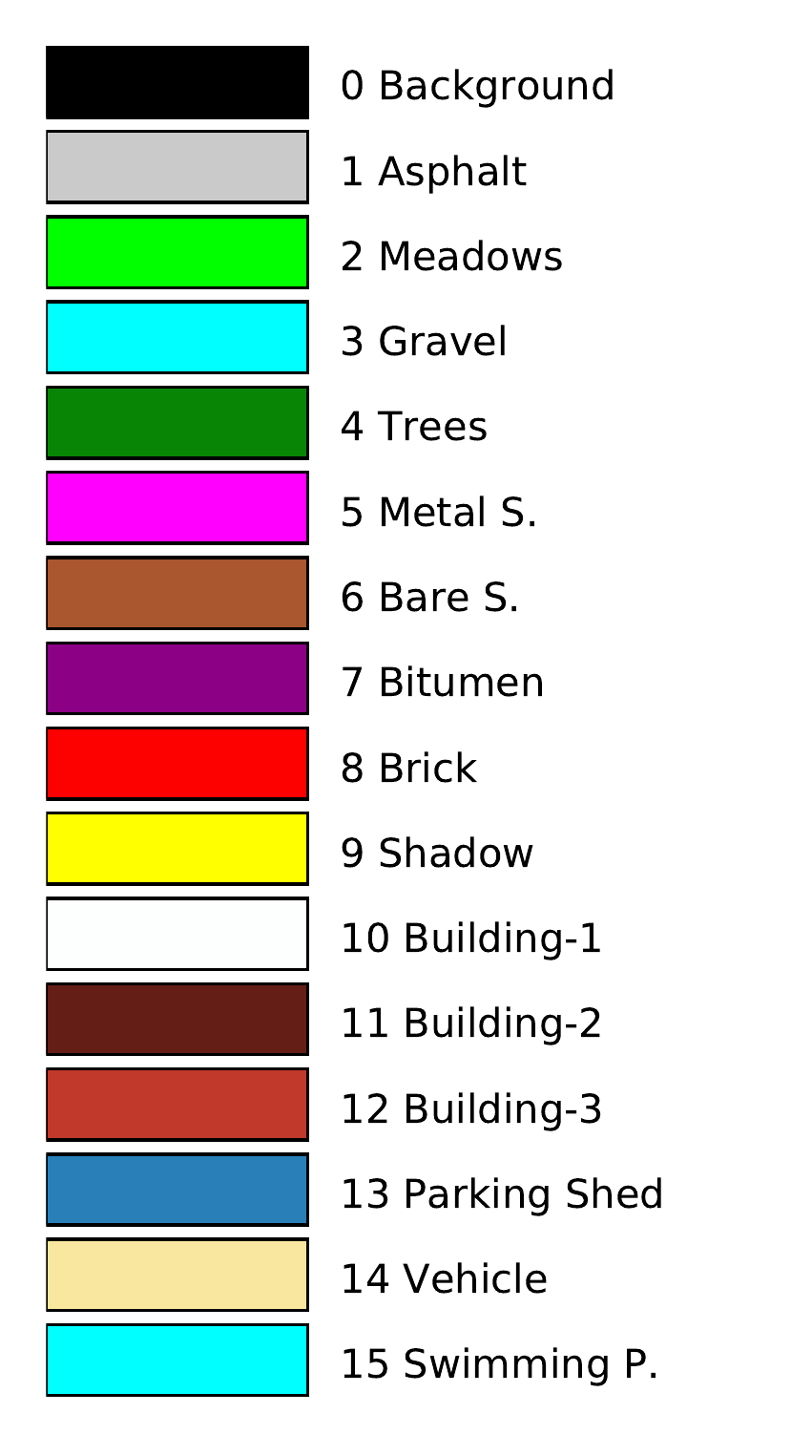}}
\caption{Example of the University of Pavia's classification map (nine classes, OA=98\%). The original ground truth covers only nine classes. We manually annotated six additional classes, which were unknown to the classifier. Note building-1, building-2 and building-3 on the right are misclassified as a mix of bare soil and other materials. The swimming pool, helicopters (vehicle) and the parking shed are all misclassified as natural land covers.}
\label{fig:paviaExample}
\end{figure} 
    
    The influence of unknown classes, although dependent on the area of an image, may even occur in a very small study region. 
The University of Pavia dataset, one of the most popular hyperspectral benchmarks, records nine land cover classes in a region of 610$\times$340 with 1.3 \textit{m} ground sampling distance (GSD), approximately 793$\times$442 $\textit{m}^2$. This dataset is small and barely covers a campus. 
    However, there is a swimming pool that should not belong to any of the nine classes. We show the false color map of the University of Pavia data in Figure \ref{fig:paviaExample}, where the swimming pool is located on the upper right side of the image. 
    In addition, on the right side of the image, three groups of buildings (building-1, building-2, building-3) are not annotated in the original reference map. 
    These buildings are often misclassified as natural land covers such as bare soil \cite{zhong2017spectral, golipour2015integrating, khodadadzadeh2014subspace, makantasis2015deep} and meadows \cite{liu2020multitask}, or other unrelated materials such as asphalt \cite{jamshidpour2020ga,liu2019kernel}.
     
    The assumption that all test data are known in machine learning is called a closed world or closed set  \cite{scheirer2012toward, bendale2015towards}. 
    In the closed world, given an unknown instance that cannot be classified as one of the known classes, a model will eventually assign a known class label to the unknown instance because it cannot identify it as unknown. In this case, if the model is used to map certain crop types, it is inevitable to overestimate their area and, therefore, the total amount of food supplies.  

    The closed-world setting is not enough for accurate hyperspectral image classification. The machine needs to tell the unknown from known classes. This is impossible for many algorithms, as they assume a closed-world setting. For example, the deep neural network is limited by the SoftMax function, which constrains the output probabilities to have a summation of one, and each is within zero to one. In this scenario, one of the known classes must be assigned for the input instance. A simple solution is using confidence thresholding, rejecting those with low classification confidence. However, low classification confidence is not the assurance of the unknown; instead, it is uncertain \cite{scheirer2012toward}. 
An uncertain instance may be a valuable sample close to the classification hyperplane’s boundary for active learning, instead of the unknown \cite{tuia2011using}. 

    To tackle the open world problem, some methods were proposed for daily image recognition. First named as open-set recognition by \cite{scheirer2012toward}, they introduced 1-vs-set machine to empower SVM with the capability to identify the unknown. 
     The OpenMax method is the first one tackling this problem under the context of deep learning~\cite{bendale2016towards}, which enables the network to estimate the probability of one instance being the unknown class. It used the activation vectors (the final layer's output features before SoftMax) to estimate the distance distribution of each class and then applied the extreme value theory to recalibrate the activation vector with $C$+1 classes, where the added class was the unknown.    
      \cite{neal2018open} proposed to use counterfactual images generated from generative adversarial networks as the unknown to train the classifier.  
      Furthermore,
      \cite{dhamija2018reducing} argued to use the background classes to reduce the network agnostophobia. 
      Further, \cite{yoshihashi2019classification} added an auxiliary task to enhance the latent features for the OpenMax method. This enhanced method for open-set recognition, named CROSR, was tested with anomaly detection algorithms, i.e., IsoForest and one-class SVM, in a class-wise fashion, to exploit the latent features effectively. The method achieved the current state of the art in open-set recognition. However, like other methods, a large amount of training samples were still needed to estimate the centroid of each class, limiting its performance in hyperspectral image classification.

    The above-mentioned methods for daily image recognition are mostly centroid-based, where the unknown score is calculated in a class-wise fashion, which limits their usage under the few-shot context. For example, the CIFAR dataset consists of 600 samples per class \cite{krizhevsky2009cifar}, and the fashion-MNIST consists of 7000 samples per class \cite{xiao2017fashion}. In hyperspectral image classification, the size of the training set is significantly smaller than the daily image recognition as collecting ground truth data is labor-intensive and time-consuming. Sometimes, only 5 to 20 instances per class are available for training \cite{liu2018deep, pan2018mugnet, xu2019abundance}, limiting the usage of these open-set methods for hyperspectral image classification. 
    Another difference between daily image recognition and hyperspectral image classification is that the number of potential classes is smaller in the latter, as hyperspectral images are taken from the same direction (overhead), whereas daily images are more complex.  For hyperspectral images, it is rare to have more than 20 classes in a dataset; but, for daily image recognition, the number of classes already exceeds one thousand, e.g., ImageNet \cite{deng2009imagenet}. As a direct result of fewer potential classes, the openness (unknown degree) of hyperspectral image classification is significantly smaller than daily image recognition. Often, the unknown or novel instances count for only a small portion of the study area, which is called the tail class in technical terms, e.g., swimming pools and vehicles.  The low openness significantly affects the performance of current open-set techniques as they are designed to face a larger open world \cite{geng2020collective}. 
    In summary (as shown in Table \ref{tab:comparison}), due to the three notable differences between hyperspectral image classification and daily image open-set recognition (few-shot, fewer classes, and low openness), it is difficult to directly use open-set techniques developed for daily image recognition in hyperspectral image classification.

   In this paper, tackling the few-shot nature, we propose a novel multitask deep learning method for hyperspectral image classification with unknown classes in the open world (MDL4OW), which is free from estimating the class centroid in a class-wise fashion. The proposed method utilizes multitask learning to conduct classification and reconstruction simultaneously. The unknown score is calculated by comparing the original and reconstructed data, whereas the unknown should be poorly reconstructed due to the lack of labels. In this way, the proposed MDL4OW does not need to calculate the mean activation vectors or latent features in a class-wise fashion. Therefore, it is suitable for few-shot hyperspectral image classification.
   The major contributions of this study are summarized as follows:
\begin{itemize}
\item We propose a novel multitask deep learning method named MDL4OW for hyperspectral image classification with unknown classes.  The proposed method can identify unknown classes and significantly improve the classification accuracy. 
\item Instead of estimating the unknown score in a class-wise manner using centroid-based methods, the proposed method can estimate the unknown score with all data using the statistical model extreme value theory. Therefore, the proposed method outperformed the state-of-the-art open-set technique for few-shot hyperspectral image classification.
\item A new evaluation metric, the mapping error, is proposed to evaluate the accuracy of hyperspectral image classification with unknown classes. This metric is especially sensitive to imbalanced classification, which is often the case in hyperspectral images.  
\end{itemize}

\begin{table*}[!t]
  \centering
  \caption{Comparison among hyperspectral image classification with unknown classes and the existing tasks.}
    \begin{tabular}{cccccc}
    \toprule
    \toprule
    Task  & Few-shot & Unknown classes & Openness & \#Class & \#Channel \\
    \midrule
    Daily photo recognition & Rare (often \textgreater  500)  & Yes & High  & 100+  & 1-4 \\
    Hyperspectral image classification & Often (\textless 50)  & No  & 0  & 20-   & Up to 300+ \\
    Hyperspectral image classification with unknown classes & Often (\textless 50)  & Yes  & Low  & 20-   & Up to 300+ \\
    \bottomrule
    \bottomrule
    \end{tabular}%
  \label{tab:comparison}%
\end{table*}%

The remainder of this paper is organized as follows. In the next section, we describe the methodology in detail.  In section \ref{sec:dataset_setup}, we introduce the evaluation metrics and  the experimental setup. 
In section \ref{sec:result}, we discuss the experimental results obtained on real-world datasets and provide analyses.  Finally, we present our  conclusions in section \ref{sec:conclusion}.

\begin{figure*}[htbp] 
\centering      
\includegraphics[width=.99\textwidth]{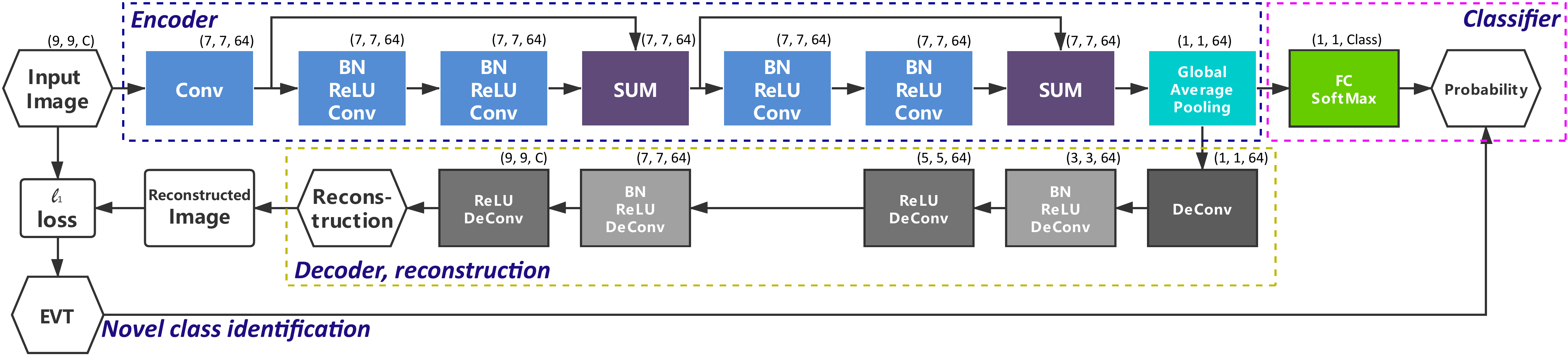}
\caption{Flowchart of the proposed method. The multitask network used here takes an input of 9 $\times$ 9 $\times$ \textit{Channel}. The first part of the network is an encoder/feature extractor with two residual units and a global average pooling layer. After extracting the latent features, a fully connected layer with the softmax function serves as the classifier and outputs the probability to the known classes. The reconstruction task here uses the deconvolutional layers to increase the spatial dimension of the latent features gradually. The output of the reconstruction task is a 9 $\times$ 9 $\times$ \textit{Channel} instance, which should be similar to the input data by minimizing the $\ell_1$ loss. The reconstruction loss is modeled by EVT to separate the known and unknown classes.}
\label{fig:flowchart}
\end{figure*}

\section{Methodology}
\label{sec:methodology}
In this section, we review a single-task classification network's components, including a feature extractor (encoder) and a classifier. Then, we describe the proposed multitask network in detail, including an encoder-decoder structure which  is used to reconstruct the input instance. Based on the assumption that instances with labels from the classification system can be better reconstructed than the unknown,  those with a large reconstructed loss should be rejected and considered as unknown. Finally, extreme value theory is used to separate the unknown from known classes.

\subsection{Deep CNNs for classification}
The proposed network with the flowchart is shown in Figure \ref{fig:flowchart}, which consists of two residual learning blocks \cite{he2016deep} and a global average pooling layer \cite{lin2013network} as the encoder, a fully connected layer with the SoftMax function as the classifier, and a decoder with deconvolutional layers \cite{pan2016shallow}. The classification part is a modified version of HResNet \cite{liu2020multitask} when adding the global average pooling layer. 

The powerful ability of deep CNNs for hyperspectral image classification tasks lies in the usage of convolutional layers to extract spectral-spatial information \cite{zhang2016deep}. Considering $X$ as the sample instance space, where each instance $x \in X$, given a limited training set with index $k$ consisting of only a few samples $\left(x^k, l^k \right)$, where $l^k \in L = \{1,..., |C|\}$ is the label index for $x^k$, the multilevel convolutional layers, along with batch normalization \cite{ioffe2015batch}, and the rectified linear unit (ReLU)  \cite{nair2010rectified}, serve as a good encoder $\phi  \left(\cdot \right)$ to extract the spectral-spatial features $x_{\phi}$ as the representation of the sample instance:
   \begin{equation}
   x_{\phi} = \phi \left(x\right). 
   \end{equation}
   
    In the encoder, the output of a convolutional layer $\phi_{conv} \left(\cdot \right)$ can be simplified as
\begin{equation}
x_{\phi_{conv}} = \phi_{conv} \left(x\right).
\end{equation}

Then, the classifier $f\left(\cdot\right)$ takes the output vector $x_{\phi}$ from the feature extractor $\phi\left(\cdot\right)$ as its input. 
In a pure deep learning scenario,  the fully connected layer with the SoftMax activation function serves as the classifier $f(\cdot)$ and gives the probability $P\left(y=j | x_{\phi} \right)$ of the $j$-th category:
\begin{equation}
P \left(y=j|x_{\phi} \right) = \frac{\exp \left(x_{\phi}^T w_{j}+b_j \right)}{\sum_{c=1}^C{\exp \left(x_{\phi}^T w_c + b_c \right)}},
\end{equation}
where $w_j$ is the weight vector of the $j$-th neuron in the fully connected layer, $b_j$ is a bias element corresponding to the $j$-th neural, and $C$ is the number of the category.  The classification task is to find the optimal parameters for the network  by minimizing the   cross-entropy loss function  $\ell_c$:
   \begin{equation}
   \ell_c \left(y,\hat{y} \right) =  - \sum_{i=1}^{C} y_i log \left(\hat{y}_i \right),
   \end{equation}
   where $C$ is the number of predefined classes,  $y$ is the ground truth label, and $\hat{y}$ is the predicted label.

As mentioned, the SoftMax function transforms the score vector into a probability vector, and the class with the largest probability is considered the predicted class. A naive solution to identify unknown classes is considering those instances with the largest probability smaller than 0.5 as unknown, which  is one of the baselines in the experiments (SoftMax with threshold = 0.5).

\subsection{Reconstruction via multitask learning}
So far, the feature extractor and classifier together form a decent machine for classification, but the network cannot still identify the unknown. To empower the network with this ability, we add a reconstruction task:
\begin{equation}
\hat{x} = f_r \left(x_{\phi} \right),
\end{equation}
where $\hat{x}$ is the reconstructed instance, $f_r \left(\cdot \right)$ is the reconstruction function or named the decoder, and $x_{\phi}$ is the output latent features from the encoder $\phi \left(\cdot \right)$. 
Here, we use $\ell_1$ distance as the reconstruction loss,
\begin{equation}
\ell_r \left(x, \hat{x} \right) = \left\Vert x - \hat{x} \right\Vert_1.
\end{equation}
In the training phase for the multitask network, we minimize the total loss via backpropagation, 
\begin{equation}
\min\limits_{\phi\left(\cdot\right),f_c\left(\cdot\right),f_r\left(\cdot\right)} \lambda_c \ell_c \left(y, \hat{y} \right) + \lambda_r \ell_r \left(x, \hat{x} \right),
\end{equation}
where $\lambda_c$ and $\lambda_r$ are the weights to control the loss influence on the multitask network for  $\ell_c$ and  $\ell_r$, respectively. 

In the decoder, the key element is the deconvolutional layer, also called the transposed convolutional layer. It can be considered as the inverse of  a convolutional layer, denoted as $\phi^{\dagger}_{conv} \left(\cdot \right)$, resulting:
\begin{equation}
x = \phi^{\dagger}_{conv} \left(\phi_{conv} \left(x \right) \right).
\end{equation}
Using the deconvolutional layer, we can gradually increase the 1$\times$1$\times$\textit{channel} latent features to patch-based hyperspectral sample instances. As shown in the flowchart (Figure \ref{fig:flowchart}), the reconstruction branch of the proposed method consists of a total of five deconvolutional layers, where all increase the spatial dimension of the instance except for the first one.

\subsection{Threshold setting with extreme value theory}

\begin{figure}[!t] 
\centering
\subfigure[Training data]{\includegraphics[width=0.23\textwidth]{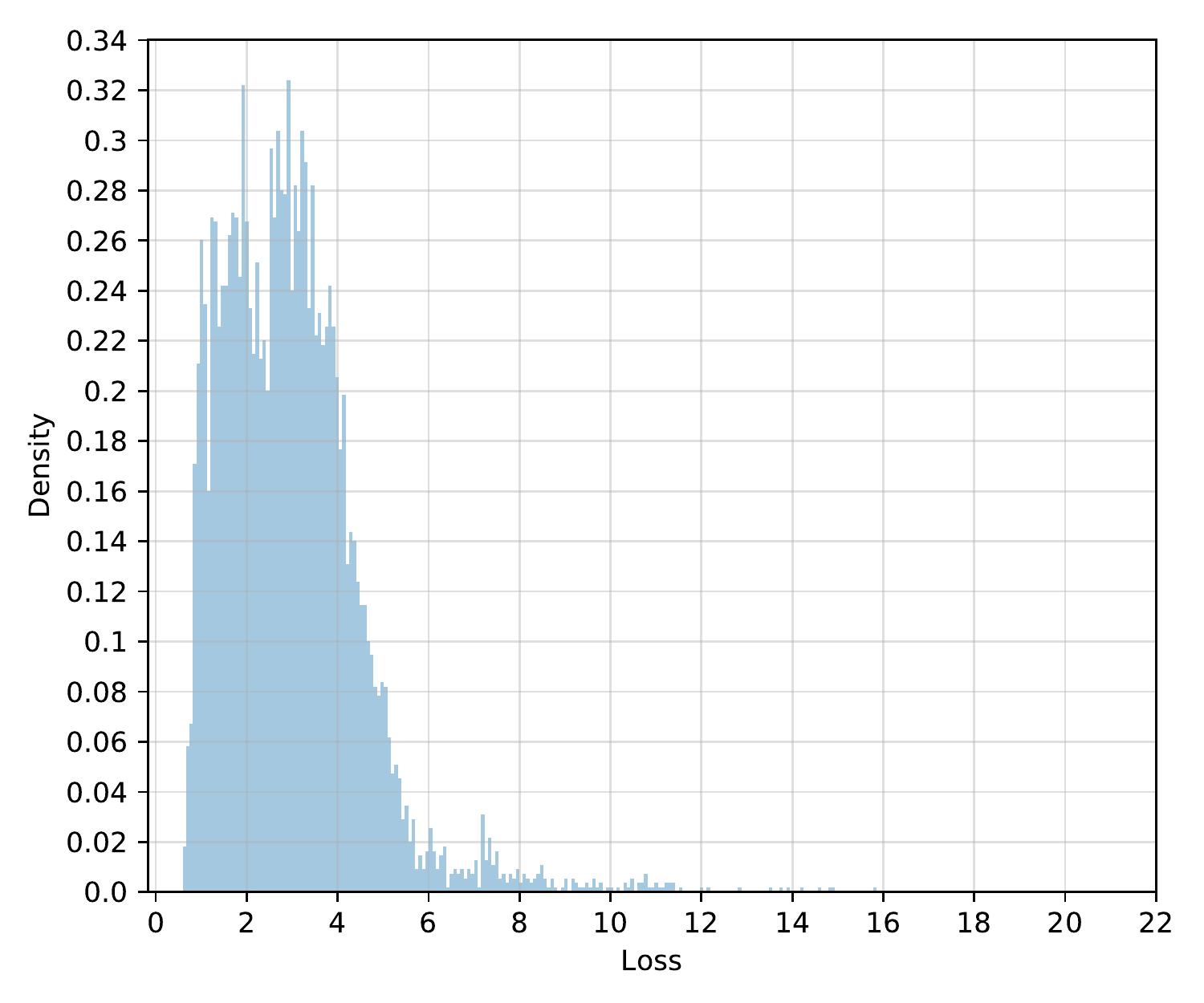}}
\subfigure[Test data]{\includegraphics[width=0.23\textwidth]{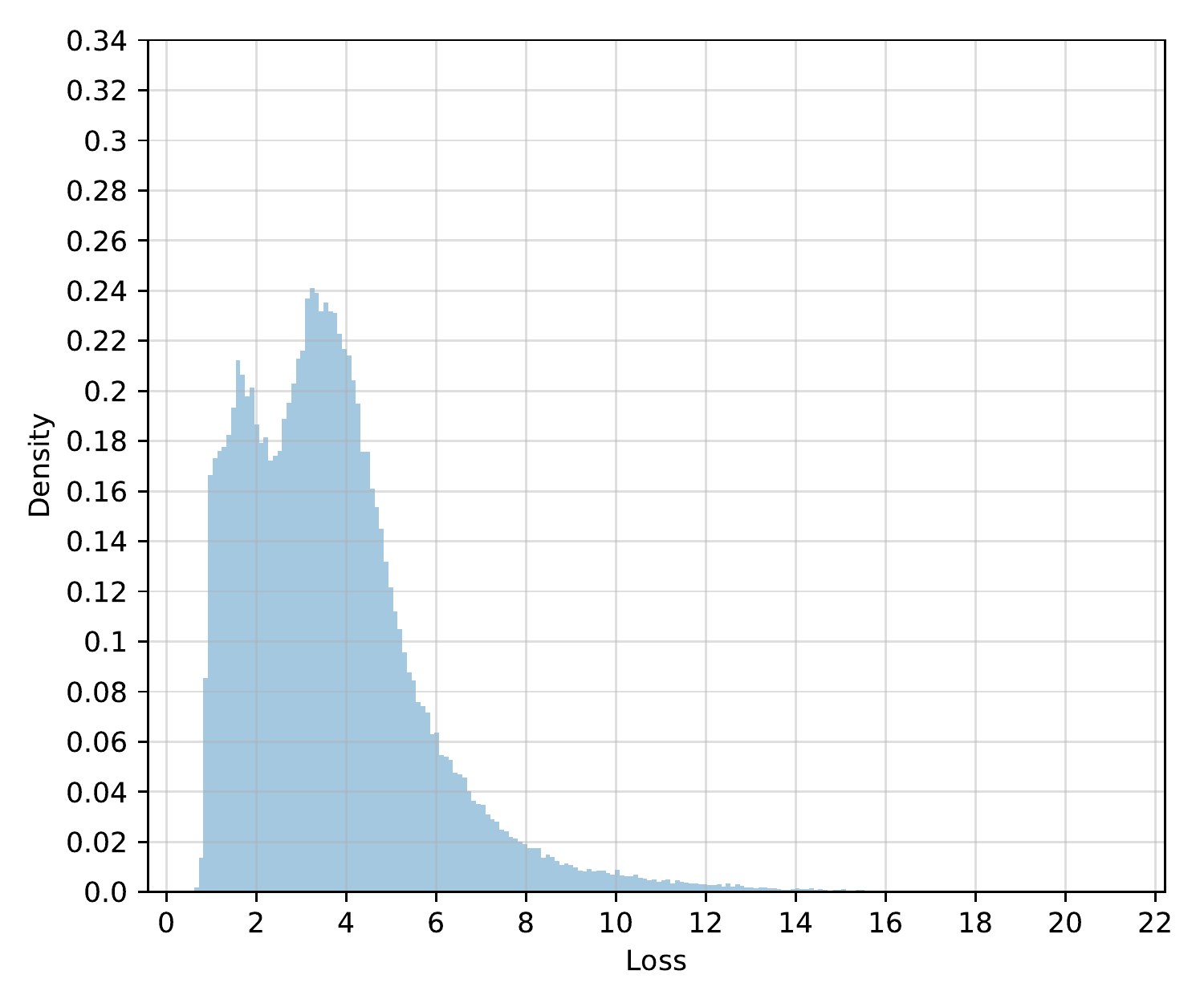}}   
\caption{Histogram samples of the reconstruction loss.}
\label{fig:loss_eaxample}
\end{figure}

After obtaining the reconstruction losses (an example is shown in Figure \ref{fig:loss_eaxample}), where a larger loss indicates that the deep learning model is not sufficiently optimized for the instance; the tail part of the loss distribution should be considered as unknown to the model. Here, we adopt the extreme value theory (EVT) to find the unknown classes by modeling a distribution of the tail part. 

The EVT suggests that the tail should be a Weibull distribution \cite{pickands1975statistical}. For a large class of distributions, $V$, and a large enough threshold, $w$, with $\{V_1,...,V_n \}$, $n$ independent and identically distributed samples, the cumulative distribution function can be approximated by a generalized pareto distribution (GPD):
\begin{align}
P \left(v - w \leq v | v > w \right) 
&= \frac{F_V \left(w+v \right)-F_V \left(v \right)}{1-F_V(w)} 
\\ &= G_{\xi, \mu} \left(v \right)
\label{eq:gev}
\end{align}
where,
\begin{equation}
G_{\xi, \mu} \left(v \right) = 
\begin{dcases}
1 - \left(1+ \frac{\xi \cdot v}{\mu}\right)^{-\frac{1}{\xi}} \ \ \ \ \ \ \ \xi \neq 0\\
1 - e^{{-\frac{v}{\mu}}} \ \ \ \ \ \ \ \ \ \ \ \ \ \xi = 0\\
\end{dcases}
\end{equation}
where $\mu > 0$, and $v \geq 0$ when $\xi \geq 0$, and $0 \leq v \leq -\frac{\mu}{\xi}$ when $\xi < 0$.
The parameters $\xi$ and $\mu$ can be estimated from the given tail data  \cite{grimshaw1993computing}. Here, $v$ is the reconstruction loss, and $G_{\xi, \mu} \left(\cdot \right)$ is the cumulative distribution function of GPD.  The pseudocode for identifying the unknown class after obtaining the trained network is shown in Algorithm \ref{alg:mdl4ow}.

To model the tail distribution, one defines the tail size, $\tau$, and a threshold value, $z$, to decide the boundary score between known and novel. The threshold is often set as 0.5 \cite{bendale2016towards,yoshihashi2019classification}, whereas the  $\tau$ value should be at least two to model a distribution. The proposed method MDL4OW is free from class centroid estimation, which enhances its performance under the few-shot context. For comparison purposes, the class-wise version MDL4OW/C is also applied.  The two methods are similar, except that MDL4OW is applied in a global fashion and MDL4OW/C is applied a class-wise fashion. Details of the threshold setting in the experiments are shown in section \ref{ssec:setup}  experimental setup.

 \begin{algorithm}[!t]
 \color{black}
 \caption{MDL4OW}
 \begin{algorithmic}[1]
 \renewcommand{\algorithmicrequire}{\textbf{Input:}}
 \renewcommand{\algorithmicensure}{\textbf{Output:}}
 \REQUIRE $X, \phi \left(\cdot\right), f\left(\cdot\right), f_r\left(\cdot\right), G_{\xi, \mu} \left(v \right), \tau, z$
 \ENSURE  $y_{pred}$
  \STATE Latent features: $X_{\phi} = \phi(X)$
  \STATE Predicted known class: $y_{pred} = f\left(X_{\phi}\right)$
  \STATE Reconstruction: $\hat{X} = f_r\left(X_{\phi}\right)$
  \STATE Reconstruction error:  $v = ||X-\hat{X}||_1$
  \STATE Compute EVT model  $G_{\xi, \mu} \left(v \right)$ based on $v, \tau$
  \IF {$G_{\xi, \mu} \left(v \right) < z $}
  \STATE $y_{pred} = y_{pred}$
  \ELSE 
  \STATE $y_{pred} = unknown$
  \ENDIF
 \RETURN $y_{pred}$
 \end{algorithmic}
 \label{alg:mdl4ow}
 \end{algorithm}

\section{Datasets and Experimental setup}
\label{sec:dataset_setup}
\subsection{Datasets}
\subsubsection{University of Pavia}
\begin{table}[!t]
  \centering
  \caption{The University of Pavia}
    \begin{tabular}{ccc}
    \toprule
    \toprule
    Index & Name  & \#sample \\
    \midrule
    1     & Asphalt & 6631 \\
    2     & Meadows & 18649 \\
    3     & Gravel & 2099 \\
    4     & Trees & 3064 \\
    5     & Metal sheet & 1345 \\
    6     & Bare soil & 5029 \\
    7     & Bitumen & 1330 \\
    8     & Brick & 3682 \\
    9     & Shadow & 947 \\
          & Novel & 5163 \\
    \midrule
          & Total & 47939 \\
    \bottomrule
    \bottomrule
    \end{tabular}%
  \label{tab:dataset_pavia}%
\end{table}%

The first dataset used in the experiment is the previously mentioned benchmark, the University of Pavia. This image was captured by the ROSIS sensor over the University of Pavia campus in 2001. After removal of 12 bands affected by noise and water absorption, it contains 103 bands with a scene size of 610$\times$340 in a GSD of 1.3 \textit{m}. The standard reference map contains nine classes.
As mentioned in the introduction section, some of the unannotated buildings have a very different spectral profile as compared to the known land covers. We manually annotated six additional  classes, as shown in the reference map in Figure \ref{fig:pavia_maps}. The number of supporting reference pixels is available in Table \ref{tab:dataset_pavia}.

\subsubsection{Salinas Valley}
\begin{table}[!t]
  \centering
  \caption{Salinas Valley}
    \begin{tabular}{ccc}
    \toprule
    \toprule
    Index & Name  & \#sample \\
    \midrule
    1     & Weeds-1 & 2009 \\
    2     & Weeds-2 & 3726 \\
    3     & Fallow & 1976 \\
    4     & Fallow-P & 1394 \\
    5     & Fallow-S & 2678 \\
    6     & Stubble & 3959 \\
    7     & Celery & 3579 \\
    8     & Grapes & 11271 \\
    9     & Soil  & 6203 \\
    10    & Corn  & 3278 \\
    11    & Lettuce-4wk & 1068 \\
    12    & Lettuce-5wk & 1927 \\
    13    & Lettuce-6wk & 916 \\
    14    & Lettuce-7wk & 1070 \\
    15    & Vinyard-U & 7268 \\
    16    & Vinyard-T & 1807 \\
          & Novel & 5613 \\
    \midrule
          & Total & 59742 \\
    \bottomrule
    \bottomrule
    \end{tabular}%
  \label{tab:dataset_salinas}%
\end{table}%
The Salinas Valley dataset was used in the experiment to evaluate the proposed method in fine-grained crop mapping using hyperspectral data. This image was collected by the AVIRIS sensor over Salinas Valley, California, with a GSD of 3.7 \textit{m}. The region covers an area of 512$\times$217 pixels with 204 bands after 20 water absorption bands were discarded. The data were at-sensor radiance data without atmospheric correction. A total of 16 agricultural land covers are identified in the image. However, some of the man-made materials, such as buildings, are not represented in the classification system and should not be classified as any of the known crops. We manually annotated these man-made materials and a water pool as the unknown, as shown in Figure \ref{fig:salinas_maps} reference map. Details of the dataset can be found in Table \ref{tab:dataset_salinas}.

\subsubsection{Indian Pines}
\begin{table}[!t]
  \centering
  \caption{Indian Pines}
    \begin{tabular}{ccc}
    \toprule
    \toprule
    Index & Name  & \#sample \\
    \midrule
    1     & Corn-notill & 1428 \\
    2     & Corn-mintill & 830 \\
    3     & Grass-pasture & 483 \\
    4     & Hay-windrowed & 478 \\
    5     & Soybean-notill & 972 \\
    6     & Soybean-mintill & 2454 \\
    7     & Soybean-clean & 593 \\
    8     & Woods & 1265 \\
          & Novel & 2007 \\
    \midrule
          & Total & 10510 \\
    \bottomrule
    \bottomrule
    \end{tabular}%
  \label{tab:dataset_indian}%
\end{table}%
The Indian Pines dataset is used to illustrate the potential of the proposed method. The dataset was  captured by the AVIRIS sensor in 1992 over the Indian Pines test site in Northwest Tippecanoe County, Indiana. The image consists of 145$\times$145 pixels with a GSD of 20 \textit{m}. A total of 16 classes were annotated in the reference data, and most were agricultural land covers. A road exists on the upper side of the image, which should  not belong to any of the known classes. We manually annotated the road, as shown in Figure \ref{fig:indian_maps} reference map. Since some of the classes only have less than ten instances, these tail classes were discarded in the experiment following \cite{lee2017going} and were considered as unknown, resulting in eight known classes.  Details of the dataset can be found in Table \ref{tab:dataset_indian}.

\subsection{Evaluation metrics}
\subsubsection{Openness}
Openness indicates the unknown or open degree of a dataset for open-world classification. It can be calculated from the number of training classes and the number of test classes \cite{geng2020recent}:
\begin{equation}
\color{black}
\textit{Openness} = 1 - \sqrt{\frac{2\times N_{train}}{N_{test}+N_{train}}}.
\end{equation}
Low openness is one of the differences between daily image recognition and hyperspectral image classification. Some studies pointed out that  the performance of certain methods could be degraded with a low openness value \cite{geng2020collective}. 
The openness values for the three hyperspectral datasets are 13\%, 3\%,  and 20\%, whereas the openness for daily image recognition is up to 63\% \cite{oza2019c2ae}.

\subsubsection{Open overall accuracy (OA)}
The evaluation of open-world classification is critical and diverse because of the inclusion of unknown classes \cite{geng2020recent}.
A naive solution is extending the current OA evaluation metric from $C$ classes to $C+1$ classes, where the added is the unknown class.
\begin{equation}
\textit{Open OA} = \frac{\sum^{C+1}_{i=1} {(TP_i+TN_i)}}{\sum^{C+1}_{i=1} (TP_i+TN_i+FP_i+FN_i)},
\end{equation}
where $TP_i$, $TN_i$, $FP_i$, and $FN_i$ are true positive, true negative, false positive, and false negative of the $i$-th class.
In hyperspectral image classification, we are interested in the accuracy of known land cover classes instead of the unknown. However, the open OA metric takes the accuracy of the unknown class into account. If the known classes only account for a small portion of the data, this metric will be dominated by the unknown classes. 

The open OA is similar to the closed OA (the one normally used in classification) but includes the unknown class. For reference purposes, we show the closed OA below:
\begin{equation}
\textit{Closed OA} = \frac{\sum^{C}_{i=1} {(TP_i+TN_i)}}{\sum^{C}_{i=1} (TP_i+TN_i+FP_i+FN_i)}.
\end{equation}

\subsubsection{Micro F1 score}
A more reasonable indicator should be the micro F1 score, where we only consider the precision and recall of known classes. Note the F1 score of the unknown class should not be considered as pointed by \cite{geng2020recent}. With micro F1, the inclusion of unknown classes plays an important role in the precision scores. If unknown instances are classified into one of the known classes, precision will be decreased. The micro F1 score is calculated as follows, 
\begin{equation}
\textit{F1} = \frac{2 \times \textit{Precision} \times \textit{Recall}}{\textit{Precision} + \textit{Recall}}
\end{equation} 
where, 
\begin{equation}
\textit{Precision} =  \frac{\sum^{C}_{i=1} TP_i}{\sum^{C}_{i=1} (TP_i+FP_i)},
\end{equation}
\begin{equation}
\textit{Recall} =  \frac{\sum^{C}_{i=1} TP_i}{\sum^{C}_{i=1} (TP_i+FN_i)}.
\end{equation}

\subsubsection{Mapping error}
The mapping error in a specific region is of direct interest to local governments, as policy-makers rely on food production data collected based on administrative regions to adapt their strategy. In this study, the mapping error is defined as,
\begin{equation}
\begin{aligned}
Error = \frac{\sum_{i=1}^{C} {|A_{p,i} - A_{gt,i}|}}{\sum_{i=1}^{C}{ A_{gt,i}}} \\
\quad s.t. \quad A_{i} \geq 0,
\quad  \sum_{i=1}^{C+1}{A_{p,i}} = \sum_{i=1}^{C+1} A_{gt,i}
\end{aligned}
\label{eq:error}
\end{equation}
where $A_{p,i}$ is the predicted area of the $i$-th class, $A_{gt,i}$ is the ground truth area, $C$ is the number of known classes, and $C+1$ class is the unknown. Note for the closed set classification, the mapping error should be, 
\begin{equation}
\begin{aligned}
Error_{close} = \frac{\sum_{i=1}^{C} {|A_{p,i} - A_{gt,i}|}}{\sum_{i=1}^{C}{ A_{gt,i}}} \\
\quad s.t. \quad A_{i} \geq 0,
\quad  \sum_{i=1}^{C}{A_{p,i}} = \sum_{i=1}^{C} A_{gt,i}.
\end{aligned}
\end{equation}
If a model overestimates a land cover class, it will also underestimate another land cover class. Note in the open-world setting, the upper bound of the individual class in theory is  $\sum_{i=1}^{C+1} A_{gt,i}$, whereas the upper bound for the closed world assumption is $\sum_{i=1}^{C} A_{gt,i}$.
    Therefore, the range of mapping error in the closed setting is 0--2, whereas the maximum error in open-world classification depends on the unknown instances included for evaluation and is calculated as $Error_{max} = 2 \times (1 + A_{gt,C+1}/\sum_{i=1}^{C}{A_{gt,i}})$.

    Mapping error is not a point-by-point evaluation metric, meaning a zero error does not necessarily lead to zero classification confusion. However, for common multiclass scenarios, it is nearly impossible for a model to have a correct guess on the mapping area randomly. As a well-classified map is supposed  to have a good estimation of the land cover area, mapping error  is the most simple and direct metric to evaluate the accuracy of hyperspectral image classification, especially with the existence of unknown classes.   Mapping error is also sensitive to imbalanced mapping (or in technical terms, the long tail), which is a great challenge when we are facing the open world \cite{liu2019large}. We show a toy example in Table \ref{tab:error_example}, where three models achieve an identical OA but different errors. A good model, like Model 1,  should carefully take care of the tail classes to reduce the mapping error. In contrast, an imbalanced model, although achieving the same OA, neglects the tail classes and thus leads to extreme overestimation and underestimation of the number and area of each land cover class. 

\begin{table}[htbp]
  \centering
  \caption{A simple example illustrating mapping error's sensitivity to imbalanced mapping (the long tail). Note $a$/$b$ indicates the model predicts $b$ instances belong to the class, where $a$ instances are correctly classified.}
    \begin{tabular}{cccccc}
    \toprule
    \toprule
          & C1    & C2    & C3    & OA    & Error \\
    \midrule
    \midrule
    GT    & 80    & 10    & 10    & -     & - \\
    Model 1 & 70/80 & 5/10  & 5/10  & 80/100    & 0 \\
    Model 2 & 78/88 & 1/6   & 1/6   & 80/100    & 16/100 \\
    Model 3 & 80/100 & 0     & 0     & 80/100    & 40/100 \\
    \bottomrule
    \bottomrule
    \end{tabular}%
  \label{tab:error_example}%
\end{table}%

\subsection{Experimental setup}
\label{ssec:setup}
Experiments were carried out under both the few-shot context (20 training samples per class) and many-shot context (200 per class). 
The experiments were conducted using TensorFlow 1.9 \cite{abadi2016tensorflow} and Keras 2.1.6 \cite{chollet2015keras} with Python 3.6 on a machine equipped with an Intel I5-8500 CPU, a Nvidia GTX 1080 Ti GPU and 32G RAM. The AdaDelta  optimizer \cite{zeiler2012adadelta} is used for backpropagation in the training phase. The learning rate was set as 1.0 for the first 170 epochs and then, as 0.1 for another 30 epochs before the training  stopped. We adopted the early stopping mechanism to accelerate the training. If the loss does not decrease for five epochs, the learning proceeds immediately to the next phase. To fully exploit the training samples, they were augmented horizontally, vertically, and diagonally; hence, the number of training samples  $NoS$ should be times four. 

In the experiments, the weights of classification and reconstruction losses are 0.5. The tail number is set as $NoS \times 4 \times 0.05 \times C$ for MDL4OW and $NoS \times 4 \times 0.05$ for MDL4OW/C, where $NoS$ is the number of training samples per class, $4$ is the number of augmentations of training samples, and $C$ is the number of training classes. If the tail number is smaller than 20 for MDL4OW, then it is set as 20. The tail number here is not the global optimum and will be discussed in section \ref{ssec:tail_sensitivity}.  All results are reported from the average of ten random training sets. We compare our method with random forest, support vector machine, deep contextual CNN (DCCNN) \cite{lee2017going}, wide contextual residual network (WCRN) \cite{liu2018wide}, the modified HResNet (the baseline of the proposed method) \cite{liu2020multitask}, and the state-of-the-art open-set method CROSR \cite{yoshihashi2019classification}. For open-set CNNs with confidence thresholding, the unknown  probability is determined with a threshold of 0.5.

\section{Results and analysis}

\begin{table}[!t]
\color{black}
  \centering
  \caption{Number of known and novel classes (in brackets) and pixels in the experiments.}
  \scalebox{0.8}[0.8]{
    \begin{tabular}{ccccc}
    \toprule
    \toprule
          &       & Pavia & Salinas & Indian \\
    \midrule
    \midrule
    \multirow{2}[1]{*}{Almost perfect} & Known Pixels (Classes) & 42776 (9) & 54129 (16) & 8503 (8) \\
          & Novel Pixels (Classes) & 5163 (6) & 5613 (2) & 2007 (9) \\
    \multirow{2}[0]{*}{OW1} & Known Pixels (Classes) & 14724 (3) & 10716 (3) & 2883 (3) \\
          & Novel Pixels (Classes) & 33215 (12) & 49026 (15) & 7627 (14) \\
    \multirow{2}[0]{*}{OW2} & Known Pixels (Classes) & 26742 (3) & 22483 (3) & 4019 (3) \\
          & Novel Pixels (Classes) & 21197 (12) & 37259 (15) & 6491 (14) \\
    \multirow{2}[1]{*}{OW3} & Known Pixels (Classes) & 11643 (3) & 24069 (3) & 3286 (3) \\
          & Novel Pixels (Classes) & 36296 (12) & 35673 (15) & 7224 (14) \\
    \bottomrule
    \bottomrule
    \end{tabular}}%
  \label{tab:setting}%
\end{table}%

The experiments were conducted in two parts. In the first part, we evaluated the proposed method in an almost perfect classification system. Under this setting, only a small portion of the instances belong to unknown or novel due to the long-tail nature of the open world. In the second part, we selected some known classes as training data and the other classes as the unknown to evaluate the proposed method. This is the standard protocol for the open classification of daily images in computer vision, and the number of unknown classes greatly exceeds the known classes. However, the second setting is rare in real-world hyperspectral image classification. The number of known and novel classes and their supporting pixels in each setting are shown in Table \ref{tab:setting}.

\label{sec:result}
\subsection{With almost perfect classification systems}
\subsubsection{Results on the University of Pavia dataset}
   In the first experiment, we applied the proposed method to the popular benchmark data from the University of Pavia.  
    In Tables \ref{tab:oa1_pu} and \ref{tab:oa2_pu}, we show the overall OA, F1, mapping error, and class-wise F1 obtained with 20 and 200 training samples per class. Under the few-shot setting, MDL4OW achieved the best classification in terms of three evaluation metrics (85.07\%, 0.9172, 14.03\%), whereas CROSR achieved the second-best F1 score (84.66\%, 0.9090, 14.95\%).  Under the many-shot setting, the best classification was obtained by MDL4OW/C. The OA and F1  are 90.53\% and 0.9482, 1.87\% and 0.0083 higher than the closed classification. When it comes to the mapping error, the proposed method significantly outperformed conventional methods. The mapping error was reduced by 4.9\% from 12.19\% to 7.29\%.
    
\begin{figure*}[!t]
\centering
\subfigure[Close]{\includegraphics[width=0.13\textwidth]{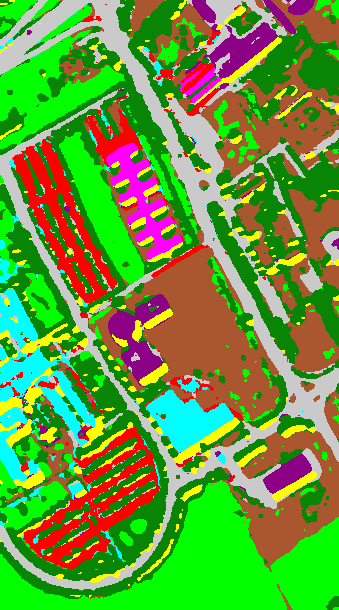}} 
\subfigure[CROSR]{\includegraphics[width=0.13\textwidth]{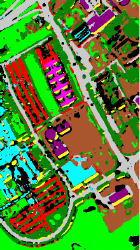}} 
\subfigure[MDL4OW]{\includegraphics[width=0.13\textwidth]{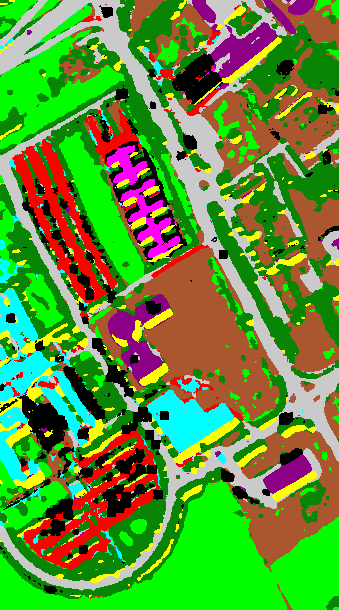}} 
\subfigure[MDL4OW/C]{\includegraphics[width=0.13\textwidth]{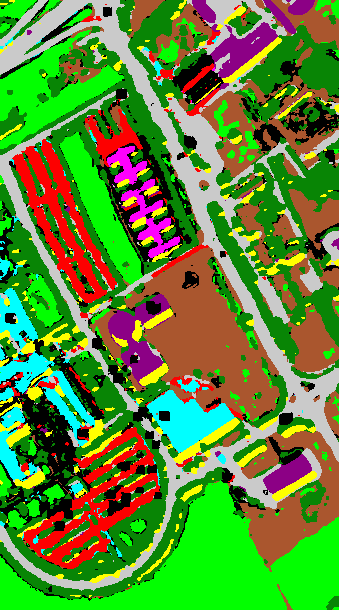}} 
\subfigure[Image]{\includegraphics[width=0.13\textwidth]{fig2/pu_im.png}}
\subfigure[Reference]{\includegraphics[width=0.13\textwidth]{fig2/paviaU_gt15.pdf}}
\subfigure{\includegraphics[width=0.13\textwidth]{fig2/paviaU_gt15_legend.pdf}}
\caption{Classification maps of the University of Pavia. Known classes: 1-9; unknown classes: 10-15. In the classification maps, those masked with black color are identified as unknown by the proposed method. }
\label{fig:pavia_maps}
\end{figure*}

    For illustrative purposes, we show the classification maps in Figure \ref{fig:pavia_maps}. We can see that the proposed method successfully identified the unknown classes (not annotated in the standard dataset) building-1, vehicles (the helicopter and some cars), the parking shed, and the swimming pool. Among the three open methods, only the CROSR identified building-2 as unknown, which is reflected in its high F1 score on the novel class. However, CROSR achieved a lower F1 in the second class meadows. This class covers a large area in the image, as indicated by the number of sample instances (Table \ref{tab:dataset_pavia}). As shown in the classification map (Figure \ref{fig:pavia_maps} CROSR), on the lower side of the image, some meadows were misclassified as unknown by CROSR. CROSR is a centroid-based method and is sensitive to the intraclass variation. Meadows consist of over 18000 instances and have a large intraclass variation, limiting the performance of centroid-based methods. Large intraclass variation is common in remote sensing images.  Therefore, the proposed MDL4OW is more robust and stable compared to CROSR. 

\begin{table*}[!t]
  \centering
  \caption{Classification accuracy using 20 samples per class on the University of Pavia dataset. * top-1 softmax probability $<$ 0.5 is considered as unknown. Error (\%): the smaller the better; else: the larger the better.}
  \scalebox{0.7}[0.7]{
    \begin{tabular}{c|ccccc|cccccc}
    \toprule
    \toprule
          & \multicolumn{5}{c|}{Close}            & \multicolumn{6}{c}{Open} \\
          & RF    & SVM   & DCCNN\cite{lee2017going} & WCRN\cite{liu2018wide}  & HResNet\cite{liu2020multitask} & DCCNN* & WCRN*  & HResNet* & CROSR \cite{yoshihashi2019classification} & MDL4OW  & MDL4OW/C \\
    \midrule
    \midrule
    1     & 76.34  & 75.82  & 84.60  & 85.50  & 91.66  & 84.60  & 85.52  & 91.72  & 85.87 & 91.56  & 89.78  \\
    2     & 69.17  & 81.67  & 91.15  & 93.77  & 95.57  & 91.17  & 93.76  & 95.57  & 93.37 & 95.57  & 93.84  \\
    3     & 51.82  & 60.32  & 57.50  & 67.46  & 79.82  & 57.78  & 67.71  & 80.15  & 83.41 & 82.70  & 81.95  \\
    4     & 68.44  & 78.96  & 89.98  & 91.09  & 92.16  & 90.08  & 91.23  & 92.35  & 92.96 & 91.69  & 89.47  \\
    5     & 64.38  & 67.13  & 84.71  & 89.02  & 88.46  & 84.92  & 89.11  & 88.78  & 95.83 & 90.27  & 91.88  \\
    6     & 40.06  & 53.04  & 65.76  & 67.50  & 74.81  & 65.81  & 67.68  & 75.00  & 77.91 & 76.37  & 78.32  \\
    7     & 61.01  & 62.79  & 64.78  & 72.19  & 72.24  & 65.03  & 72.59  & 72.94  & 90.57 & 75.71  & 76.01  \\
    8     & 67.41  & 75.17  & 72.24  & 70.35  & 82.53  & 72.41  & 70.50  & 82.62  & 84.38 & 83.93  & 80.42  \\
    9     & 99.79  & 99.88  & 91.27  & 91.23  & 86.63  & 91.51  & 91.46  & 87.25  & 93.19 & 86.72  & 84.51  \\
    Novel & 0.00  & 0.00  & 0.00  & 0.00  & 0.00  & 2.50  & 2.73  & 4.87  & 57.21 &  32.77  & 42.06  \\
    \midrule
    \midrule
    OA  (\%)   & 60.36$\pm$2.19  & 69.28$\pm$3.22  & 76.81$\pm$2.78  & 79.40$\pm$1.56  & 83.52$\pm$1.42  & 76.90$\pm$2.82  & 79.44$\pm$1.56  & 83.73$\pm$1.36 & 84.66$\pm$3.61 & \textbf{85.07$\pm$1.58}  & 83.35$\pm$2.43  \\
    F1$\times$100 & 75.25$\pm$1.71  & 81.81$\pm$2.30  & 86.86$\pm$1.81  & 88.51$\pm$0.97  & 91.01$\pm$0.84  & 86.89$\pm$1.83  & 88.51$\pm$0.97  & 91.11$\pm$0.81 & 90.90$\pm$2.28 & \textbf{91.72$\pm$0.96}  & 90.44$\pm$1.50  \\
    Error (\%) & 44.22$\pm$9.17  & 30.27$\pm$6.26  & 24.05$\pm$6.67  & 17.19$\pm$4.40  & 16.59$\pm$3.31 & 23.81$\pm$6.72  & 16.93$\pm$4.51  & 16.31$\pm$3.22 & 14.95$\pm$5.23  & \textbf{14.03$\pm$3.69}  & 14.31$\pm$3.69  \\
    \bottomrule
    \bottomrule
    \end{tabular}}%
  \label{tab:oa1_pu}%
\end{table*}%

\begin{table*}[!t]
  \centering
  \caption{Classification accuracy using 200 samples per class on the University of Pavia dataset. * top-1 softmax probability $<$ 0.5 is considered as unknown. Error (\%): the smaller the better; else: the larger the better.}
  \scalebox{0.7}[0.7]{
    \begin{tabular}{c|ccccc|cccccc}
    \toprule
    \toprule
          & \multicolumn{5}{c|}{Close}            & \multicolumn{6}{c}{Open} \\
          & RF    & SVM   & DCCNN\cite{lee2017going} & WCRN\cite{liu2018wide}  & HResNet\cite{liu2020multitask} & DCCNN* & WCRN*  & HResNet* & CROSR\cite{yoshihashi2019classification} & MDL4OW  & MDL4OW/C \\
    \midrule
    \midrule
    1     & 82.77  & 83.22  & 94.90  & 93.44  & 95.29  & 94.88  & 93.46  & 95.34 & 93.05 & 93.68  & 94.54  \\
    2     & 83.73  & 92.93  & 97.84  & 98.56  & 99.03  & 97.85  & 98.56  & 99.04 & 96.64 & 99.03  & 98.81  \\
    3     & 72.67  & 76.75  & 78.36  & 93.00  & 93.75  & 78.54  & 93.03  & 93.82 & 93.96 & 97.39  & 96.72  \\
    4     & 83.37  & 92.24  & 95.83  & 97.01  & 90.16  & 95.89  & 97.04  & 90.24 & 95.02 & 90.42  & 90.61  \\
    5     & 75.79  & 75.08  & 90.53  & 94.61  & 88.74  & 90.61  & 94.64  & 88.86 & 96.08 & 98.11  & 97.47  \\
    6     & 52.88  & 73.54  & 76.59  & 77.48  & 83.93  & 76.66  & 77.50  & 84.01 & 86.55 & 85.23  & 86.29  \\
    7     & 72.67  & 79.91  & 84.09  & 75.77  & 75.59  & 84.35  & 75.83  & 75.77 & 96.17 & 81.93  & 80.56  \\
    8     & 80.28  & 83.63  & 88.75  & 87.48  & 96.65  & 88.94  & 87.53  & 96.71 & 95.04 & 94.81  & 95.74  \\
    9     & 99.51  & 99.92  & 97.31  & 96.04  & 94.46  & 97.39  & 96.07  & 94.61 & 96.29 & 95.21  & 94.52  \\
    Novel & 0.00  & 0.00  & 0.00  & 0.00  & 0.00  & 1.85  & 0.52  & 1.91  & 61.70 & 38.43  & 45.94  \\
    \midrule
    \midrule
    OA (\%)    & 72.77$\pm$0.88  & 81.36$\pm$0.43  & 86.55$\pm$0.50  & 87.39$\pm$0.18  & 88.66$\pm$0.12  & 86.58$\pm$0.54  & 87.41$\pm$0.17  & 88.76$\pm$0.13 & 90.39$\pm$0.67 & 90.17$\pm$0.13  & \textbf{90.53$\pm$0.62}  \\
    F1$\times$100 & 84.24$\pm$0.59  & 89.72$\pm$0.26  & 92.78$\pm$0.29  & 93.27$\pm$0.10  & 93.99$\pm$0.07  & 92.80$\pm$0.31  & 93.28$\pm$0.10  & 94.03$\pm$0.07 & 94.51$\pm$0.37 & 94.66$\pm$0.08  & \textbf{94.82$\pm$0.33}  \\
    Error (\%) & 29.87$\pm$2.34  & 14.72$\pm$1.62  & 13.76$\pm$1.14  & 12.66$\pm$0.48  & 12.19$\pm$0.18  & 13.56$\pm$1.15  & 12.61$\pm$0.49  & 12.06$\pm$0.18 & 7.39$\pm$1.44 & 8.36$\pm$0.04  & \textbf{7.29$\pm$0.95}  \\
    \bottomrule
    \bottomrule
    \end{tabular}}%
  \label{tab:oa2_pu}%
\end{table*}%

\subsubsection{Results on the Salinas Valley dataset}
\begin{figure*}[!t]
\centering
\subfigure[Close]{\includegraphics[width=0.13\textwidth]{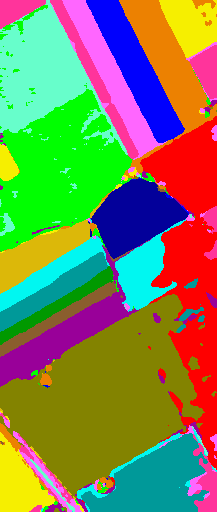}}
\subfigure[CROSR]{\includegraphics[width=0.13\textwidth]{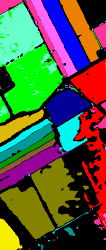}}
\subfigure[MDL4OW]{\includegraphics[width=0.13\textwidth]{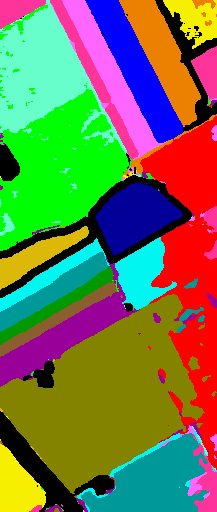}} 
\subfigure[MDL4OW/C]{\includegraphics[width=0.13\textwidth]{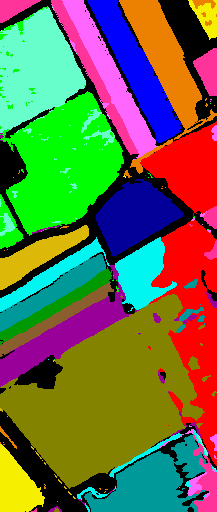}}
\subfigure[Image]{\includegraphics[width=0.13\textwidth]{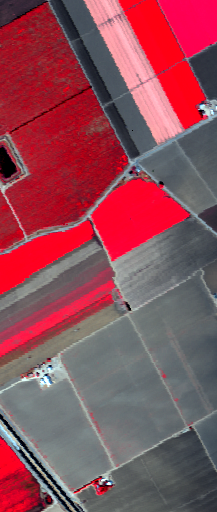}}  
\subfigure[Reference]{\includegraphics[width=0.13\textwidth]{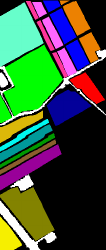}}
\subfigure{\includegraphics[width=0.13\textwidth]{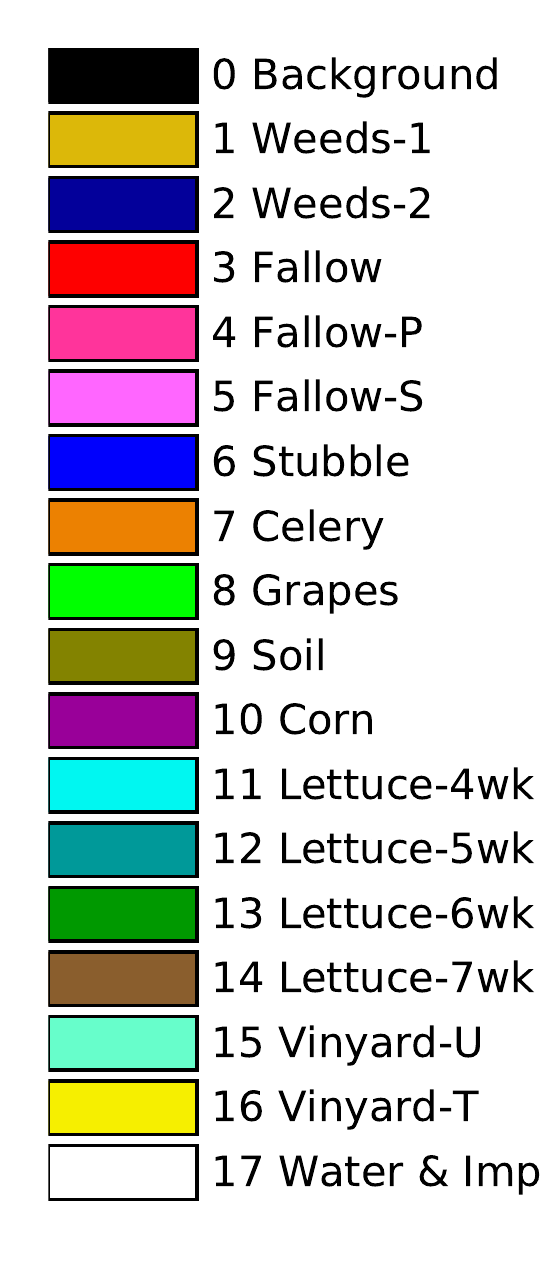}}
\caption{Classification maps of the Salinas Valley. Known classes: 1-16; unknown classes: 17. In the classification maps, those masked with black color are identified as unknown by the proposed method. }
\label{fig:salinas_maps}
\end{figure*}

To evaluate the potential of the proposed method in estimating crop area,  a popular agricultural  dataset,  the Salinas Valley, is used in the experiment. This dataset contains 16 agricultural classes, where some of the man-made materials are not considered (a road on the lower left and a water pool on the upper left). We manually annotated some of the man-made materials in white, as shown in Figure \ref{fig:salinas_maps} reference map. The classification results are presented in Tables \ref{tab:oa1_salinas} and \ref{tab:oa2_salinas} for few-shot and many-shot settings. Under the few-shot context, MDL4OW achieved the best classification result in terms of three evaluation metrics. Compared with the close classification, the OA increased by 4.94\% from 82.46\% to 87.40\%; the F1 score increased by 0.0235 from 0.9038 to 0.9273. The mapping error is more sensitive to the unknown classes and reduced from 23.24\% to 17.04\%, up to 6\%. This experiment strongly shows the potential of the proposed method in precisely estimating crop area. 
    As for the many-shot setting, the class-wise MDL4OW/C achieved the best classification results, followed by MDL4OW. The mapping error was significantly reduced from 10.84\% to 3.96\%.

   For illustrative purposes, we show the classification maps obtained from the many-shot setting in Figure \ref{fig:salinas_maps}. We can see that a lot of known instances (the lower part) are misclassified as unknown with the CROSR method. In contrast, the proposed MDL4OW successfully identified the unknown (a road on the lower left and a water pool on the upper left) while maintaining high accuracy on the known classes. As in remotely sensed hyperspectral images, the unknown should only count for a small portion of the image; the proposed method is more suitable for the classification of hyperspectral data.

\begin{table*}[!t]
  \centering
  \caption{Classification accuracy using 20 samples per class on the Salinas Valleys dataset. * top-1 softmax probability $<$ 0.5 is considered as unknown. Error (\%): the smaller the better; else: the larger the better.}
  \scalebox{0.7}[0.7]{
    \begin{tabular}{c|ccccc|cccccc}
    \toprule
    \toprule
          & \multicolumn{5}{c|}{Close}            & \multicolumn{6}{c}{Open} \\
          & RF    & SVM   & DCCNN\cite{lee2017going} & WCRN\cite{liu2018wide}  & HResNet\cite{liu2020multitask} & DCCNN* & WCRN*  & HResNet* & CROSR\cite{yoshihashi2019classification} & MDL4OW  & MDL4OW/C \\
    \midrule
    \midrule
    1     & 97.25  & 97.64  & 80.93  & 92.02  & 99.86  & 81.02  & 92.09  & 99.86 & 93.24 & 94.58  & 94.49  \\
    2     & 97.80  & 97.90  & 85.30  & 98.05  & 99.74  & 85.21  & 98.07  & 99.77 & 92.60 & 96.95  & 97.99  \\
    3     & 80.22  & 91.99  & 73.05  & 89.78  & 97.90  & 72.42  & 89.74  & 97.93 & 93.27 & 98.58  & 95.42  \\
    4     & 70.62  & 75.44  & 71.92  & 74.86  & 69.31  & 72.03  & 74.99  & 69.42 & 96.23 & 97.22  & 95.93  \\
    5     & 90.24  & 92.25  & 77.16  & 71.11  & 67.05  & 77.36  & 71.26  & 67.13 & 93.86 & 83.50  & 85.75  \\
    6     & 81.66  & 90.72  & 95.75  & 93.97  & 97.49  & 95.83  & 94.07  & 97.60 & 93.17 & 95.19  & 99.12  \\
    7     & 96.87  & 98.49  & 98.09  & 98.38  & 98.59  & 98.09  & 98.37  & 98.55 & 94.67 & 96.31  & 95.59  \\
    8     & 62.70  & 70.79  & 39.08  & 71.62  & 74.71  & 35.33  & 71.44  & 74.72 & 74.66 & 75.22  & 75.18  \\
    9     & 93.38  & 87.11  & 89.36  & 94.57  & 97.24  & 89.48  & 94.57  & 97.26 & 95.12 & 98.02  & 96.81  \\
    10    & 74.87  & 77.92  & 81.30  & 84.25  & 90.97  & 81.57  & 84.32  & 91.04 & 89.11 & 91.89  & 90.28  \\
    11    & 62.30  & 61.27  & 63.14  & 78.83  & 73.03  & 63.61  & 79.50  & 73.35 & 84.83 & 82.37  & 81.84  \\
    12    & 92.82  & 94.40  & 91.20  & 98.51  & 99.54  & 91.93  & 98.52  & 99.55 & 93.61 & 99.89  & 98.70  \\
    13    & 89.68  & 93.88  & 90.60  & 97.28  & 99.16  & 91.83  & 97.29  & 99.22 & 94.80 & 99.93  & 99.56  \\
    14    & 80.83  & 88.93  & 92.94  & 94.30  & 98.54  & 93.20  & 94.42  & 98.58 & 95.37 & 98.83  & 96.65  \\
    15    & 56.67  & 62.46  & 54.08  & 65.87  & 77.40  & 49.56  & 65.96  & 77.40 & 76.45 & 77.49  & 77.31  \\
    16    & 84.67  & 87.57  & 93.90  & 80.50  & 90.89  & 94.11  & 80.67  & 90.94 & 94.64 & 99.14  & 95.43  \\
    Novel & 0.00  & 0.00  & 0.00  & 0.00  & 0.00  & 4.21  & 2.32  & 1.69  & 64.94 & 74.39 &  75.97  \\
    \midrule
    \midrule
    OA (\%)  & 74.50$\pm$1.63  & 78.02$\pm$1.76   & 70.48$\pm$3.19   & 78.74$\pm$1.69   & 82.46$\pm$1.65   & 68.84$\pm$5.69   & 78.69$\pm$1.77   & 82.51$\pm$1.64 & 84.27$\pm$1.29 & \textbf{87.40$\pm$1.60}   & 87.12$\pm$1.57   \\
    F1$\times$100 & 85.38$\pm$1.08   & 87.64$\pm$1.11   & 82.64$\pm$2.17   & 88.09$\pm$1.06   & 90.38$\pm$1.39   & 81.35$\pm$4.19   & 88.05$\pm$1.11   & 90.41$\pm$1.41 & 90.57$\pm$0.85  & \textbf{92.73$\pm$1.36}   & 92.43$\pm$1.37   \\
    Error (\%) & 19.91$\pm$5.19   & 17.37$\pm$4.50   & 36.71$\pm$14.27   & 21.12$\pm$6.09   & 23.24$\pm$5.59   & 36.62$\pm$13.66  & 20.97$\pm$6.14   & 23.15$\pm$5.59  & 18.68$\pm$2.12 & \textbf{17.04$\pm$5.37}   & 18.20$\pm$5.67   \\
    \bottomrule
    \bottomrule
    \end{tabular}}%
  \label{tab:oa1_salinas}%
\end{table*}%

\begin{table*}[!t]
  \centering
  \caption{Classification accuracy using 200 samples per class on the Salinas Valleys dataset. * top-1 softmax probability $<$ 0.5 is considered as unknown. Error (\%): the smaller the better; else: the larger the better.}
  \scalebox{0.7}[0.7]{
    \begin{tabular}{c|ccccc|cccccc}
    \toprule
    \toprule
          & \multicolumn{5}{c|}{Close}            & \multicolumn{6}{c}{Open} \\
          & RF    & SVM   & DCCNN\cite{lee2017going} & WCRN\cite{liu2018wide}  & HResNet\cite{liu2020multitask} & DCCNN* & WCRN*  & HResNet* & CROSR\cite{yoshihashi2019classification} & MDL4OW  & MDL4OW/C \\
    \midrule
    \midrule
    1     & 99.66  & 99.83  & 84.00  & 87.70  & 91.18  & 84.10  & 87.81  & 91.30 & 96.58 & 94.01  & 94.60  \\
    2     & 99.63  & 99.73  & 99.35  & 99.99  & 99.93  & 99.40  & 99.99  & 99.94 & 97.05 & 97.74  & 99.04  \\
    3     & 91.53  & 97.53  & 93.92  & 88.70  & 88.94  & 94.18  & 88.79  & 89.02 & 96.25 & 92.18  & 98.45  \\
    4     & 71.94  & 81.47  & 88.35  & 80.35  & 64.48  & 88.37  & 80.45  & 64.53 & 97.12 & 98.19  & 98.55  \\
    5     & 92.54  & 85.04  & 73.29  & 74.36  & 87.25  & 73.37  & 74.46  & 87.36 & 96.97 & 95.17  & 96.47  \\
    6     & 85.91  & 93.21  & 98.14  & 97.94  & 98.80  & 98.19  & 98.00  & 98.83 & 96.82 & 99.73  & 97.26  \\
    7     & 98.56  & 99.34  & 99.51  & 99.68  & 95.57  & 99.53  & 99.68  & 95.60 & 96.86 & 97.16  & 97.87  \\
    8     & 75.99  & 82.22  & 80.59  & 85.64  & 90.95  & 80.54  & 85.65  & 90.95 & 90.17 & 91.71  & 92.55  \\
    9     & 96.76  & 94.76  & 94.99  & 96.03  & 95.99  & 95.01  & 96.05  & 96.01 & 96.36 & 97.44  & 97.72  \\
    10    & 74.97  & 74.49  & 89.85  & 89.02  & 93.51  & 90.09  & 89.18  & 93.61 & 94.35 & 96.24  & 97.39  \\
    11    & 73.20  & 91.95  & 84.13  & 99.05  & 96.55  & 84.42  & 99.08  & 96.61 & 96.35 & 97.48  & 97.53  \\
    12    & 97.25  & 98.92  & 99.69  & 99.91  & 99.79  & 99.71  & 99.91  & 99.80 & 96.76 & 99.85  & 97.76  \\
    13    & 97.16  & 98.85  & 99.68  & 99.75  & 99.56  & 99.71  & 99.76  & 99.61 & 97.15 & 99.91  & 99.12  \\
    14    & 91.60  & 94.26  & 94.74  & 99.19  & 97.70  & 95.03  & 99.21  & 97.81 & 96.12 & 98.79  & 98.51  \\
    15    & 67.87  & 75.37  & 79.08  & 82.16  & 91.07  & 79.13  & 82.21  & 91.08 & 88.70 & 91.18  & 89.99  \\
    16    & 93.92  & 94.59  & 98.40  & 93.94  & 92.84  & 98.47  & 93.97  & 92.86 & 96.83 & 99.61  & 98.62  \\
    Novel & 0.00  & 0.00  & 0.00  & 0.00  & 0.00  & 2.84  & 1.95  & 1.57  & 75.26 & 76.87 & 83.32  \\
    \midrule
    \midrule
    OA (\%)   & 80.85$\pm$0.50  & 83.87$\pm$0.30   & 84.08$\pm$2.19   & 85.72$\pm$1.60   & 88.14$\pm$1.90   & 84.16$\pm$2.16   & 85.81$\pm$1.60   & 88.22$\pm$1.90 & 91.82$\pm$0.52 & 93.73$\pm$0.80   & \textbf{94.34$\pm$0.61}   \\
    F1$\times$100 & 89.41$\pm$0.30   & 91.23$\pm$0.17   & 91.34$\pm$1.32   & 92.30$\pm$0.95   & 93.70$\pm$1.50   & 91.37$\pm$1.31   & 92.35$\pm$0.95   & 93.73$\pm$1.51 & 95.29$\pm$0.31 & 96.53$\pm$0.60   & \textbf{96.82$\pm$0.71}   \\
    Error (\%) & 15.16$\pm$2.08   & 11.53$\pm$0.90   & 17.31$\pm$7.29   & 14.30$\pm$5.24   & 10.84$\pm$4.50   & 17.12$\pm$7.30   & 14.21$\pm$5.24   & 10.76$\pm$4.50 & 6.98$\pm$0.83 & 4.63$\pm$2.79   & \textbf{3.96$\pm$2.61}   \\
    \bottomrule
    \bottomrule
    \end{tabular}}%
  \label{tab:oa2_salinas}%
\end{table*}%

\subsubsection{Results on the Indian Pines dataset}
\begin{figure*}[!t]
\centering
\subfigure[Close]{\includegraphics[width=0.13\textwidth]{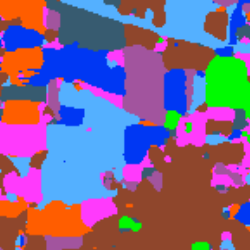}}
\subfigure[CROSR]{\includegraphics[width=0.13\textwidth]{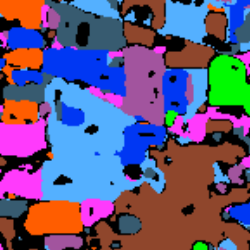}}
\subfigure[MDL4OW]{\includegraphics[width=0.13\textwidth]{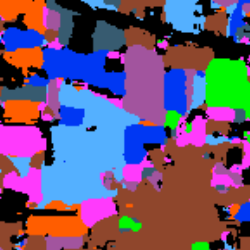}}
\subfigure[MDL4OW/C]{\includegraphics[width=0.13\textwidth]{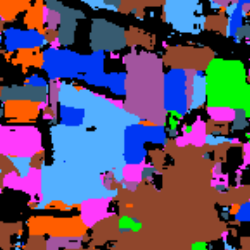}}
\subfigure[Image]{\includegraphics[width=0.13\textwidth]{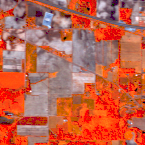}}
\subfigure[Reference]{\includegraphics[width=0.13\textwidth]{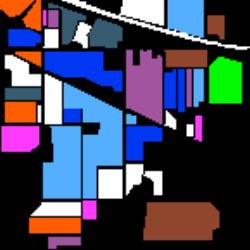}}
\subfigure{\includegraphics[width=0.13\textwidth]{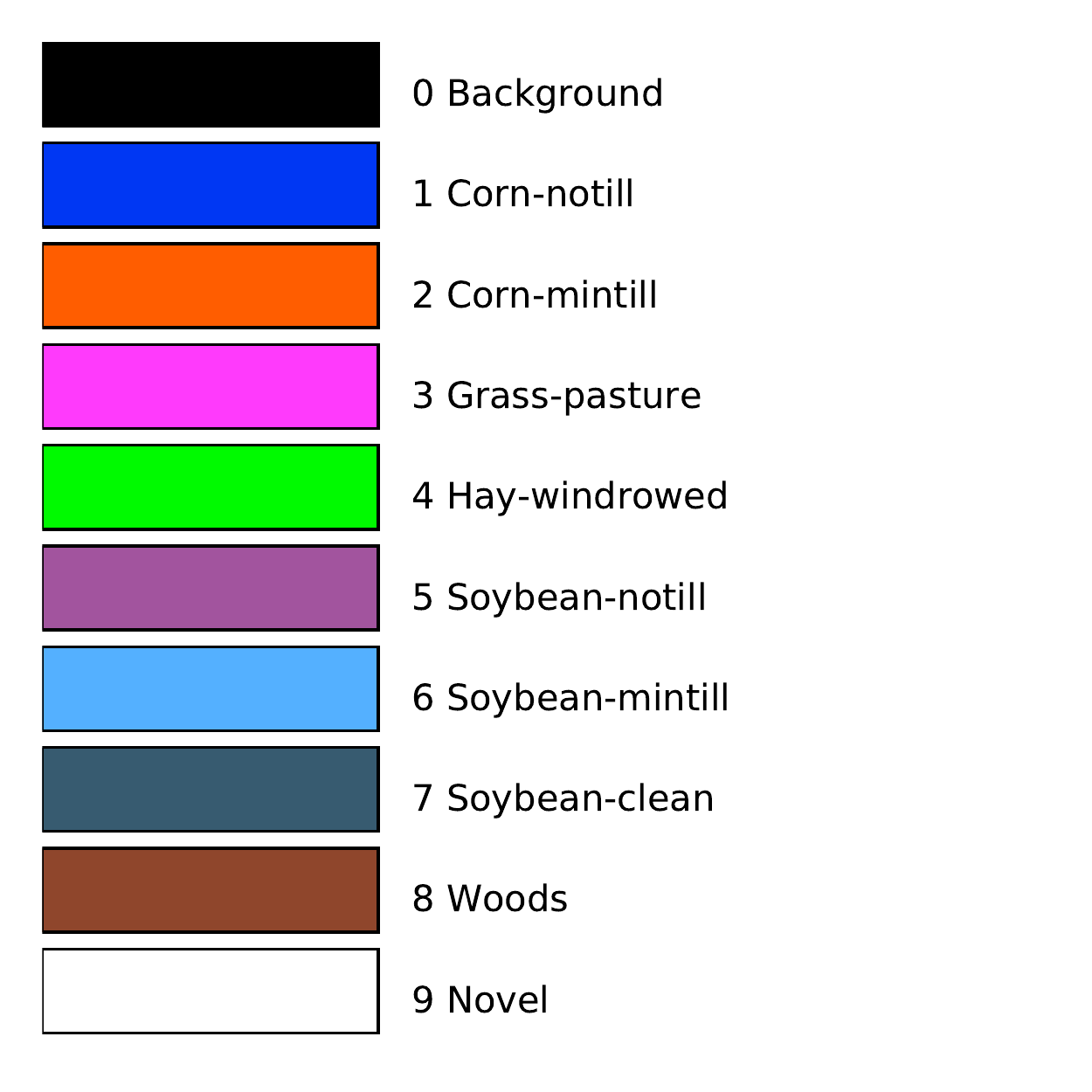}}
\caption{Classification maps of the Indian Pines. Known classes: 1-8; unknown classes: 9. In the classification maps, those masked with black color are identified as unknown by the proposed method.}
\label{fig:indian_maps}
\end{figure*}

To show the generalization of the proposed method, another popular dataset, Indian Pines, is tested. This dataset, as shown in the false color map in Figure \ref{fig:indian_maps}, mainly covers an agricultural region. In the standard 16-class reference map,  one non-agricultural class exists. However, the road on the upper side of the image is not represented in the reference data.  It should not be considered as an agricultural class.  As mentioned before, since some of the land covers in this dataset consist of very few samples, we only included eight classes as the training set following \cite{lee2017going}. The remaining 8+1 classes were naturally considered as unknown. The results are reported in Tables \ref{tab:oa1_indian} and \ref{tab:oa2_indian}.  

Under the few-shot context, the three open methods are competitive and resulted in a mapping error of 28\%--29\%, 3--4\% lower compared to the closed error of 33.27\%, whereas the proposed MDL4OW  is slightly better. As for the many-shot setting, the class-wise MDL4OW/C obtained the best classification result in three evaluation metrics. The OA and F1 are 82.61\% and 0.9001, 2.47\% and 0.0104 higher than the closed classification. The mapping error is more sensitive, and is 16.00\%, 7.60\% lower than the baseline.  

We show the classification maps obtained using 200 samples per class in Figure \ref{fig:indian_maps}. The road on the upper side of the image is successfully identified as unknown by the proposed method, both MDL4OW and MDL4OW/C. The close CNN classified it as one of the agricultural classes, soybean-mintill, leading to its overestimation. CROSR failed to recognize the road as unknown. From the classification maps, we can see that most man-made materials unrelated to crops can be identified as unknown by the proposed method, showing its promise in precise crop mapping.

\begin{table*}[!t]
  \centering
  \caption{Classification accuracy using 20 samples per class on the Indian Pines dataset. * top-1 softmax probability $<$ 0.5 is considered as unknown. Error (\%): the smaller the better; else: the larger the better.}
  \scalebox{0.7}[0.7]{
    \begin{tabular}{c|ccccc|cccccc}
    \toprule
    \toprule
          & \multicolumn{5}{c|}{Close}            & \multicolumn{6}{c}{Open} \\
          & RF    & SVM   & DCCNN\cite{lee2017going} & WCRN\cite{liu2018wide}  & HResNet\cite{liu2020multitask} & DCCNN* & WCRN*  & HResNet* & CROSR\cite{yoshihashi2019classification} & MDL4OW  & MDL4OW/C \\
    \midrule
    \midrule
    1     & 47.95  & 48.43  & 50.24  & 62.79  & 73.48  & 50.23  & 62.92  & 73.54 & 71.58 & 74.89  & 74.51  \\
    2     & 46.15  & 41.52  & 39.08  & 51.76  & 69.12  & 39.42  & 52.22  & 69.35 & 68.63 & 68.22  & 68.45  \\
    3     & 35.75  & 44.91  & 62.45  & 62.52  & 67.03  & 62.70  & 63.48  & 67.55 & 70.05 & 65.12  & 66.34  \\
    4     & 82.43  & 89.64  & 90.58  & 90.24  & 89.57  & 90.90  & 90.88  & 89.79 & 88.04 & 89.61  & 89.24  \\
    5     & 53.92  & 54.54  & 46.07  & 60.47  & 76.42  & 45.93  & 61.08  & 76.58 & 75.03 & 76.79  & 78.21  \\
    6     & 54.45  & 54.87  & 56.35  & 64.98  & 77.06  & 56.28  & 65.06  & 77.01 & 74.43 & 77.23  & 76.55  \\
    7     & 37.06  & 45.99  & 43.56  & 48.42  & 65.57  & 43.79  & 48.92  & 66.10 & 66.99 & 65.50  & 66.01  \\
    8     & 87.23  & 83.34  & 75.06  & 84.99  & 84.49  & 75.40  & 85.52  & 84.64 & 87.85 & 84.34  & 83.85  \\
    Novel & 0.00  & 0.00  & 0.00  & 0.00  & 0.00  & 6.98  & 9.70  & 3.11  & 36.16 & 21.18  & 27.93  \\
    \midrule
    \midrule
    OA (\%)    & 49.27$\pm$1.48  & 51.01$\pm$1.90   & 50.99$\pm$2.98   & 58.67$\pm$2.10   & 67.98 $\pm$2.58  & 51.35$\pm$2.87   & 59.35$\pm$2.35   & 68.14$\pm$2.57 & 68.49$\pm$3.94  & 68.92$\pm$3.26   & \textbf{69.00$\pm$3.90}   \\
    F1$\times$100 & 66.00$\pm$1.33   & 67.53$\pm$1.68   & 67.49$\pm$2.61   & 73.93$\pm$1.68   & 80.91$\pm$1.87   & 67.43$\pm$2.67   & 74.12$\pm$1.81   & 80.95$\pm$1.87 & 79.83$\pm$2.40 & \textbf{80.99$\pm$2.19}   & 80.73$\pm$2.42   \\
    Error (\%) & 46.48$\pm$8.97   & 41.74$\pm$9.19   & 48.43$\pm$9.27   & 38.26$\pm$8.64   & 33.27$\pm$10.10   & 47.11$\pm$11.16   & 36.71$\pm$9.62   & 32.62$\pm$9.93 & 28.90$\pm$11.85 & \textbf{28.32$\pm$10.74}   & 28.37$\pm$11.30   \\
    \bottomrule
    \bottomrule
    \end{tabular}}%
  \label{tab:oa1_indian}%
\end{table*}%

\begin{table*}[!t]
  \centering
  \caption{Classification accuracy using 200 samples per class on the Indian Pines dataset. * top-1 softmax probability $<$ 0.5 is considered as unknown. Error (\%): the smaller the better; else: the larger the better.}
  \scalebox{0.7}[0.7]{
    \begin{tabular}{c|ccccc|cccccc}
    \toprule
    \toprule
          & \multicolumn{5}{c|}{Close}            & \multicolumn{6}{c}{Open} \\
          & RF    & SVM   & DCCNN\cite{lee2017going} & WCRN\cite{liu2018wide}  & HResNet\cite{liu2020multitask} & DCCNN* & WCRN*  & HResNet* & CROSR\cite{yoshihashi2019classification} & MDL4OW  & MDL4OW/C \\
    \midrule
    \midrule
    1     & 76.23  & 78.72  & 77.63  & 94.14  & 94.97  & 77.72  & 94.22  & 95.06 & 94.34 & 95.31  & 95.03  \\
    2     & 72.88  & 72.83  & 73.07  & 84.24  & 82.66  & 72.52  & 84.40  & 82.77 & 82.89 & 83.26  & 83.47  \\
    3     & 54.32  & 54.74  & 64.01  & 71.76  & 85.79  & 64.51  & 72.31  & 86.31 & 86.05 & 85.33  & 85.66  \\
    4     & 91.77  & 93.89  & 95.12  & 95.01  & 95.17  & 95.15  & 95.05  & 95.23 & 95.91 & 94.88  & 95.31  \\
    5     & 78.01  & 77.04  & 81.76  & 91.28  & 93.79  & 82.16  & 91.65  & 94.08 & 93.97 & 94.08  & 94.52  \\
    6     & 77.84  & 78.80  & 82.44  & 94.80  & 87.64  & 81.57  & 94.90  & 87.80 & 87.12 & 88.82  & 88.67  \\
    7     & 55.72  & 85.09  & 81.23  & 85.49  & 82.91  & 81.86  & 85.95  & 83.44 & 88.76 & 84.98  & 88.35  \\
    8     & 79.32  & 86.43  & 75.90  & 75.73  & 86.91  & 76.03  & 76.04  & 87.16 & 88.91 & 87.58  & 87.70  \\
    Novel & 0.00  & 0.00  & 0.00  & 0.00  & 0.00  & 3.86  & 4.76  & 4.50  & 22.09 & 32.21  & 33.58  \\
    \midrule
    \midrule
    OA (\%)    & 66.73$\pm$0.58  & 70.53$\pm$0.53  & 70.97$\pm$6.56  & 79.13$\pm$0.54  & 80.14$\pm$0.40  & 70.41$\pm$7.29  & 79.58$\pm$0.56  & 80.57$\pm$0.46 & 81.55$\pm$1.53 & 82.23$\pm$1.64  & \textbf{82.61$\pm$1.91}  \\
    F1$\times$100 & 80.04$\pm$0.04  & 82.72$\pm$0.36  & 82.84$\pm$4.62  & 88.35$\pm$0.34  & 88.97$\pm$0.24  & 82.32$\pm$5.24  & 88.57$\pm$0.35  & 89.19$\pm$0.27 & 89.54$\pm$0.79 & 89.79$\pm$0.89  & \textbf{90.01$\pm$1.02}  \\
    Error (\%) & 34.95$\pm$1.70  & 28.13$\pm$4.02  & 33.75$\pm$8.95  & 24.54$\pm$1.81  & 23.60$\pm$0.23  & 33.22$\pm$9.02  & 23.95$\pm$1.81  & 23.00$\pm$0.31 & 18.76$\pm$2.17 & 16.26$\pm$1.49  & \textbf{16.00$\pm$2.36}  \\
    \bottomrule
    \bottomrule
    \end{tabular}}%
  \label{tab:oa2_indian}%
\end{table*}%

\subsubsection{Threshold analysis}
\begin{figure*}[!t]
\color{black}
\centering
\subfigure[Pavia]{\includegraphics[width=0.3\textwidth]{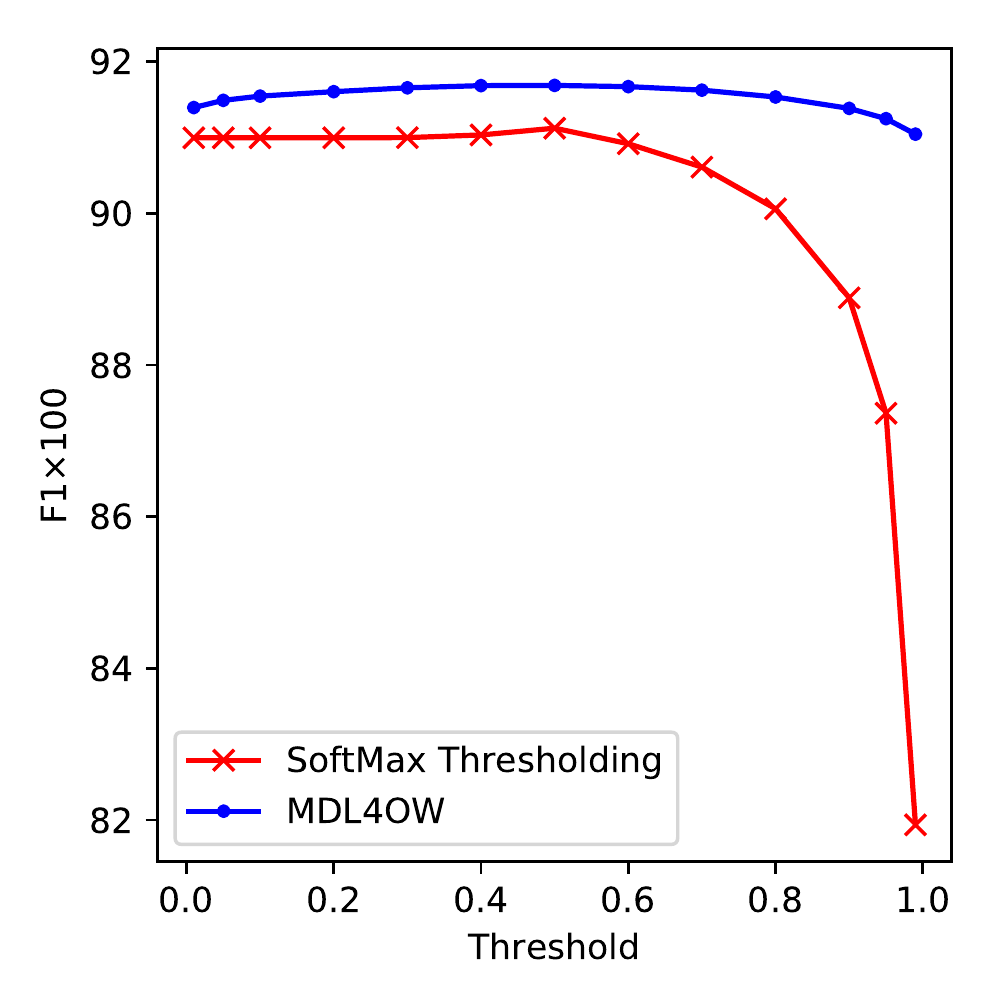}}
\subfigure[Salinas]{\includegraphics[width=0.3\textwidth]{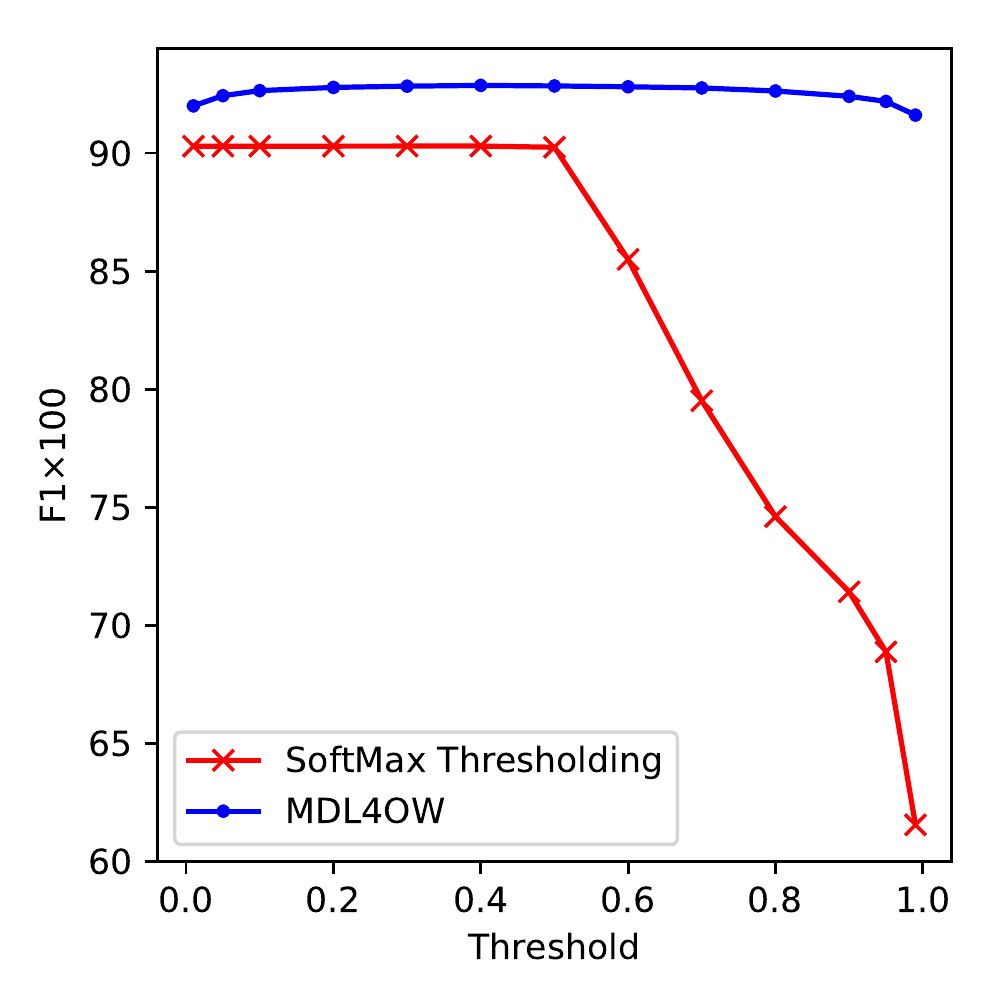}}
\subfigure[Indian]{\includegraphics[width=0.3\textwidth]{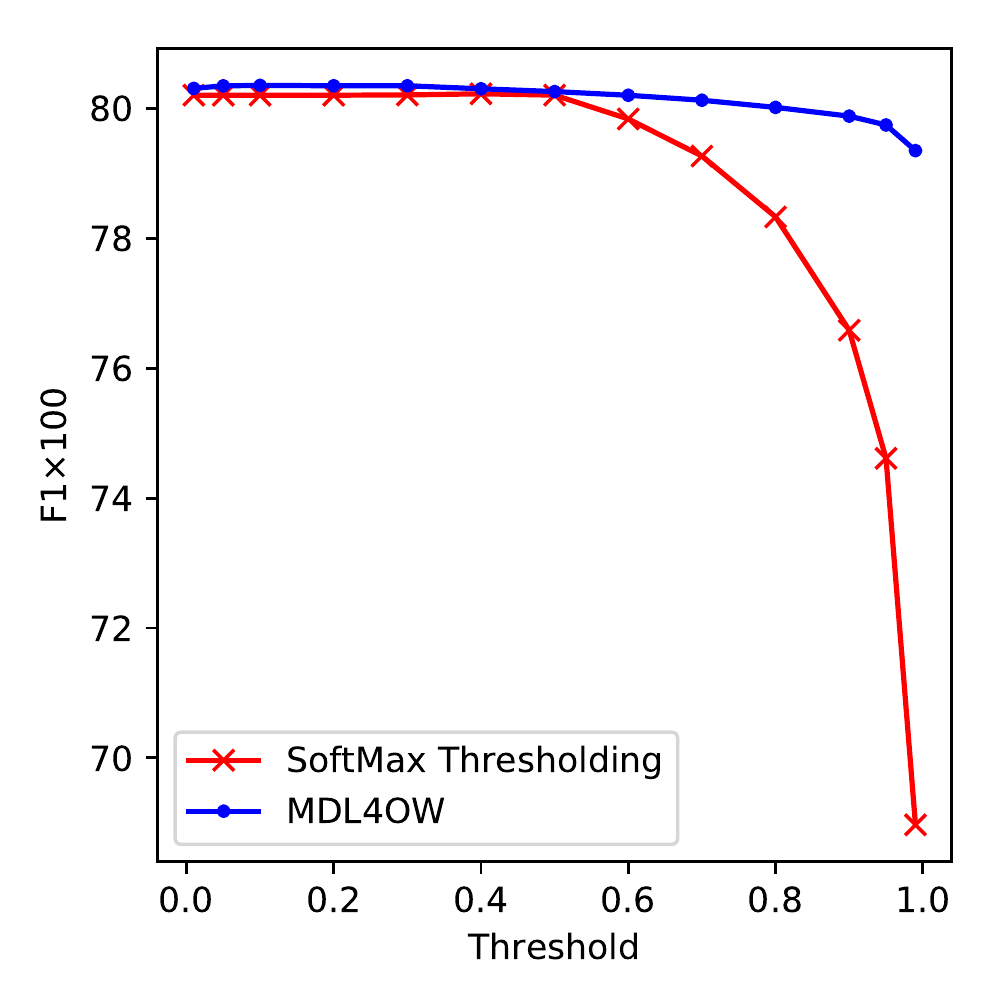}}
\caption{Confidence threshold analysis in terms of F1 score under the few-shot setting. }
\label{fig:threshold}
\end{figure*}
We show a threshold analysis under the few-shot context in Figure \ref{fig:threshold} to demonstrate that  a) SoftMax confidence thresholding is not enough for detecting the unknown, and b) MDL4OW is robust to the threshold setting ($z$).  From Figure \ref{fig:threshold} we can see that with a larger threshold, the F1 score first slightly improves and then drops sharply. A low SoftMax probability indicates the classifier is uncertain with its decision, but uncertainty is not equal to unknown. A sample instance that is difficult to classify may be the valuable instance close to the decision boundary between two classes (the valuable sample in active learning) \cite{scheirer2012toward}.

\subsection{Comparison with other methods}
\begin{table}[!t]
\color{black}
  \centering
  \caption{Comparison with other methods in terms of closed OA (\%) and open OA (\%). * Top-1 SoftMax probability $<$ 0.5 is considered as unknown.}
  \scalebox{0.75}[0.75]{
    \begin{tabular}{|c|c|cc|cc|}
    \hline
    \multicolumn{2}{|c|}{\multirow{2}[2]{*}{Method}} & \multicolumn{2}{c|}{Pavia University} & \multicolumn{2}{c|}{Salinas Valley} \\
    \multicolumn{2}{|c|}{} & Closed OA & Open OA & Closed OA & Open OA \\
    \hline
    \multirow{2}[2]{*}{Non-deep} & RF    & 81.56$\pm$0.99 & 72.77$\pm$0.88 & 89.24$\pm$0.55 & 80.85$\pm$0.50 \\
          & SVM   & 91.18$\pm$0.48 & 81.36$\pm$0.43 & 92.60$\pm$0.34 & 83.87$\pm$0.30 \\
    \hline
    \multirow{8}[2]{*}{Close} & DCCNN \cite{lee2017going} & 96.99$\pm$0.56 & 86.55$\pm$0.50 & 92.84$\pm$2.41 & 84.08$\pm$2.19 \\
          & WCRN \cite{liu2018wide} & 97.94$\pm$0.20 & 87.39$\pm$0.18 & 94.66$\pm$1.76 & 85.72$\pm$1.60 \\
          & HResNet & 99.36$\pm$0.15 & 88.66$\pm$0.12 & 97.39$\pm$0.64 & 88.14$\pm$1.90 \\
          & CNN \cite{hu2015deep} & 87.90  & 78.43 & 90.25 & 81.77 \\
          & R-PCA-CNN \cite{makantasis2015deep} & 94.38 & 84.62 & 92.39 & 83.71 \\
          & CNN-PPF \cite{li2016hyperspectral} & 96.48 & 86.09 & 94.80  & 85.89 \\
          & C-CNN\cite{mei2017learning} & 98.41 & 87.81 & 97.42 & 88.27 \\
          & SSRN\cite{zhong2017spectral} & 99.79 & 89.04 & 98.68 & 89.41 \\
   \hline
    \multirow{3}[2]{*}{Few-shot} & DFSL+NN\cite{liu2018deep} & 97.85 & 87.31 & 97.78 & 88.59 \\
          & DFSL+SVM\cite{liu2018deep} & 98.62 & 88.00    & 97.81 & 88.62 \\
          & SS-RN\cite{rao2019spatial} & 99.69 & 88.95 & 98.62 & 89.35 \\
  \hline
    \multirow{6}[2]{*}{Open} & DCCNN* & 96.92$\pm$0.59 & 86.58$\pm$0.54 & 92.77$\pm$2.41 & 84.16$\pm$2.16 \\
          & WCRN* & 97.93$\pm$0.20 & 87.41$\pm$0.17 & 94.65$\pm$1.76 & 85.81$\pm$1.60 \\
          & HResNet* & 99.35$\pm$0.14 & 88.76$\pm$0.13 & 97.38$\pm$0.64 & 88.22$\pm$1.90 \\
          & CROSR\cite{yoshihashi2019classification} & 96.67$\pm$0.25 & 90.39$\pm$0.67 & 94.75$\pm$0.53 & 91.82$\pm$0.52 \\
          & MDL4OW & 97.93$\pm$0.26 & 90.17$\pm$0.13 & 96.37$\pm$0.80 & 93.73$\pm$0.80 \\
          & MDL4OW/C & 97.23$\pm$0.37 & 90.53$\pm$0.62 & 94.99$\pm$0.63 & 94.34$\pm$0.61 \\
    \hline
    \end{tabular}}%
  \label{tab:compareOthers}%
\end{table}%

In Table \ref{tab:compareOthers}, we compare the proposed methods with other competitors in terms of closed OA and open OA using 200 samples per class. The comparison is based on the Pavia and Salinas data. We do not compare the results on the Indian data since the used known classes are different with other studies. Closed OA is calculated based on known classes without considering the existence of unknown class. Open OA includes the unknown class. It is easy to convert a closed OA to open OA once the unknown instances are determined from a reference map. 
Besides two non-deep methods (RF and SVM), we also include eight deep learning methods and three deep learning-based few-shot methods. They are DCCNN \cite{lee2017going}, WCRN \cite{liu2018wide}, HResNet (baseline of the proposed method), CNN \cite{hu2015deep}, R-PCA-CNN \cite{makantasis2015deep}, CNN-PPF \cite{li2016hyperspectral}, C-CNN \cite{mei2017learning}, SSRN \cite{zhong2017spectral}, DFSL+NN \cite{liu2018deep}, DFSL+SVM \cite{liu2018deep}, and SS-RN \cite{rao2019spatial}.

MDL4OW/C achieves the best open OA, followed by MDL4OW. Those closed methods lack the ability to tell the unknown from known classes and therefore achieve a low open OA. Take the SSRN  \cite{zhong2017spectral} as an example. It achieves the highest closed OA on the Salinas data (98.68\%), but only obtains an open OA of 89.41\%. MDL4OW/C  obtains the highest open OA (94.34\%), whereas MDL4OW obtains the second highest open OA (93.73\%) and a reasonable closed OA (96.37\%). It should note that although open methods are promising in identifying the novel class, they inevitably lead to either a large or a slightly decrease in the closed OA, since the closed methods do not bother with the novel class. The trade-off here is similar to target detection \cite{lu2017hybrid}. As the novel class is inevitable in the real world, a method that is capable of identifying the unknown and maintaining good performance on the known classes is essential.  The proposed MDL4OW achieves a good trade-off here if we look at both the closed OA and open OA.

\subsection{With imperfect classification systems}
\begin{table*}[!t]
  \centering
  \caption{Selected imperfect classification systems.  The number is the class index. * incomplete classification system.}
  \scalebox{0.8}[0.8]{
    \begin{tabular}{ccccccc}
    \toprule
    \toprule
    Dataset & \#Train & \#Test & Openness (\%) & OW1   & OW2   & OW3 \\
    \midrule
    \midrule
    Pavia & 3     & 15    & 42.3  & \makecell{[1, 4, 6] \\ Asphalt, Trees, Bare S.}  & \makecell{[2, 4, 6]* \\ Meadows, Trees, Bare S.} & \makecell{[1, 7, 8]* \\ Asphalt, Bitumen, Brick} \\
    Salinas & 6     & 18    & 29.3  & \makecell{[1, 2, 11, 12, 13, 14]* \\ Weed-1, Weed-2, Lett.-4wk \\ Lett.-5wk, Lett.-6wk, Lett.-7wk} & \makecell{[1, 3, 6, 9, 11, 15] \\ Weed-1, Fallow, Stubble \\ Soil, Lett.-4wk, Viny.-U} & \makecell{[2, 5, 9, 10, 13, 15] \\ Weed-2, Fallow-S, Soil \\ Corn, Lett.-6wk, Viny.-U} \\
    Indian & 3     & 17    & 45.2  & \makecell{[1, 3, 5] \\ Corn-n., Grass, Soybean-n.} & \makecell{[5, 6, 7]* \\ Soybean-n., Soybean-m., Soybean-c.} & \makecell{[1, 7, 8] \\ Corn-n., Soybean-c., Woods} \\
    \bottomrule
    \bottomrule
    \end{tabular}}%
  \label{tab:open_select}%
\end{table*}%

\begin{figure*}[!t]
\centering
\subfigure[Pavia, OW1]{\includegraphics[width=0.3\textwidth]{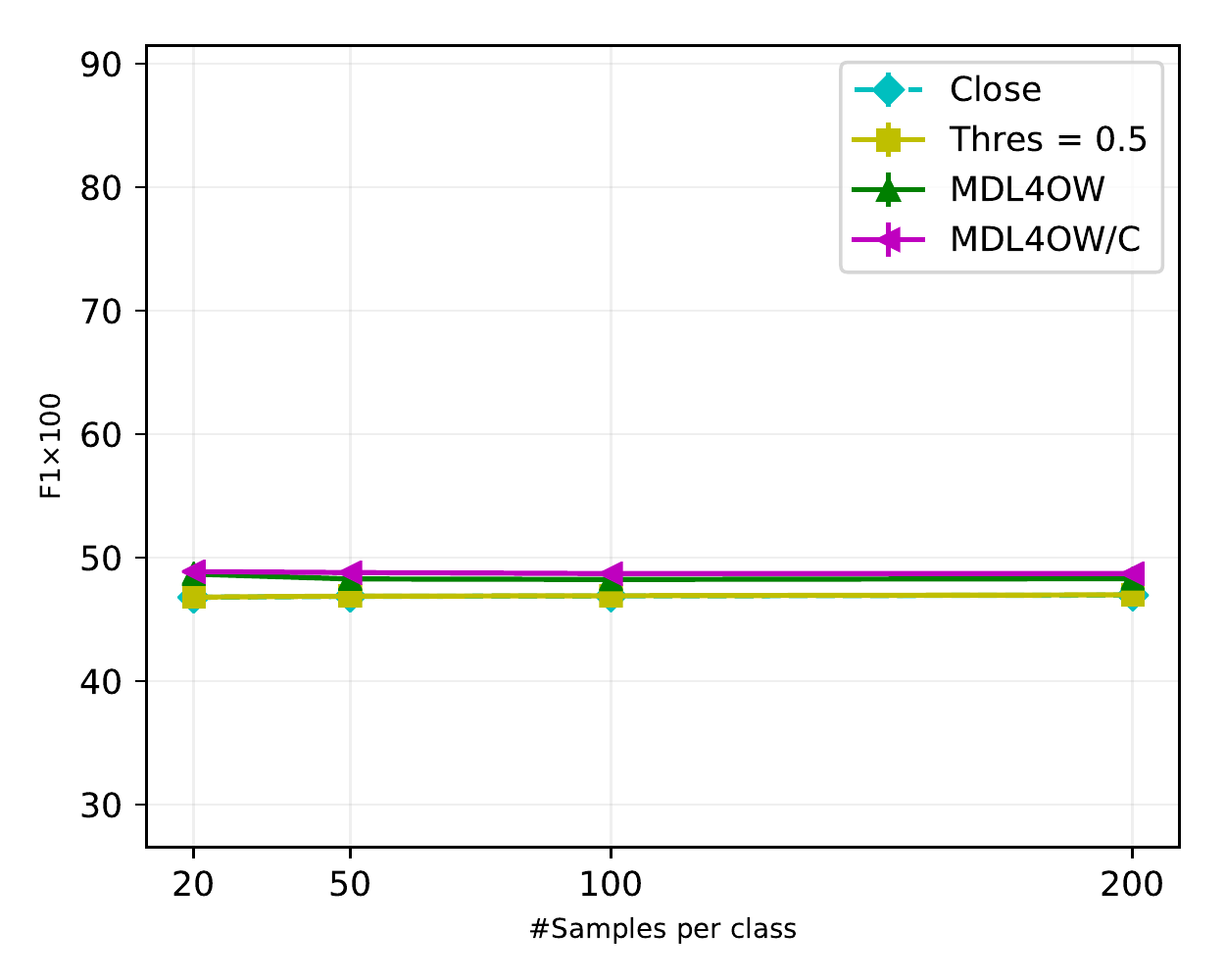}}
\subfigure[Pavia, OW2*]{\includegraphics[width=0.3\textwidth]{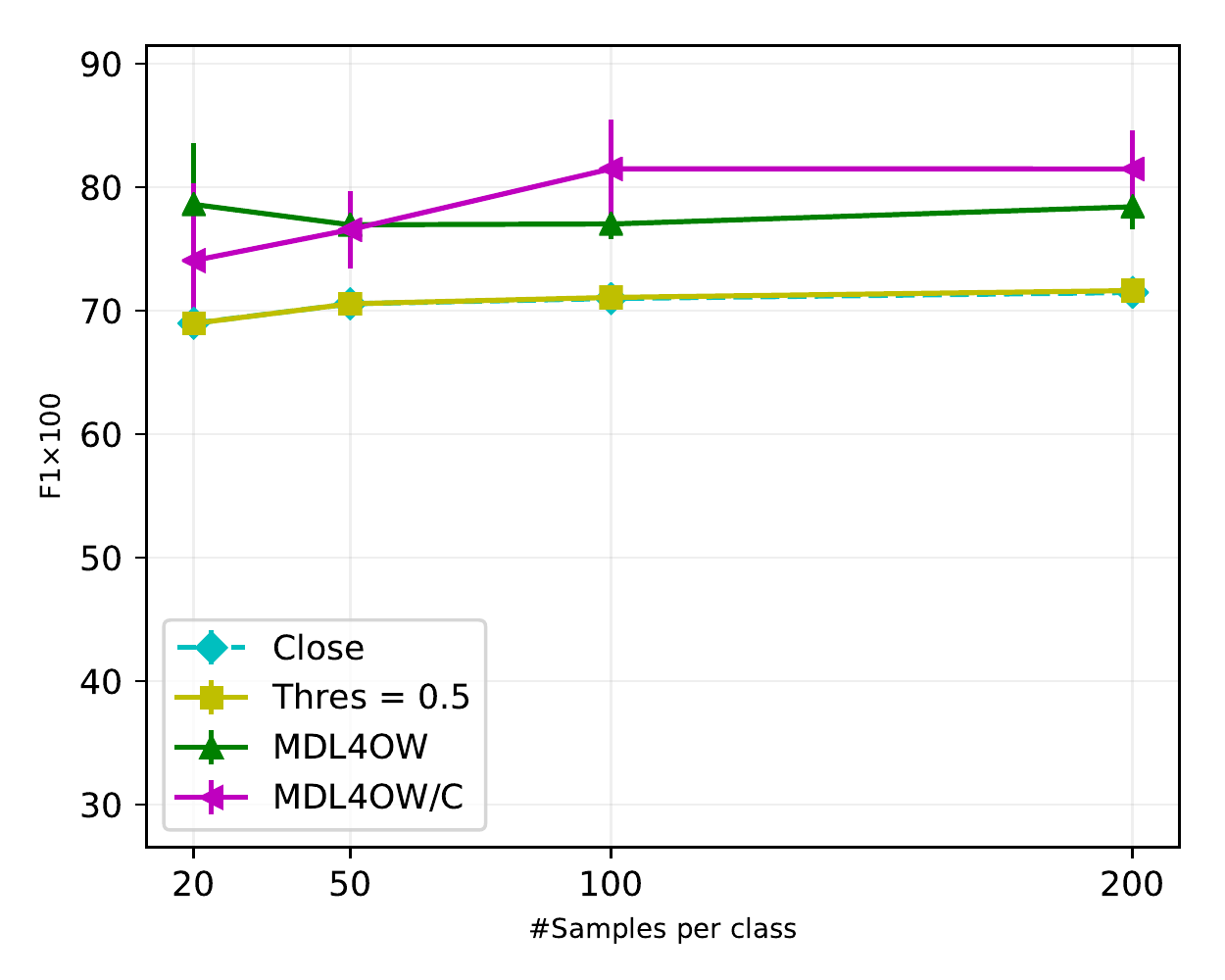}} 
\subfigure[Pavia, OW3*]{\includegraphics[width=0.3\textwidth]{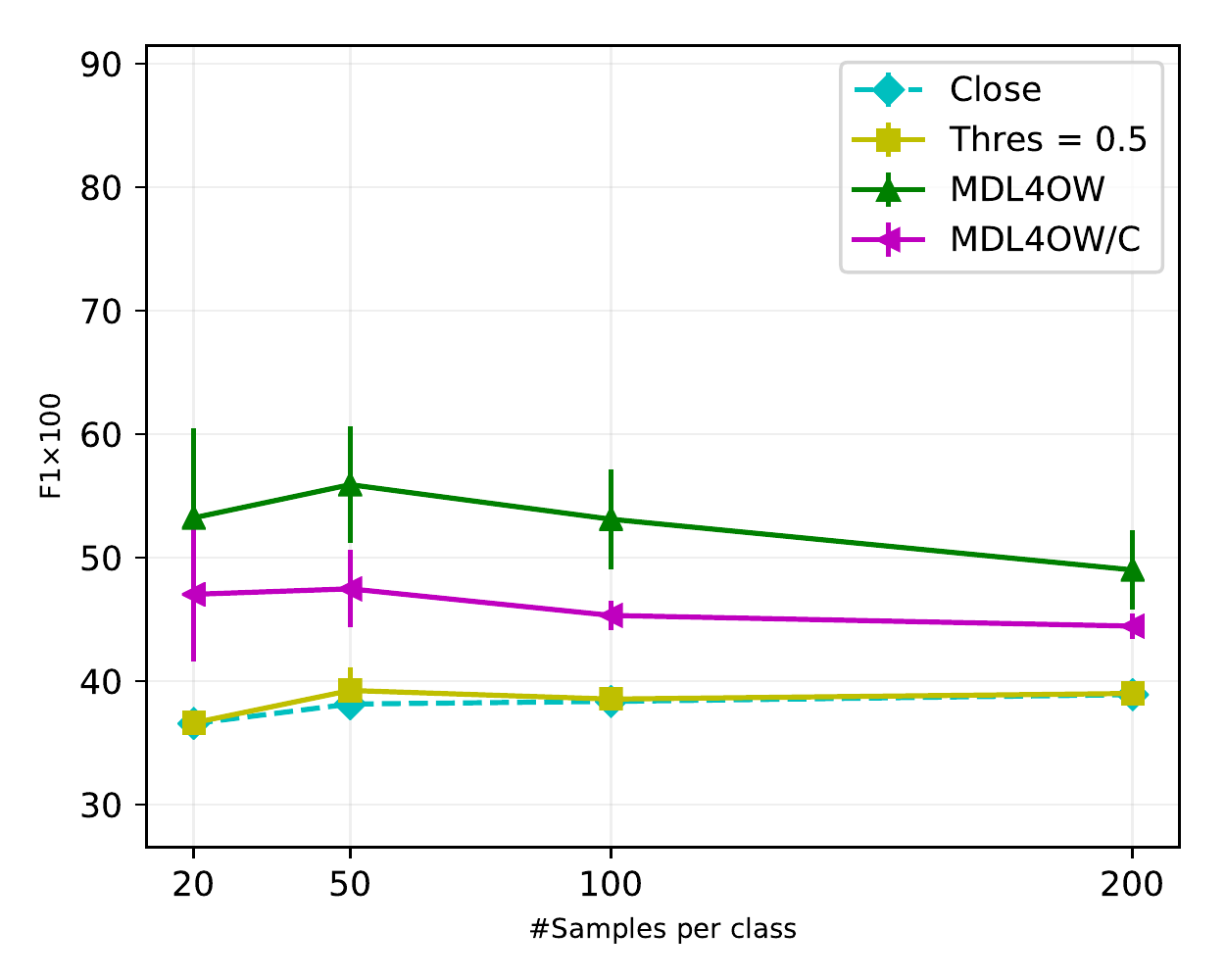}}
\vfill
\subfigure[Salinas, OW1*]{\includegraphics[width=0.3\textwidth]{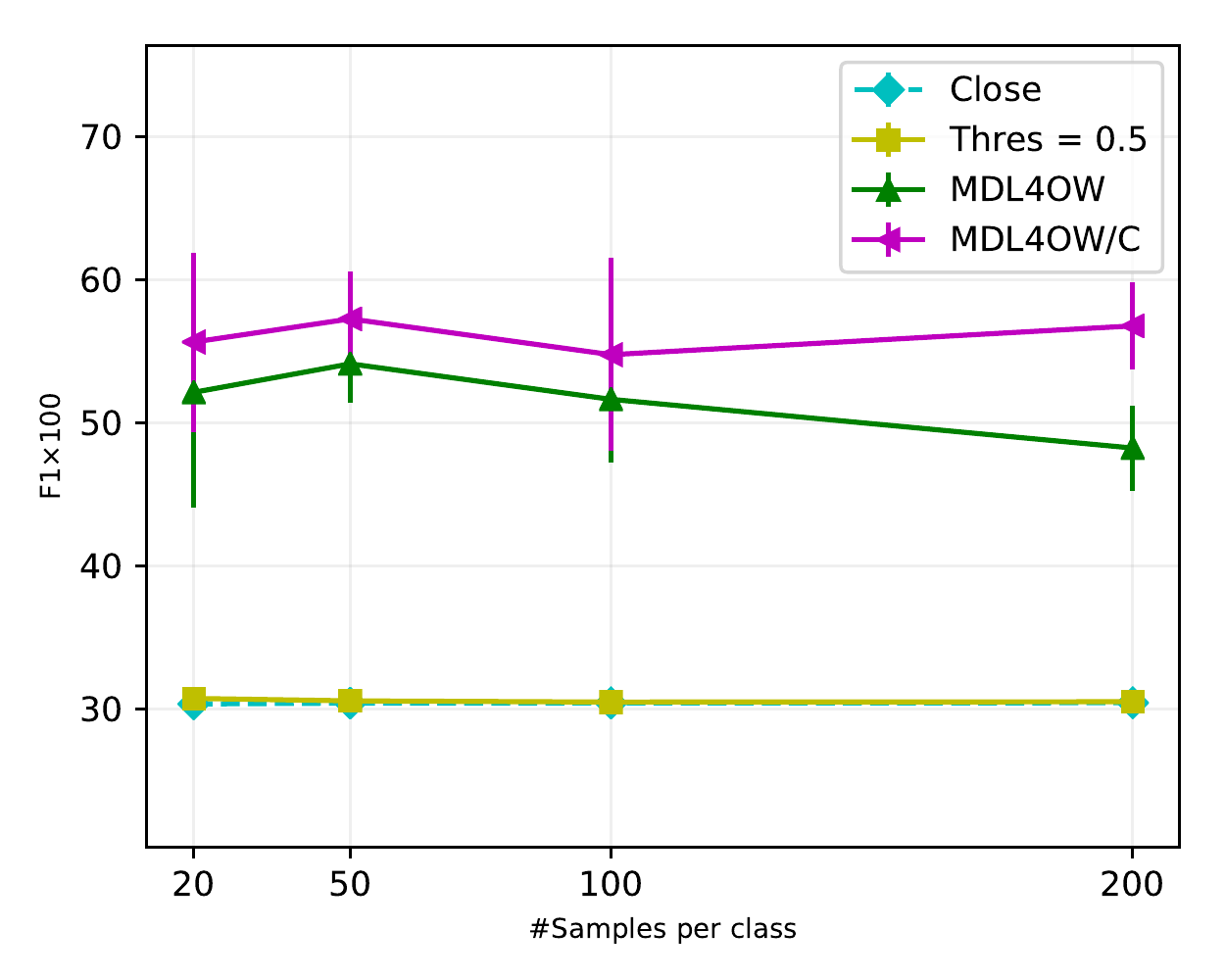}} 
\subfigure[Salinas, OW2]{\includegraphics[width=0.3\textwidth]{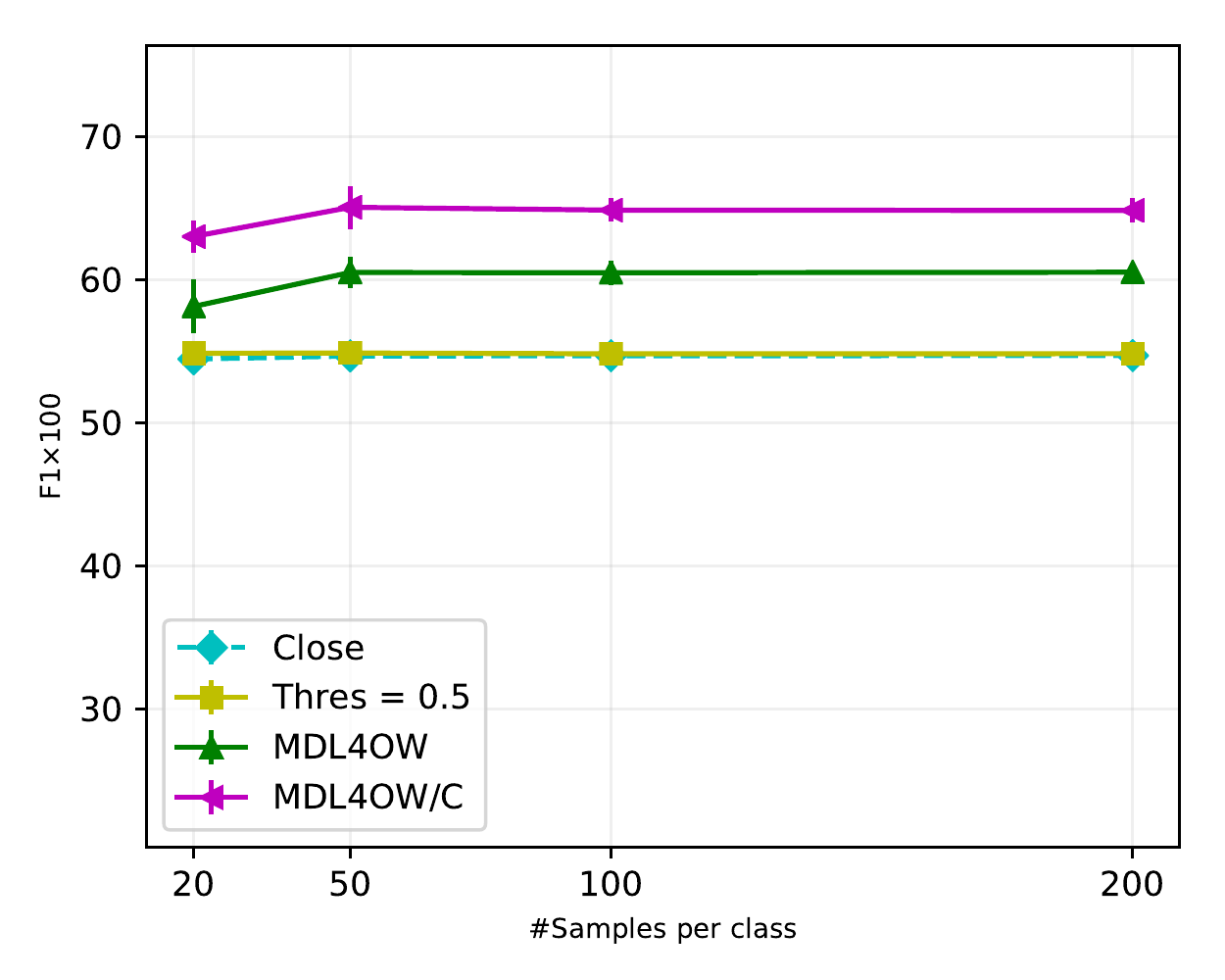}} 
\subfigure[Salinas, OW3]{\includegraphics[width=0.3\textwidth]{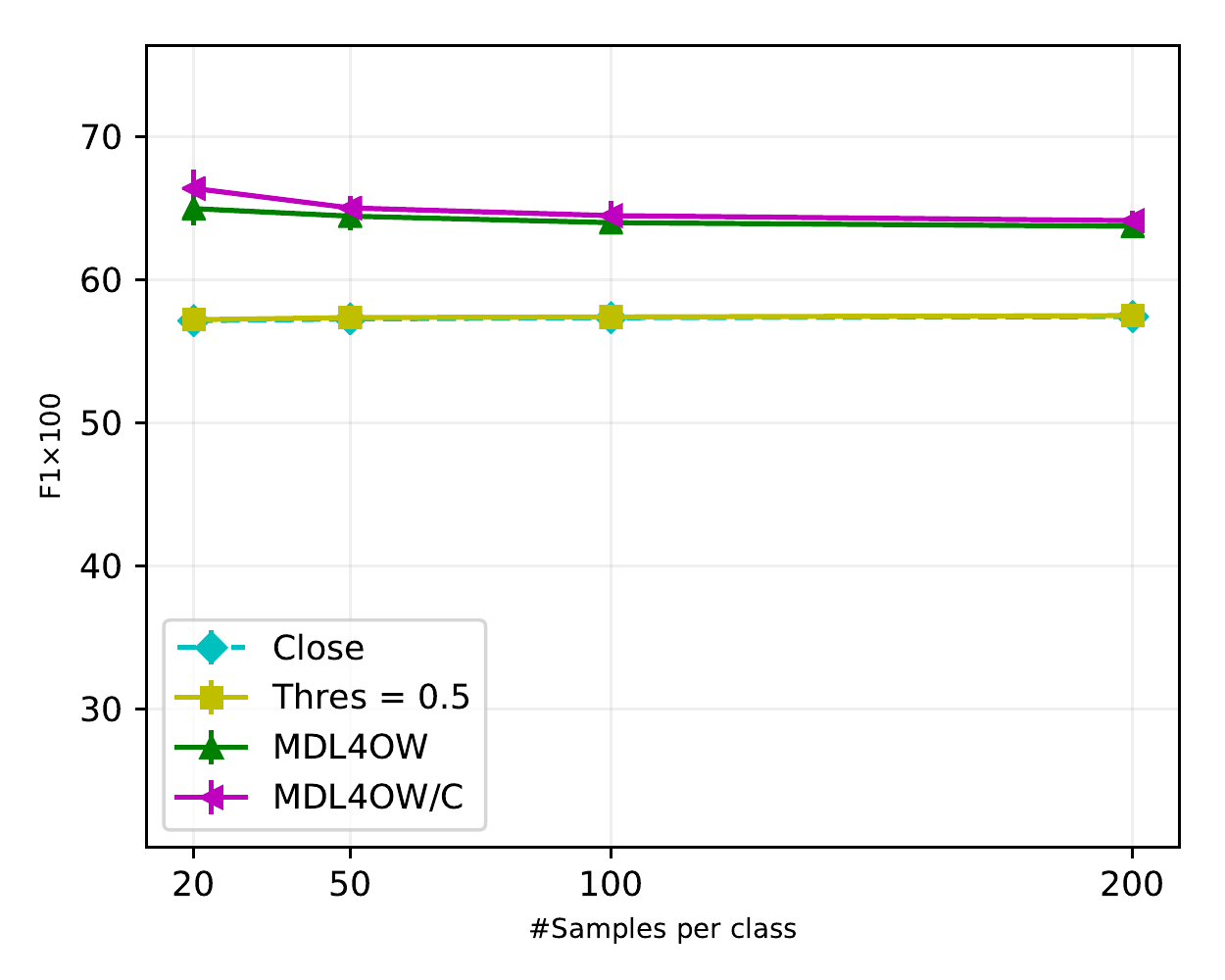}} 
\vfill
\subfigure[Indian, OW1]{\includegraphics[width=0.3\textwidth]{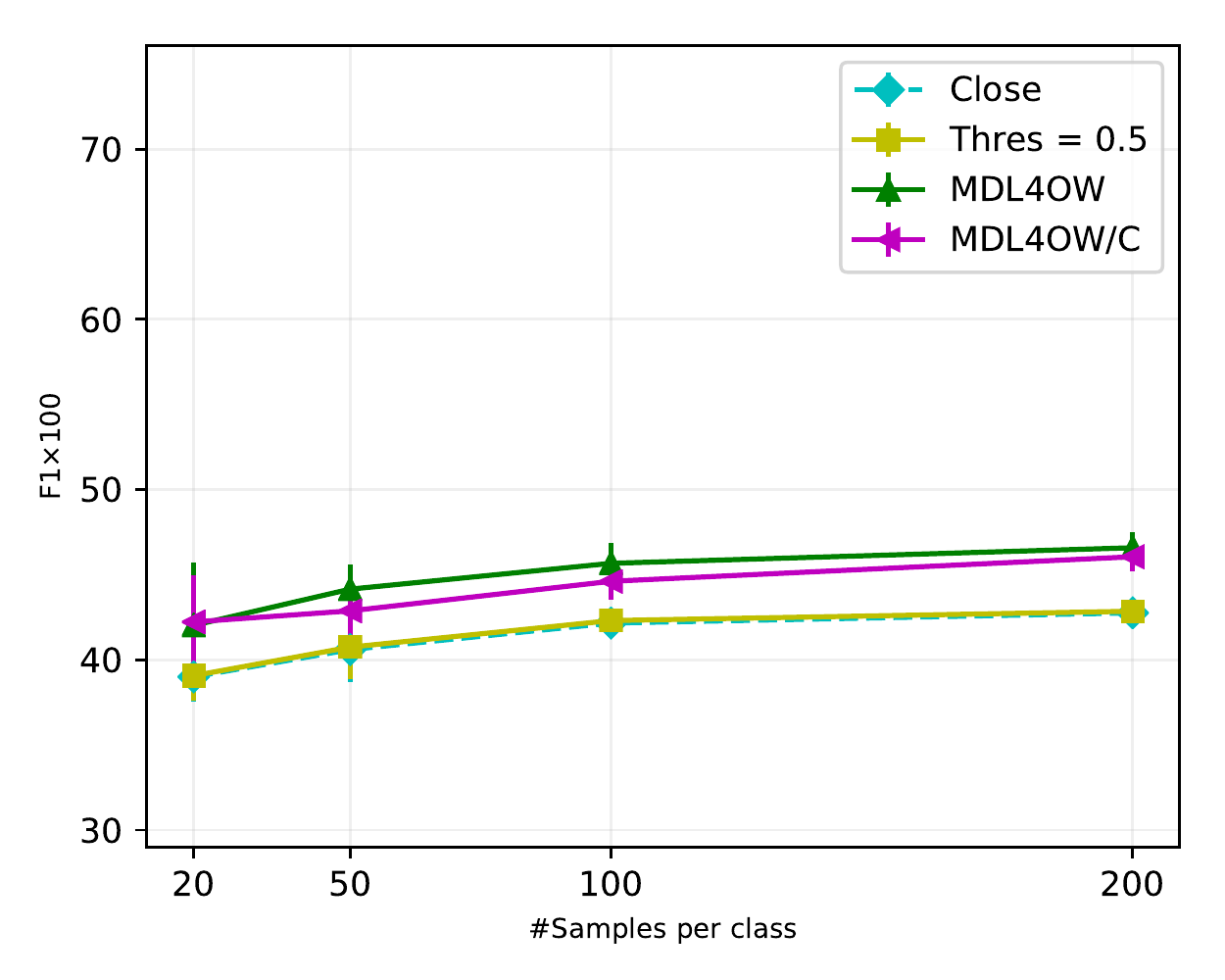}} 
\subfigure[Indian, OW2*]{\includegraphics[width=0.3\textwidth]{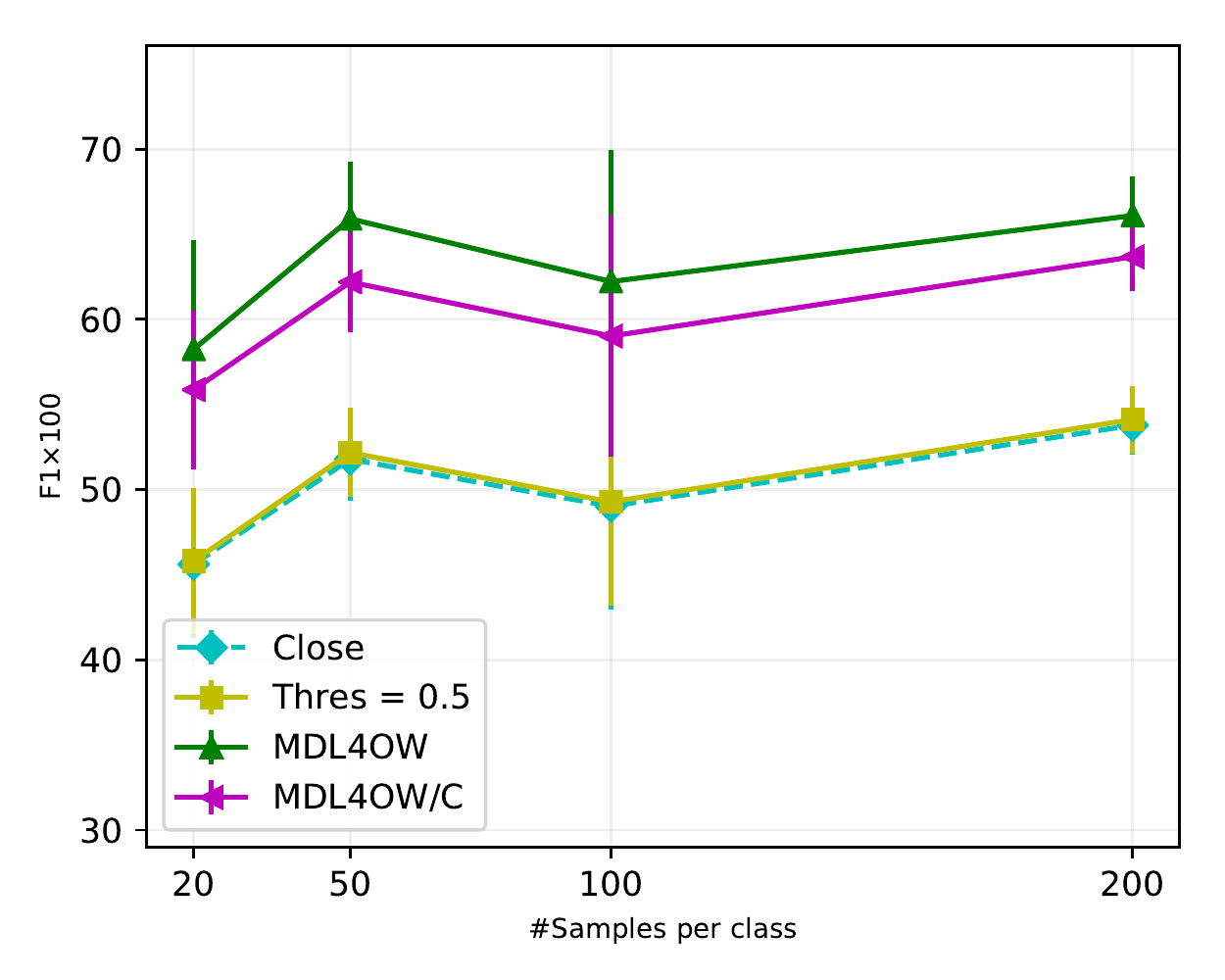}} 
\subfigure[Indian, OW3]{\includegraphics[width=0.3\textwidth]{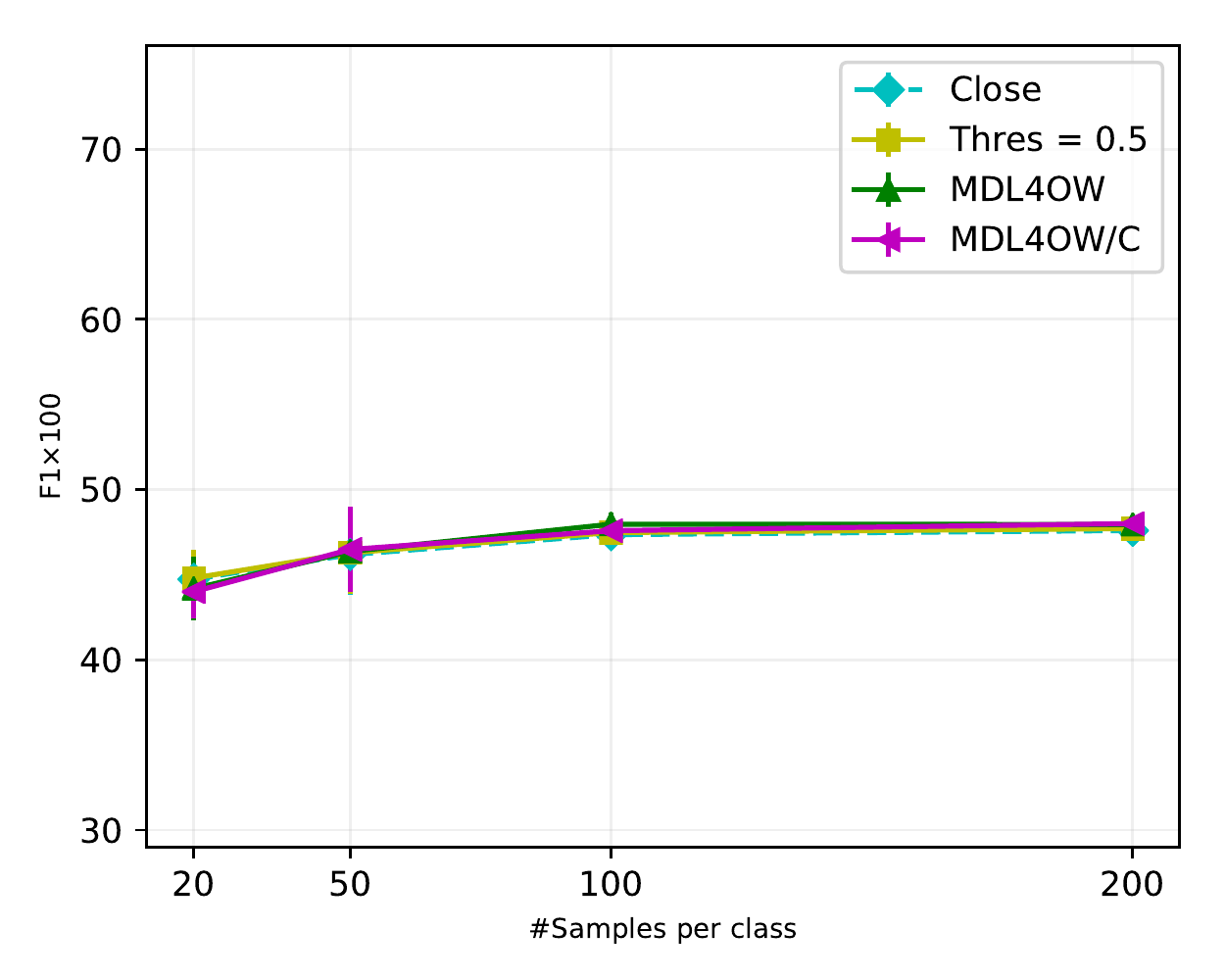}} 
\caption{F1 as a function of the number of training samples per class. * incomplete classification system.} 
\label{fig:f1_curve}
\end{figure*}

    In this section, we evaluate the proposed method with imperfect classification systems. Land cover classification systems are multilevel. For fine-grained crop mapping, each crop can be categorized as an individual class, but they all belong to a larger class,  vegetation. The hierarchy of land cover classification systems makes it easy to distinguish between impervious and vegetation but difficult to distinguish between meadows and trees. Therefore, we designed a series of experiments, where we selected some classes into training, as shown in Table \ref{tab:open_select}. Some of them are all from impervious classes, leading to a classification system without vegetation (Pavia OW3); some of them are the same crop type with a minor difference in terms of growing weeks (Salinas OW1). 
    
    A classification system lacking the basic elements (impervious, vegetation, and water) is extremely incomplete and in need of the proposed method to reject the unknown classes. For a complete classification system, this task will be extremely difficult, as the interclass variation might be too small. 
    For the experimental details,  we select 3, 6, and 3 training classes for the Pavia, Salinas, and Indian datasets, respectively, and all classes as the test set, as shown in Table \ref{tab:open_select}, leading to openness values of 42.3\%, 29.3\%, and 45.2\%, respectively. Three open-world (OW) scenarios are tested for each dataset: some are complete, and some are incomplete (denoted with *). 
    
    The F1 score as a function of the number of training samples per class is shown in Figure \ref{fig:f1_curve}. As shown, the performance of the proposed method varies. For the incomplete classification systems (with *), the proposed MDL4OW significantly improves the classification in terms of the F1 score. For the others, the improvement is less significant. Interestingly, the accuracy in terms of F1 is not affected by the number of training samples per class.  In this case, collecting more samples cannot enhance the classification in the OW with the existence of unknown classes. This provides the insight that, for hyperspectral image classification, a machine learning model needs to identify the unknown classes; otherwise the classification accuracy cannot be improved by simply adding more data, as those unknowns will decrease the precision because the model is forced to assign a label.
    
    More details compared with the state-of-the-art methods in terms of the three evaluation metrics (OA, F1, and mapping error) are shown in Tables \ref{tab:pavia_compare}, \ref{tab:salinas_compare}, and \ref{tab:indian_compare} for the Pavia, Salinas, and Indian datasets, respectively. The best classification is highlighted in bold in each setting. In most cases, the proposed method outperforms the competitors. 
    
    Although, in theory, MDL4OW is supposed to have a better classification compared to MDL4OW/C under the few-shot context, in this experiment, MDL4OW/C can sometimes achieve very good results. The problem of the class-wise method is that, when it comes to few-shot learning, there are not enough instances to estimate the distribution. But if the openness is large, meaning the unknown data exceeds the known data (this is rare in hyperspectral image classification, since it is not difficult to define an acceptable classification system where there are more knowns than unknowns), then the overlapped region of the known and unknown instances would be large enough. In this case, the tail size of the known instances is large enough; even a small set of training data can achieve a reasonable estimation of the distribution. More details about the tail analysis can be found in section \ref{ssec:tail_sensitivity}

\begin{table}[!t]
  \centering
  \caption{Classification comparison on the Pavia dataset: OA (\%), F1$\times$100 and mapping error (\%). Openness = 42.3\%. * incomplete classification system. Due to the existence of unknown instances, the maximum error can be larger than 100\% and is calculated as $Error_{max} = 2 \times (1 + A_{gt,C+1}/\sum_{i=1}^{C}{A_{gt,i}})$.}
  \scalebox{0.7}[0.7]{
    \begin{tabular}{|c|c|ccc|ccc|}
    \hline
    &  &    \multicolumn{3}{c|}{Few-shot}     &   \multicolumn{3}{c|}{Many-shot}   \\
    \multirow{6}[4]{*}{OW1} & Method & OA    & F1    & Error & OA    & F1    & Error \\ \hline          
     & Close & 30.53  & 46.78  & 225.58  & 30.68  & 46.96  & 225.58  \\
          & Softmax & 30.59  & 46.80  & 225.38  & 30.75  & 46.98  & 225.35  \\
          & CROSR & 38.94 & 25.53 & 196.67 & 38.17 & 25.32 & 201.25 \\
          & MDL4OW & \textbf{39.94 } & 48.66  & \textbf{187.63 } & \textbf{38.18 } & 48.29  & \textbf{189.38 } \\
          & MDL4OW/C & 38.85  & \textbf{48.86 } & 189.62  & 36.96  & \textbf{48.71 } & 200.18  \\ \hline
    \multirow{5}[2]{*}{OW2*} & Close & 52.65  & 68.98  & 82.33  & 55.61  & 71.47  & 79.29  \\
          & Softmax & 52.65  & 68.96  & 82.19  & 55.94  & 71.62  & 78.69  \\
          & CROSR & 33.67  & 29.56  & 114.32  & 28.12  & 22.85  & 116.41  \\
          & MDL4OW & \textbf{71.71 } & \textbf{78.60 } & \textbf{44.03 } & 70.21  & 78.41  & 48.26  \\
          & MDL4OW/C & 63.99  & 74.06  & 58.60  & \textbf{75.00 } & \textbf{81.48 } & \textbf{41.98 } \\ \hline
    \multirow{5}[2]{*}{OW3*} & Close & 22.37  & 36.56  & 311.74  & 24.15  & 38.90  & 311.74  \\
          & Softmax & 22.62  & 36.61  & 310.40  & 24.46  & 38.99  & 310.43  \\
          & CROSR & 25.19  & 19.26  & 307.32  & 30.23  & 18.78  & 287.55  \\
          & MDL4OW & \textbf{61.44 } & \textbf{53.22 } & \textbf{141.99 } & \textbf{52.54 } & \textbf{49.00 } & \textbf{182.28 } \\
          & MDL4OW/C & 50.19  & 47.02  & 193.03  & 41.04  & 44.45  & 237.39  \\ \hline
    \end{tabular}}%
  \label{tab:pavia_compare}%
\end{table}%

\begin{table}[!t]
  \centering
  \caption{Classification comparison on the Salinas dataset: OA (\%), F1$\times$100 and mapping error (\%). Openness = 29.3\%.  * incomplete classification system. Due to the existence of unknown instances, the maximum error can be larger than 100\% and is calculated as $Error_{max} = 2 \times (1 + A_{gt,C+1}/\sum_{i=1}^{C}{A_{gt,i}})$.}
  \scalebox{0.7}[0.7]{
    \begin{tabular}{|c|c|ccc|ccc|}
    \hline
    &  &    \multicolumn{3}{c|}{Few-shot}     &   \multicolumn{3}{c|}{Many-shot}   \\
    \multirow{6}[4]{*}{OW1*} & Method & OA    & F1    & Error & OA    & F1    & Error \\ \hline
          & Close & 17.87 & 30.33 & 457.54 & 17.93  & 30.41  & 457.50  \\
          & Softmax & 19.34  & 30.71  & 449.38  & 18.21  & 30.48  & 455.98  \\
          & CROSR & \textbf{73.24} & 17.91 & \textbf{144.16} & 51.82 & 7.17  & 257.43 \\
          & MDL4OW & 66.96  & 52.11  & 178.38  & 63.09  & 48.22  & 199.96  \\
          & MDL4OW/C & 71.80  & \textbf{55.63 } & 151.29  & \textbf{73.16 } & \textbf{56.75 } & \textbf{145.44 } \\ \hline
    \multirow{5}[2]{*}{OW2} & Close & 37.40  & 54.44  & 165.72  & 37.63  & 54.68  & 165.72  \\
          & Softmax & 38.42  & 54.84  & 162.94  & 37.96  & 54.81  & 164.85  \\
          & CROSR & 45.09  & 31.09  & 144.81  & 42.29  & 29.66  & 153.30  \\
          & MDL4OW & 47.41  & 58.11  & 135.71  & 52.16  & 60.50  & 124.12  \\
          & MDL4OW/C & \textbf{57.75 } & \textbf{63.00 } & \textbf{106.92 } & \textbf{60.05 } & \textbf{64.82 } & \textbf{102.06 } \\ \hline
    \multirow{5}[2]{*}{OW3} & Close & 39.97  & 57.12  & 148.21  & 40.26  & 57.41  & 148.21  \\
          & Softmax & 40.16  & 57.19  & 147.71  & 40.43  & 57.47  & 147.79  \\
          & CROSR & 48.92  & 32.17  & 126.45  & 50.71  & 33.03  & 122.28  \\
          & MDL4OW & 58.12  & 64.95  & 96.97  & 55.39  & 63.72  & 105.17  \\
          & MDL4OW/C & \textbf{61.71 } & \textbf{66.37 } & \textbf{83.93 } & \textbf{55.96 } & \textbf{64.11 } & \textbf{105.01 } \\ \hline
    \end{tabular}}%
  \label{tab:salinas_compare}%
\end{table}%

\begin{table}[!t]
  \centering
  \caption{Classification comparison on the Indian dataset: OA (\%), F1$\times$100 and mapping error (\%). Openness = 45.2\%.  * incomplete classification system. Due to the existence of unknown instances, the maximum error can be larger than 100\% and is calculated as $Error_{max} = 2 \times (1 + A_{gt,C+1}/\sum_{i=1}^{C}{A_{gt,i}})$.}
  \scalebox{0.7}[0.7]{
    \begin{tabular}{|c|c|ccc|ccc|}
    \hline
    &  &    \multicolumn{3}{c|}{Few-shot}     &   \multicolumn{3}{c|}{Many-shot}   \\
    \multirow{6}[4]{*}{OW1} & Method & OA    & F1    & Error & OA    & F1    & Error \\ \hline
          & Close & 24.24  & 39.00  & 264.55  & 27.19  & 42.76  & 264.55  \\
          & Softmax & 24.50  & 39.07  & 263.34  & 27.49  & 42.85  & 263.46  \\
          & CROSR & 36.25 & 41.54 & 206.9 & 31.96 & 44.31 & 245.84 \\
          & MDL4OW & \textbf{36.99 } & 42.02  & \textbf{205.39 } & \textbf{41.78 }  & \textbf{46.58 } & \textbf{198.04 } \\
          & MDL4OW/C & 36.92  & \textbf{42.22 } & 207.55  & 37.77  & 46.05  & 221.29  \\ \hline
    \multirow{5}[2]{*}{OW2*} & Close & 29.65  & 45.61  & 171.87  & 36.80  & 53.78  & 161.84  \\
          & Softmax & 30.33  & 45.80  & 169.56  & 37.65  & 54.12  & 159.58  \\
          & CROSR & 34.77  & 18.27  & 144.18  & 42.55  & 18.61  & 109.71  \\
          & MDL4OW & \textbf{60.08 } & \textbf{58.24 } & \textbf{87.09 } & \textbf{64.31 } & \textbf{66.10 } & \textbf{79.19 } \\
          & MDL4OW/C & 54.84  & 55.86  & 104.96  & 58.89  & 63.68  & 100.12  \\ \hline
    \multirow{5}[2]{*}{OW3} & Close & 28.82  & 44.73  & 219.84  & 31.23  & 47.60  & 219.84  \\
          & Softmax & 29.04  & 44.80  & 219.09  & 31.54  & 47.71  & 218.86  \\
          & CROSR & \textbf{42.13 } & \textbf{48.29 } & \textbf{160.43 } & \textbf{49.63 } & 45.11  & \textbf{123.94 } \\
          & MDL4OW & 33.08  & 46.18  & 190.60  & 35.97  & 47.95  & 193.51  \\
          & MDL4OW/C & 31.06  & 46.00  & 201.28  & 34.02  & \textbf{48.00 } & 205.96  \\ \hline
    \end{tabular}}%
  \label{tab:indian_compare}%
\end{table}%

\subsection{Reconstruction analysis}
\begin{figure}[!t]
\centering
\subfigure[Original Trees]{\includegraphics[width=0.23\textwidth]{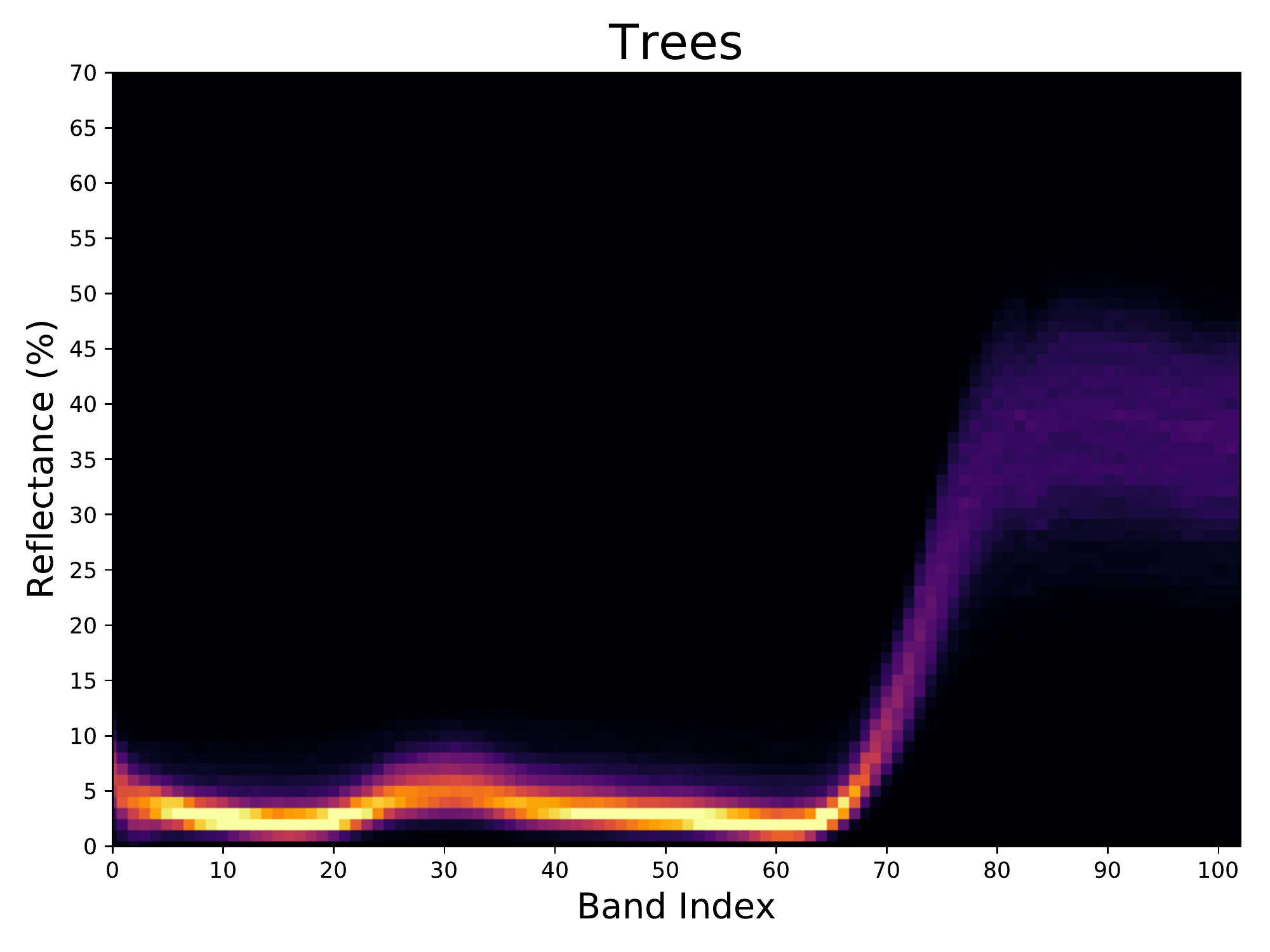}}
\subfigure[Original Building-1]{\includegraphics[width=0.23\textwidth]{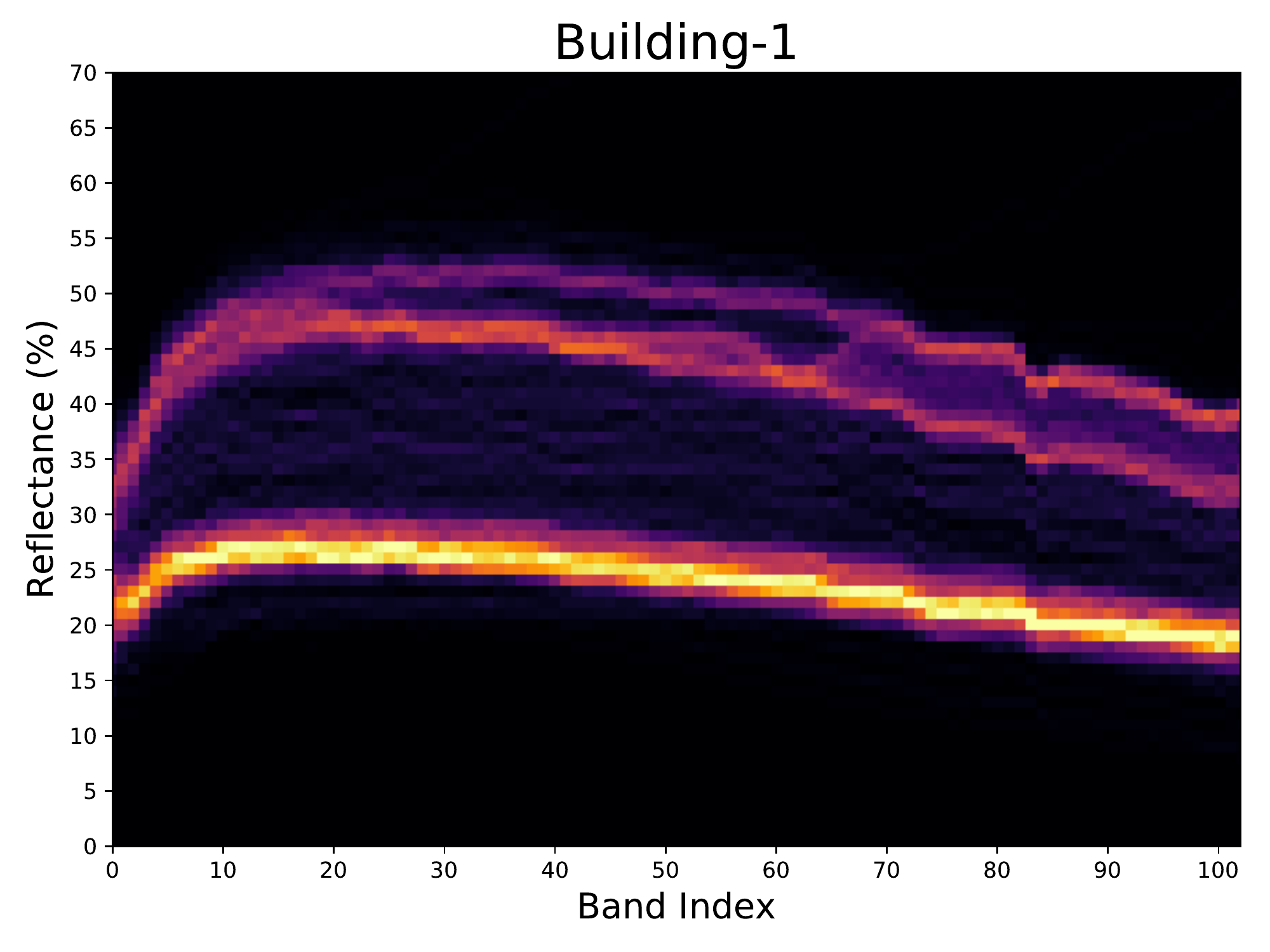}}
\subfigure[Reconstructed Trees]{\includegraphics[width=0.23\textwidth]{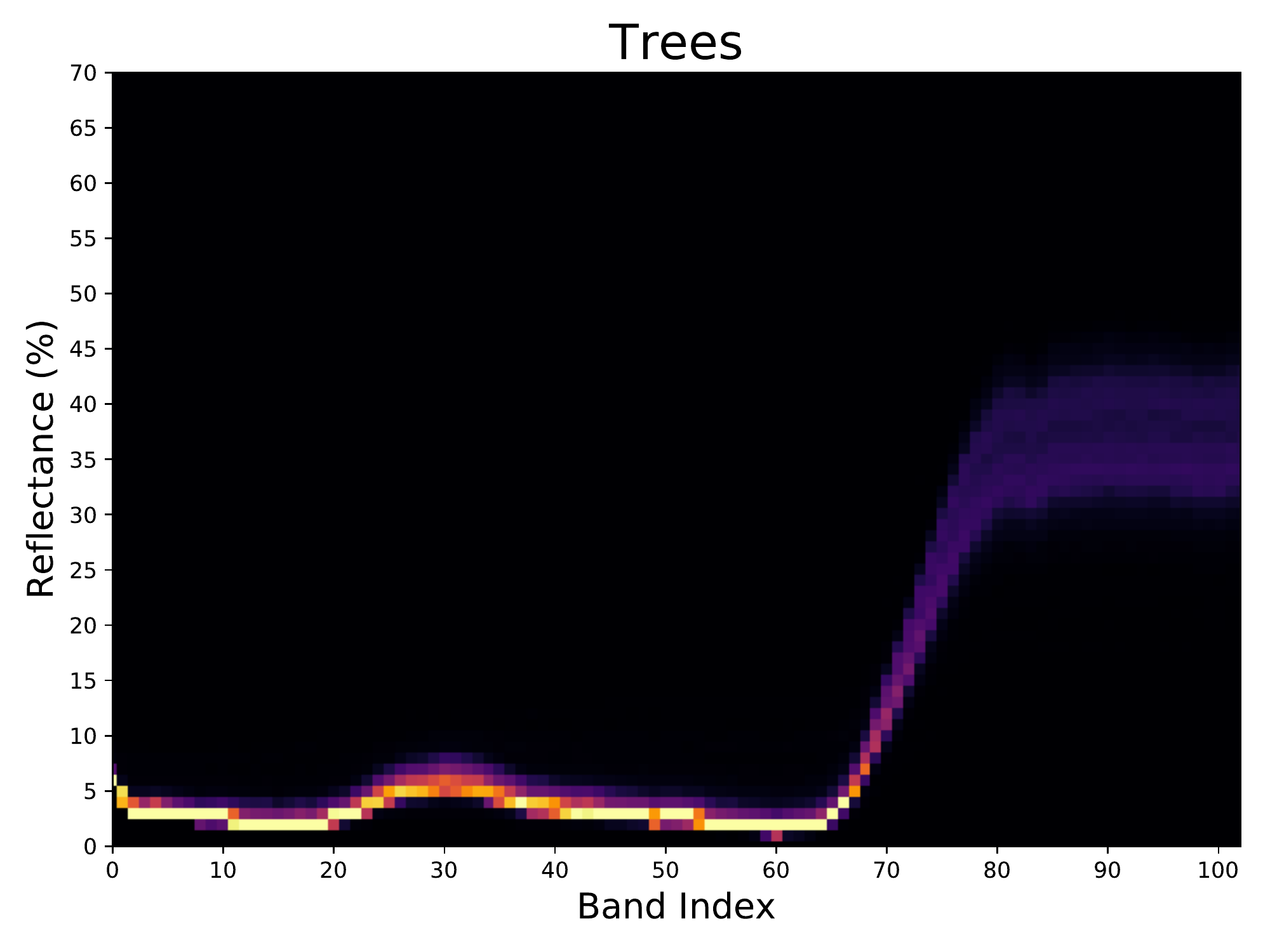}} 
\subfigure[Reconstructed Building-1]{\includegraphics[width=0.23\textwidth]{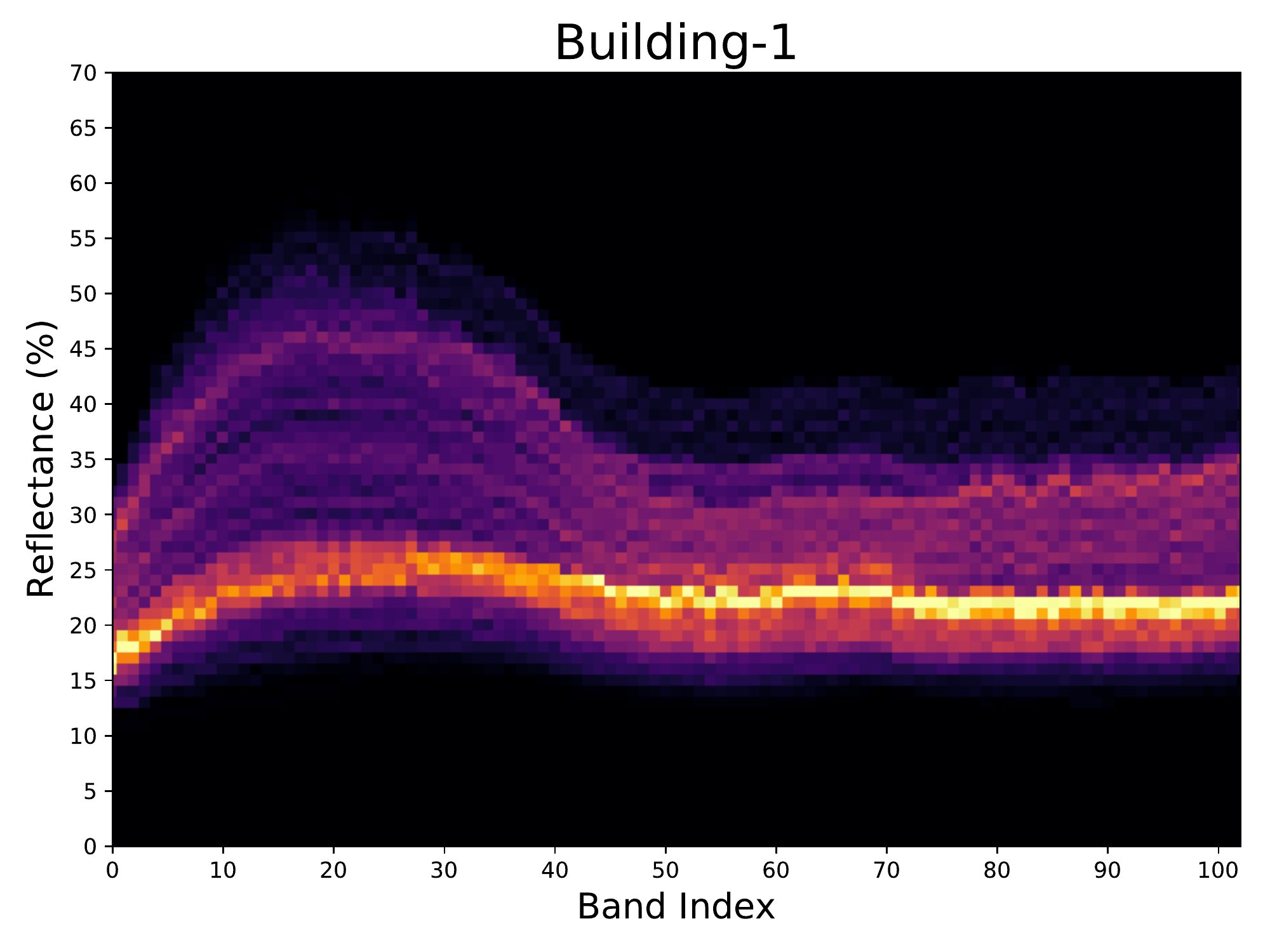}}
\caption{Density of the original (top) and reconstructed (bottom) spectral profiles of the University of Pavia data set. A brighter color indicates that more sample instances have the similar spectral profiles.}
\label{fig:spectral_pavia}
\end{figure}

Since a critical assumption of the proposed method is that the unknown classes will be poorly reconstructed, we show the original and reconstructed spectral profiles of the Pavia dataset in Figure \ref{fig:spectral_pavia}. Trees are one of the known classes in the Pavia dataset. The reconstructed spectral profiles remain as the classic vegetation curve. The reconstructed profiles are tighter, indicating less intraclass variation.  Building-1 is one of the unknown classes. We can see that the reconstructed spectral profiles are very different from the original ones. This analysis confirms the assumption that the unknown classes are poorly reconstructed, indicating the effectiveness of the proposed method. 

\subsection{Sensitivity of the tail number}
\label{ssec:tail_sensitivity}
\begin{figure}[!t]
\centering
\subfigure[Pavia, MDL4OW]{\includegraphics[width=0.23\textwidth]{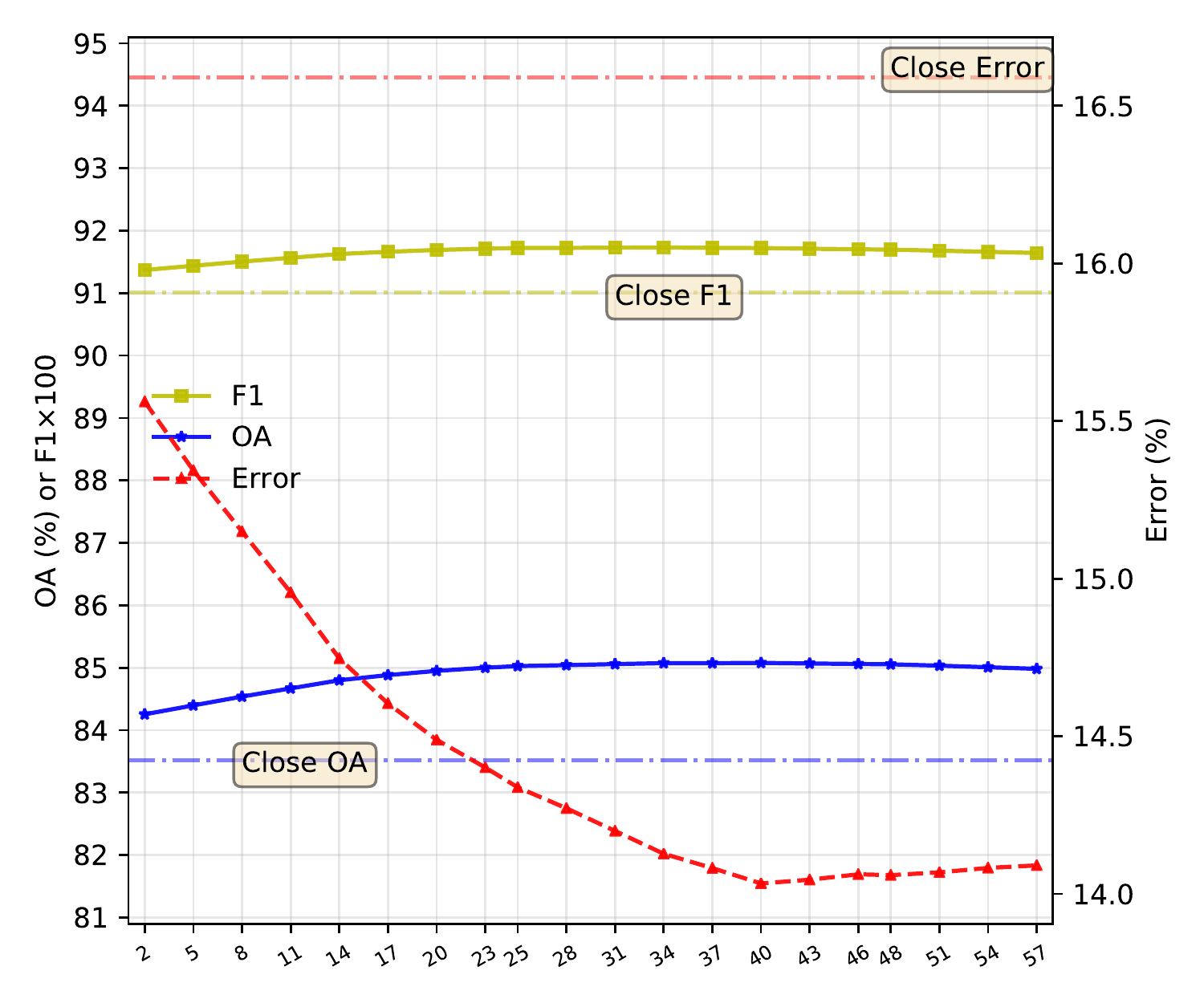}}
\subfigure[Pavia, MDL4OW/C]{\includegraphics[width=0.23\textwidth]{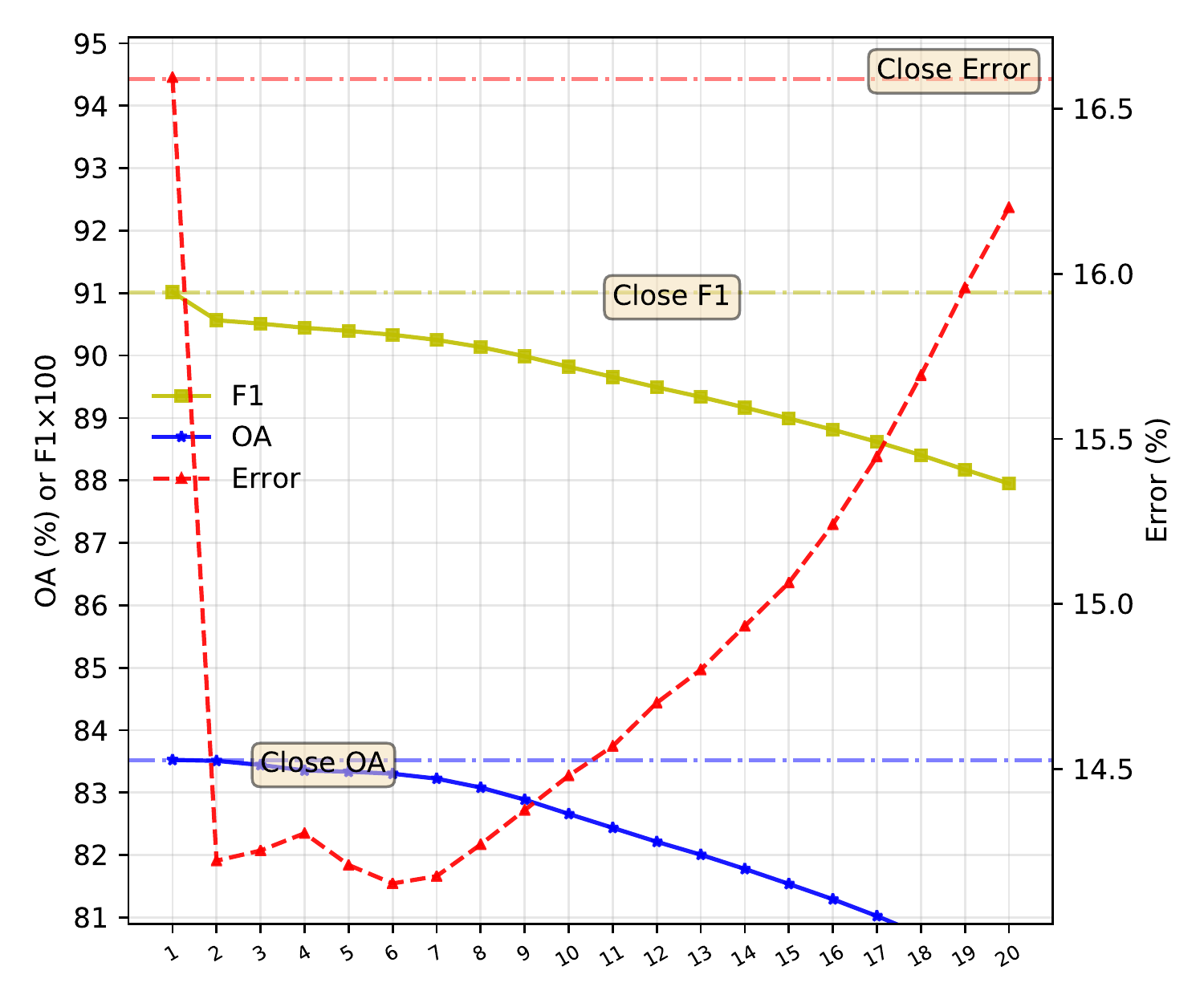}}
\subfigure[Salinas, MDL4OW]{\includegraphics[width=0.23\textwidth]{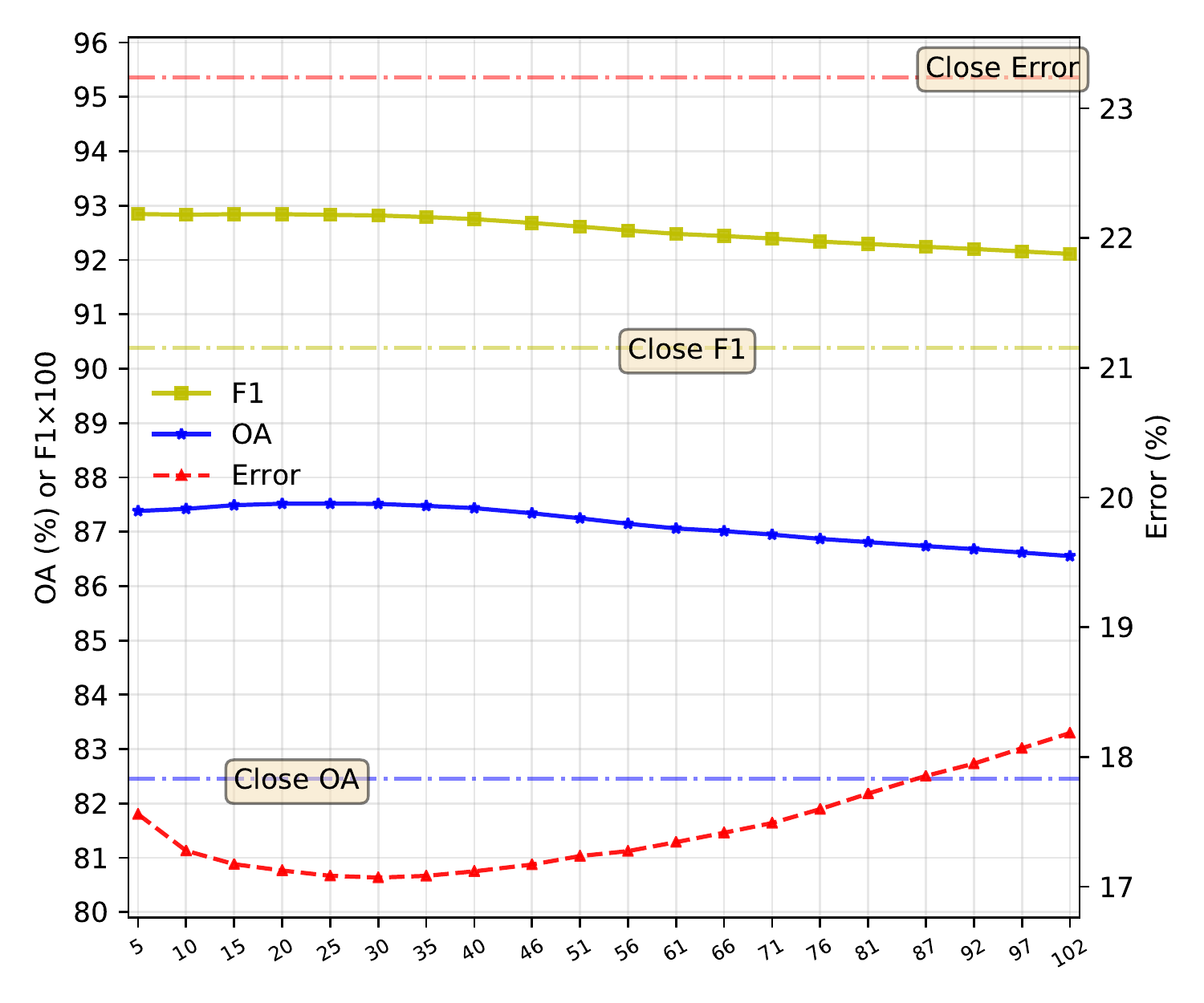}}
\subfigure[Salinas, MDL4OW/C]{\includegraphics[width=0.23\textwidth]{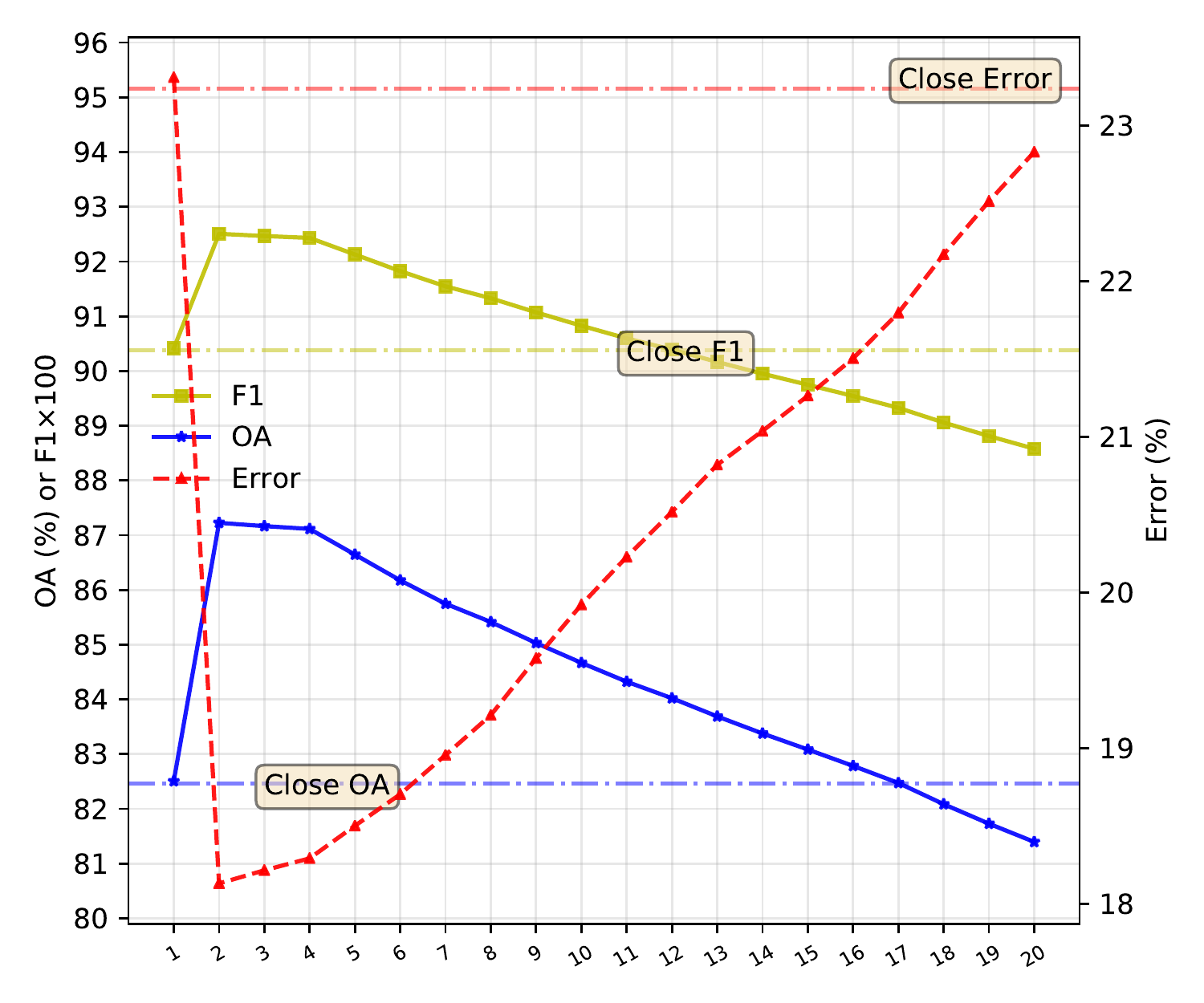}}
\subfigure[Indian, MDL4OW]{\includegraphics[width=0.23\textwidth]{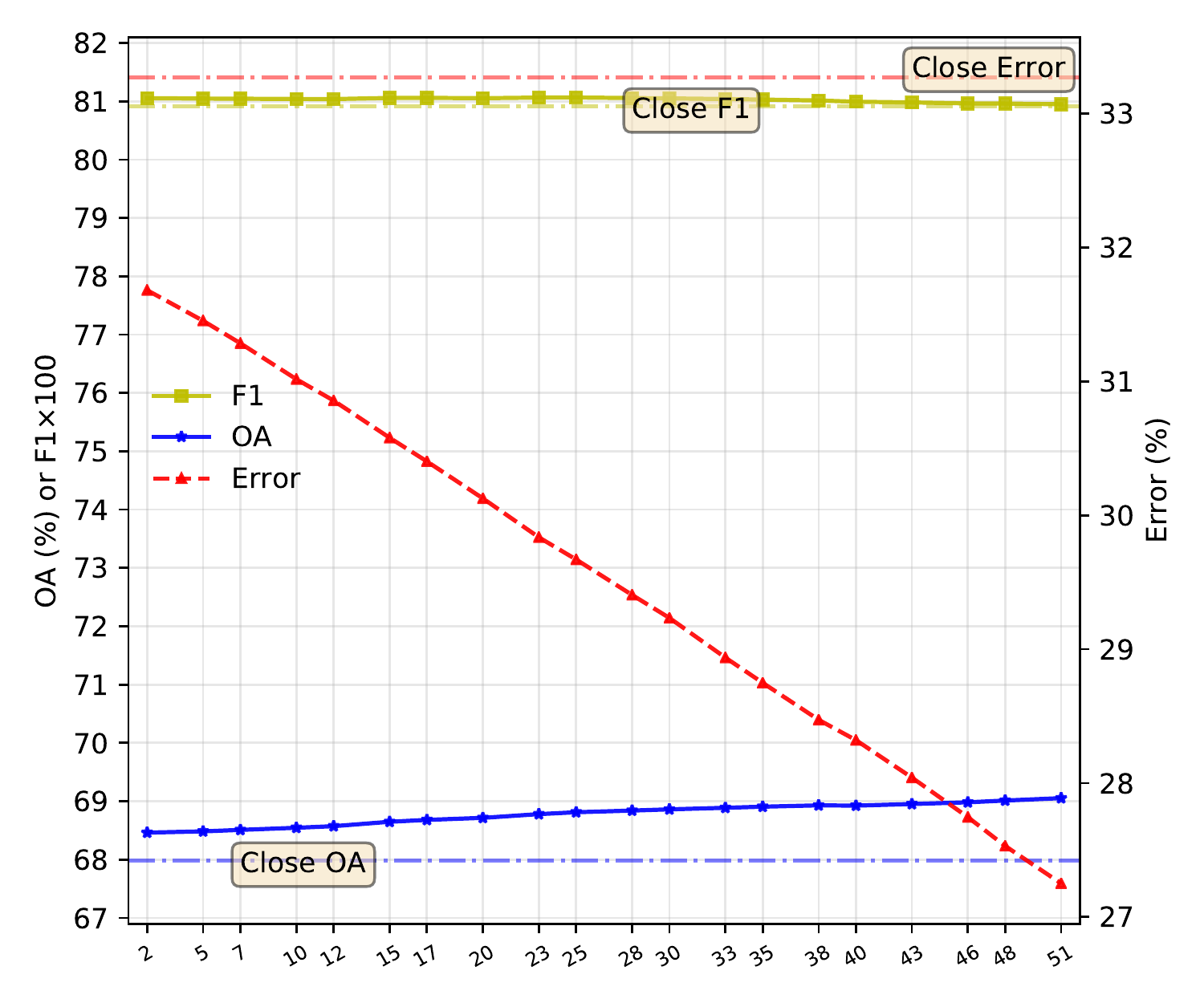}}
\subfigure[Indian,  MDL4OW/C]{\includegraphics[width=0.23\textwidth]{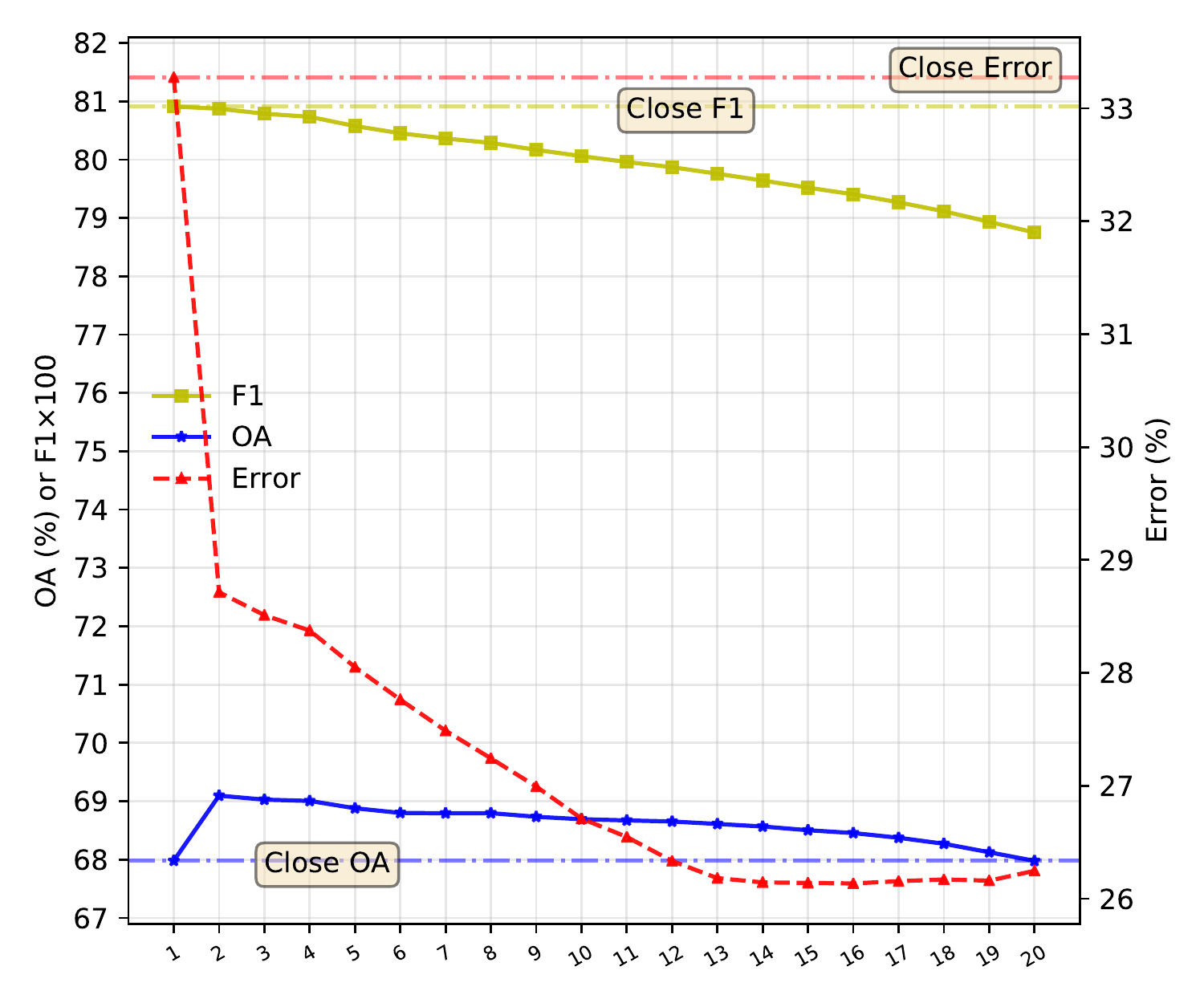}}
\caption{Sensitivity of the tail number $\tau$. Note we augmented the training samples in the experiments, so the size of training set should times 4.}
\label{fig:taileffect}
\end{figure}

    The tail number $\tau$ is an important factor in the proposed method and should be set carefully. We show the OA, F1, and mapping error as a function of the tail number under the few-shot context in Figure \ref{fig:taileffect}. We can see that the proposed MDL4OW is less sensitive to the tail number than MDL4OW/C. The former is conducted on all training data and free from estimating the class centroid in a class-wise fashion. Under the few-shot context, since the number of training samples per class is limited (20 per class), it is  difficult to apply the EVT to estimate the extreme cases. Note, we augmented the training samples so the number of training samples per class should $\times$ 4,  which in a way eases the problem. 
    
    For the Pavia and Indian datasets, the F1 scores are already lower than the closed classification  when the tail size $\tau$ is 2 (EVT does not work when the tail size $\tau$ is 1, because one instance cannot estimate a distribution). For the Salinas dataset, the safe range of the tail number in terms of F1 is 2--12, where $\tau =2$ is the optimum, showing that a smaller tail should be set, but there are not enough instances in the class-wise method.  
    
    Comparing three evaluation metrics, F1 is always higher than OA and is stricter in evaluation. Take Salinas MDL4OW/C as an example.  F1 is lower than the baseline when the tail number is greater than 12, but OA is still higher than the baseline until the tail number is greater than 17. Recall that the F1 score is calculated based on the known classes, and OA includes the unknown class. When the tail size is between 12 and 17, the accuracy of the unknown class makes OA above the baseline. But, since we care more about the known classes than the unknown, it makes more sense to use the F1 score instead of OA. As for mapping error, its real meaning is the average error on the predicted number of all known classes (i.e., land cover area) and is of direct interest to local governments. We can see that it benefits the most from the proposed method where the unknown classes are considered. 
    The evaluation of open-set classification is still an ongoing problem \cite{geng2020recent}. Although OA and F1 are the most common evaluation metrics, as the real-world meaning of mapping error is straightforward in hyperspectral image classification, it can be used as another standard metric for this application.
    
     To conclude, the proposed method MDL4OW achieved state-of-the-art results in terms of all three of the evaluation metrics under the few-shot context and is recommended for hyperspectral image classification. The class-wise version should be used only when there are enough training samples. 

\subsection{Sensitivity of loss weights}
\begin{figure*}[!t]
\centering
\subfigure[Pavia, F1, few-shot]{\includegraphics[width=0.24\textwidth]{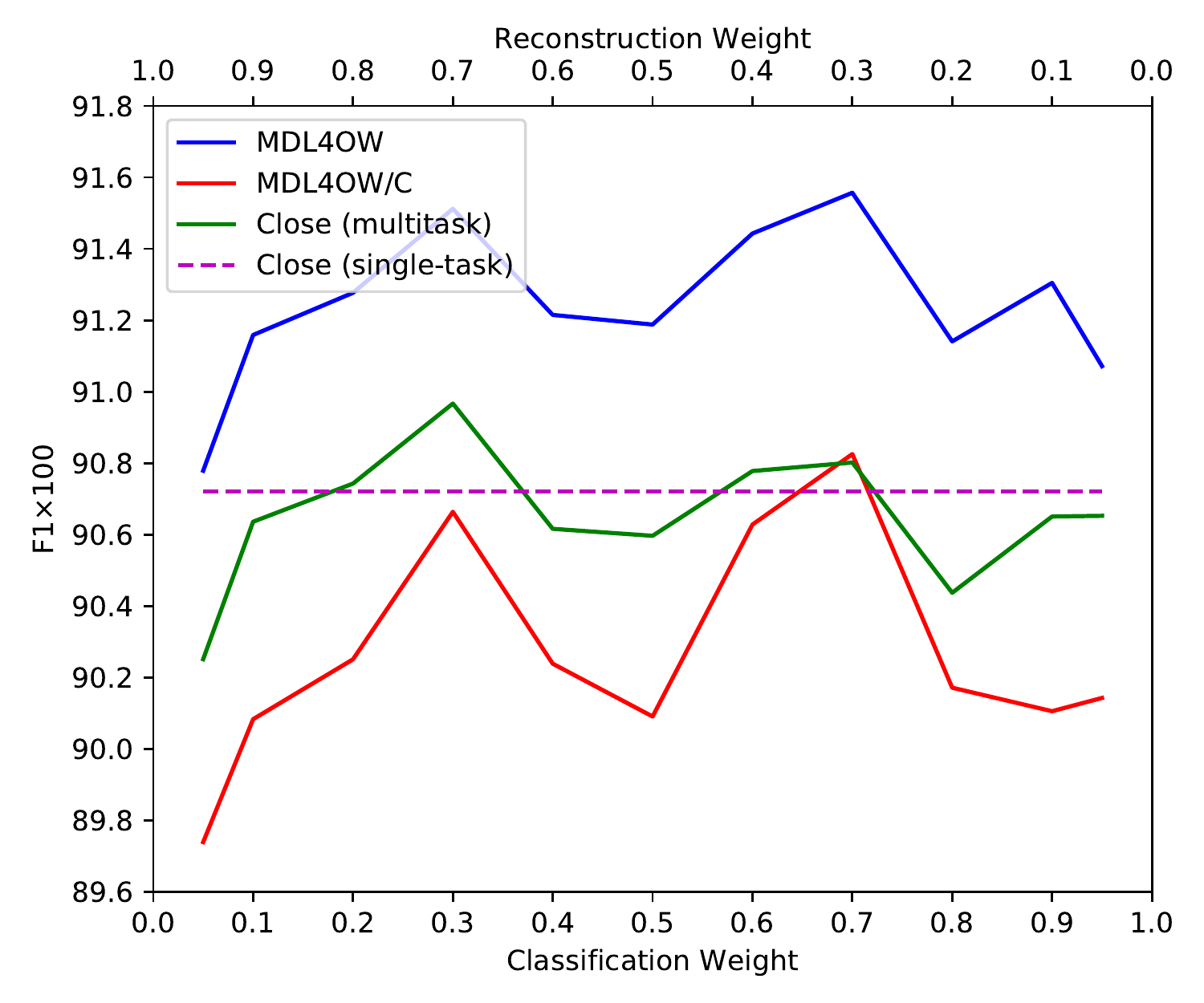}}
\subfigure[Pavia, F1, many-shot]{\includegraphics[width=0.24\textwidth]{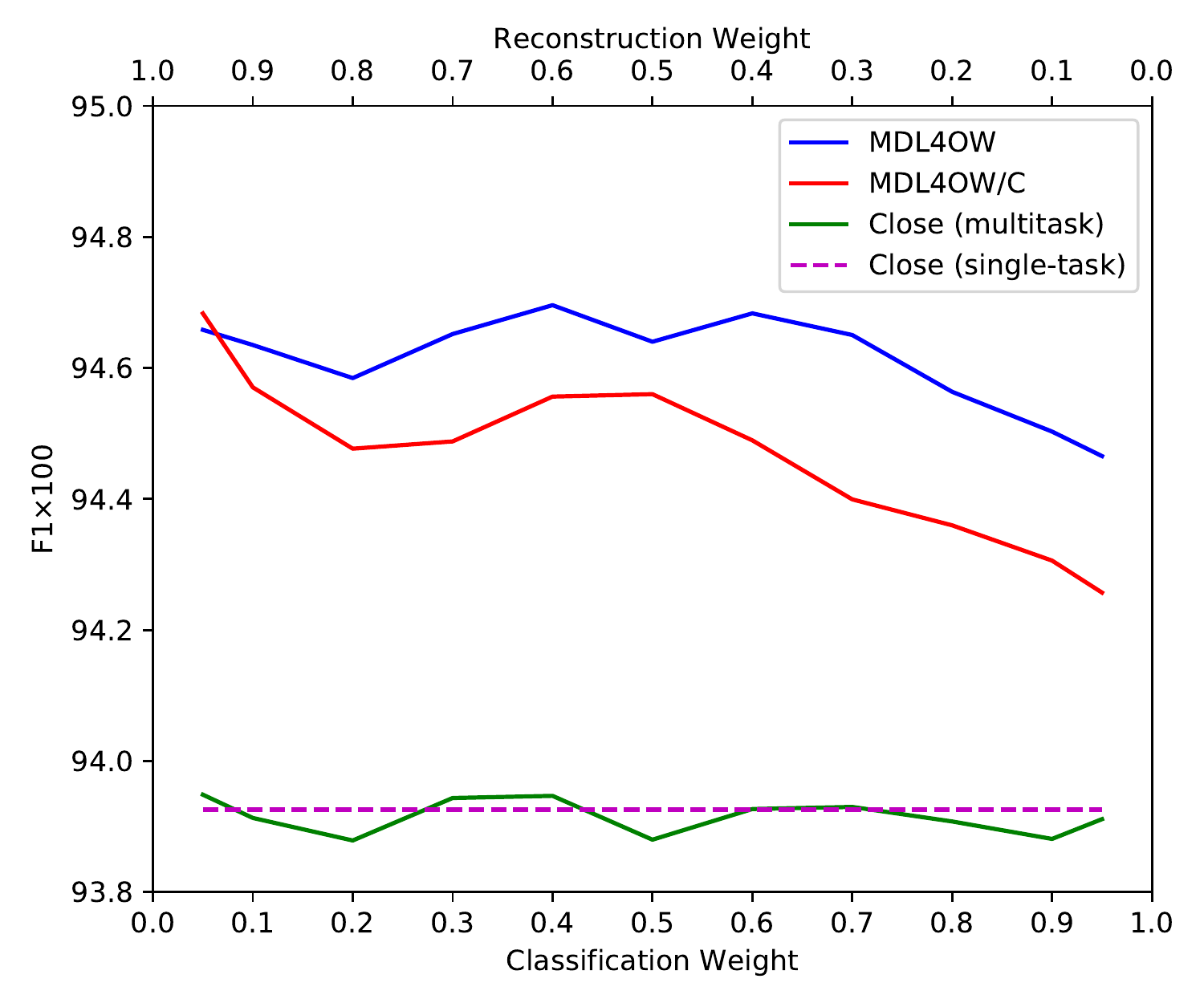}}
\subfigure[Pavia, mapping error, few-shot]{\includegraphics[width=0.24\textwidth]{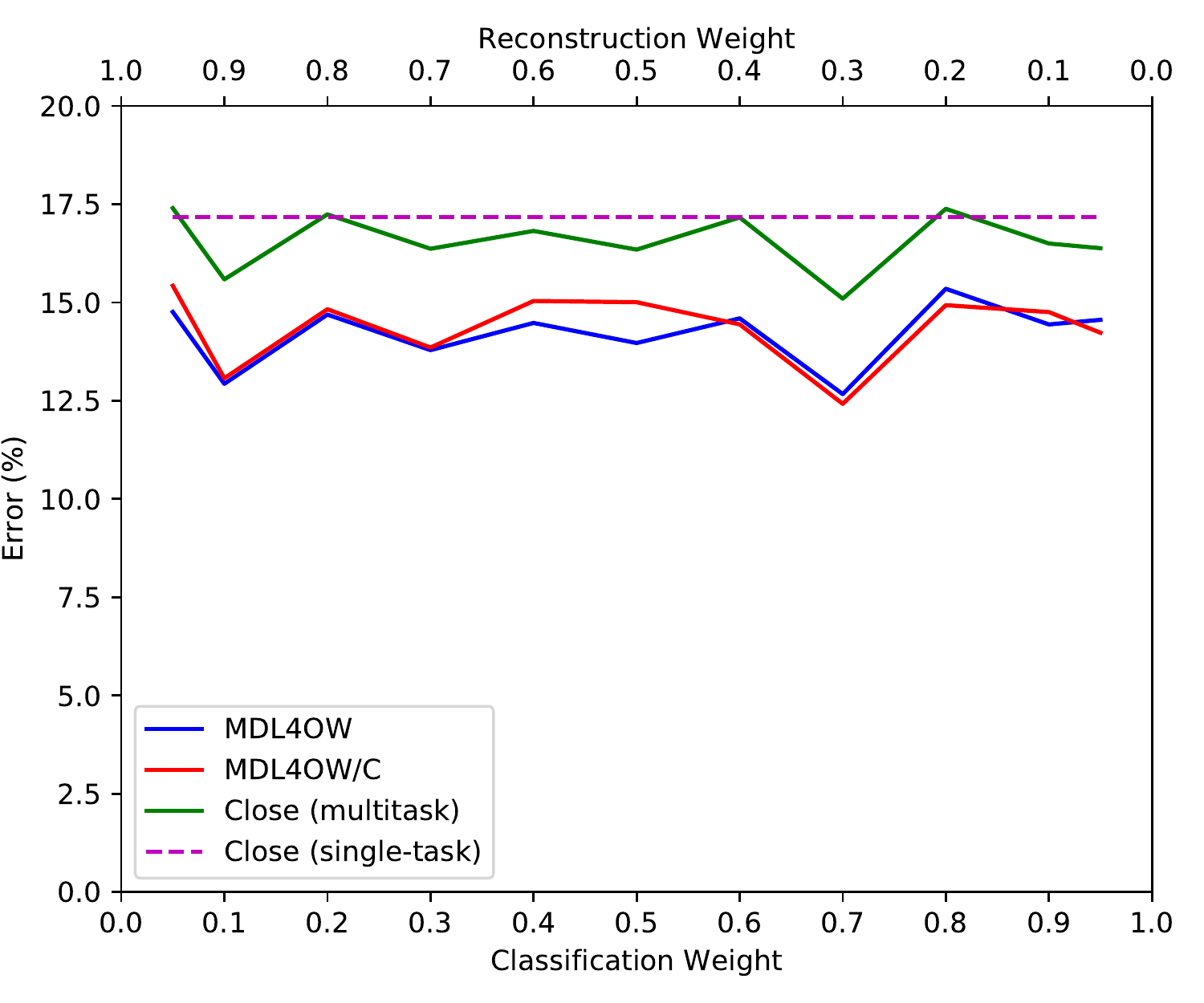}}
\subfigure[Pavia, mapping error, many-shot]{\includegraphics[width=0.24\textwidth]{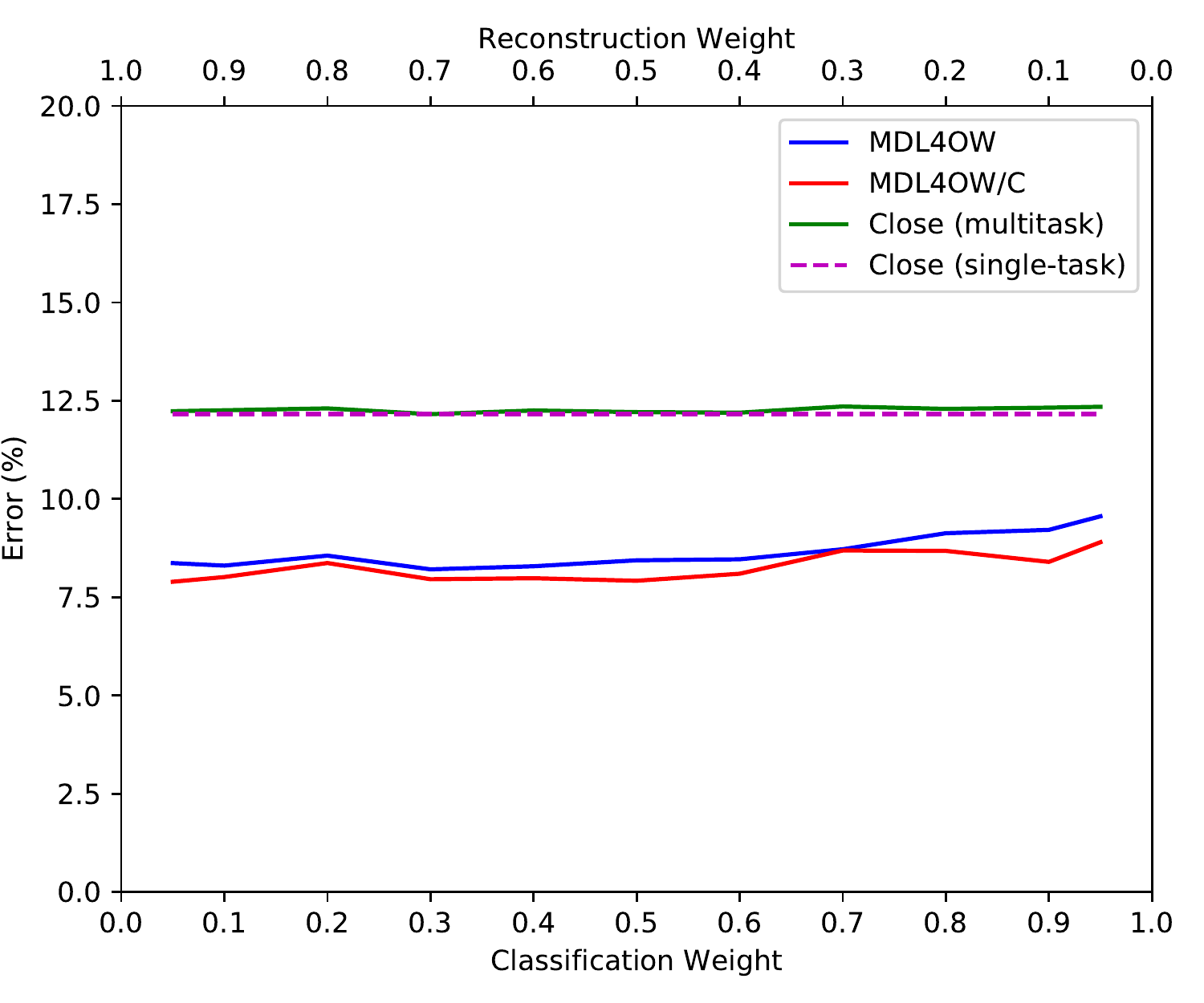}}

\subfigure[Salinas, F1, few-shot]{\includegraphics[width=0.24\textwidth]{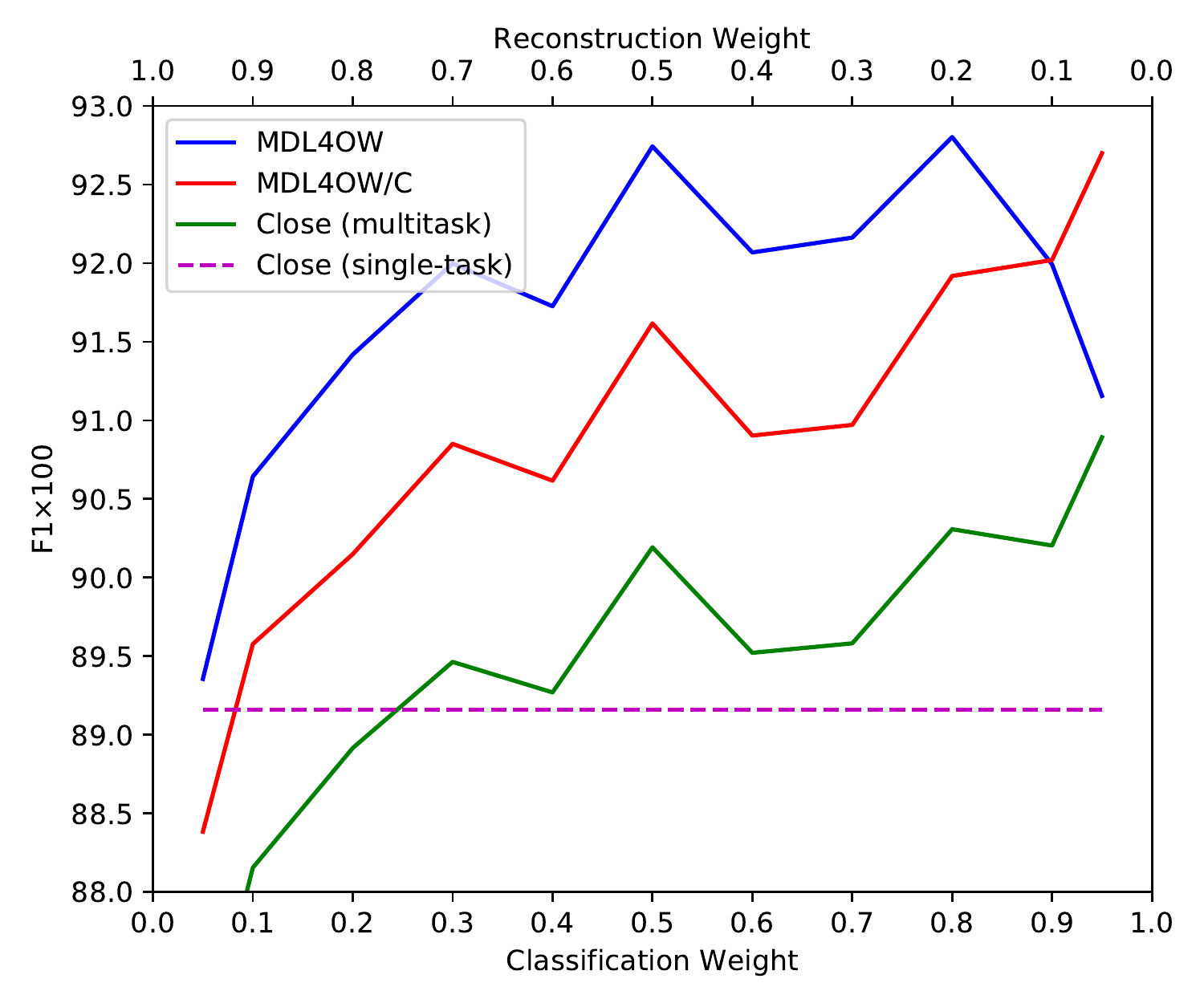}}
\subfigure[Salinas, F1, many-shot]{\includegraphics[width=0.24\textwidth]{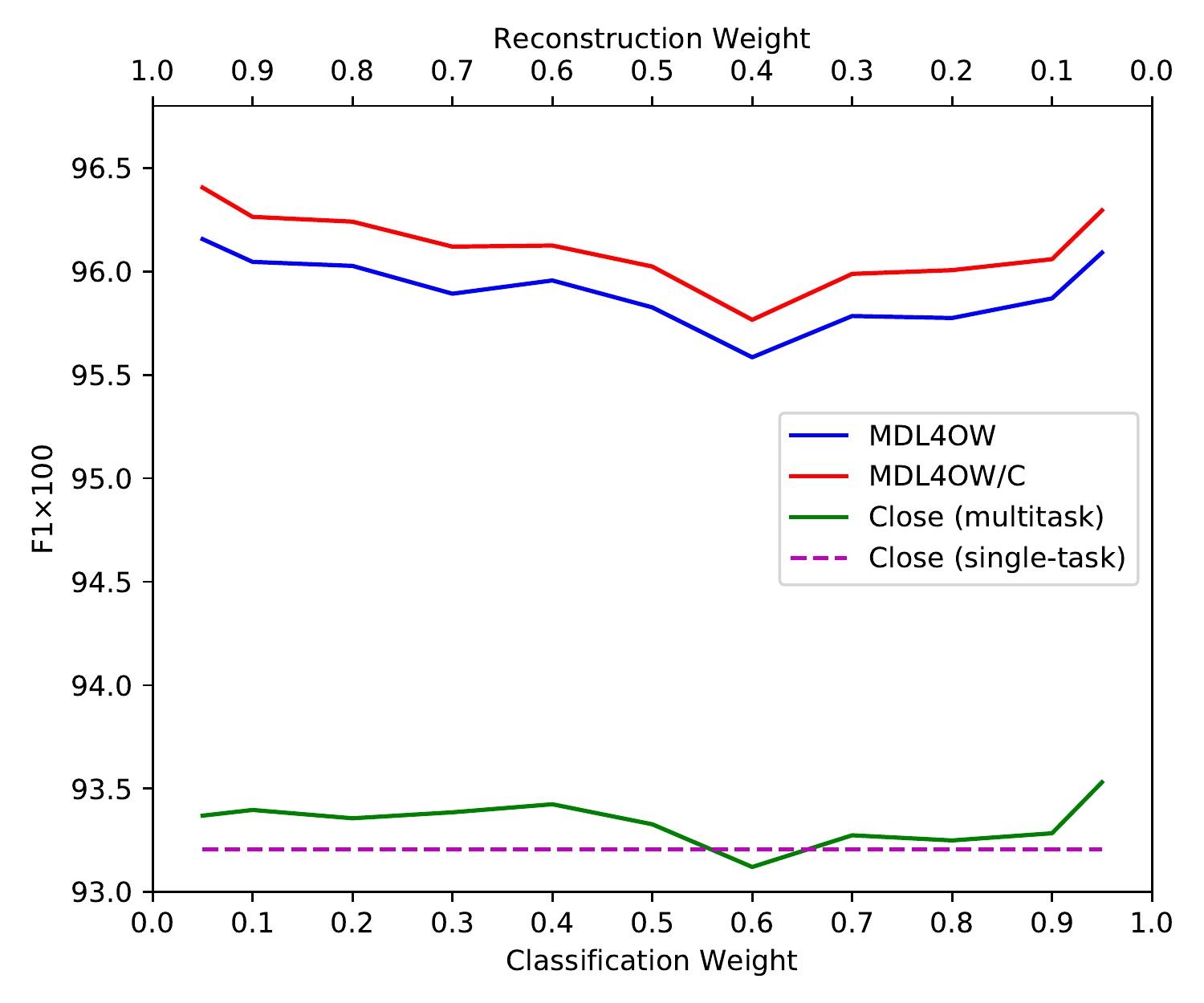}}
\subfigure[Salinas, mapping error, few-shot]{\includegraphics[width=0.24\textwidth]{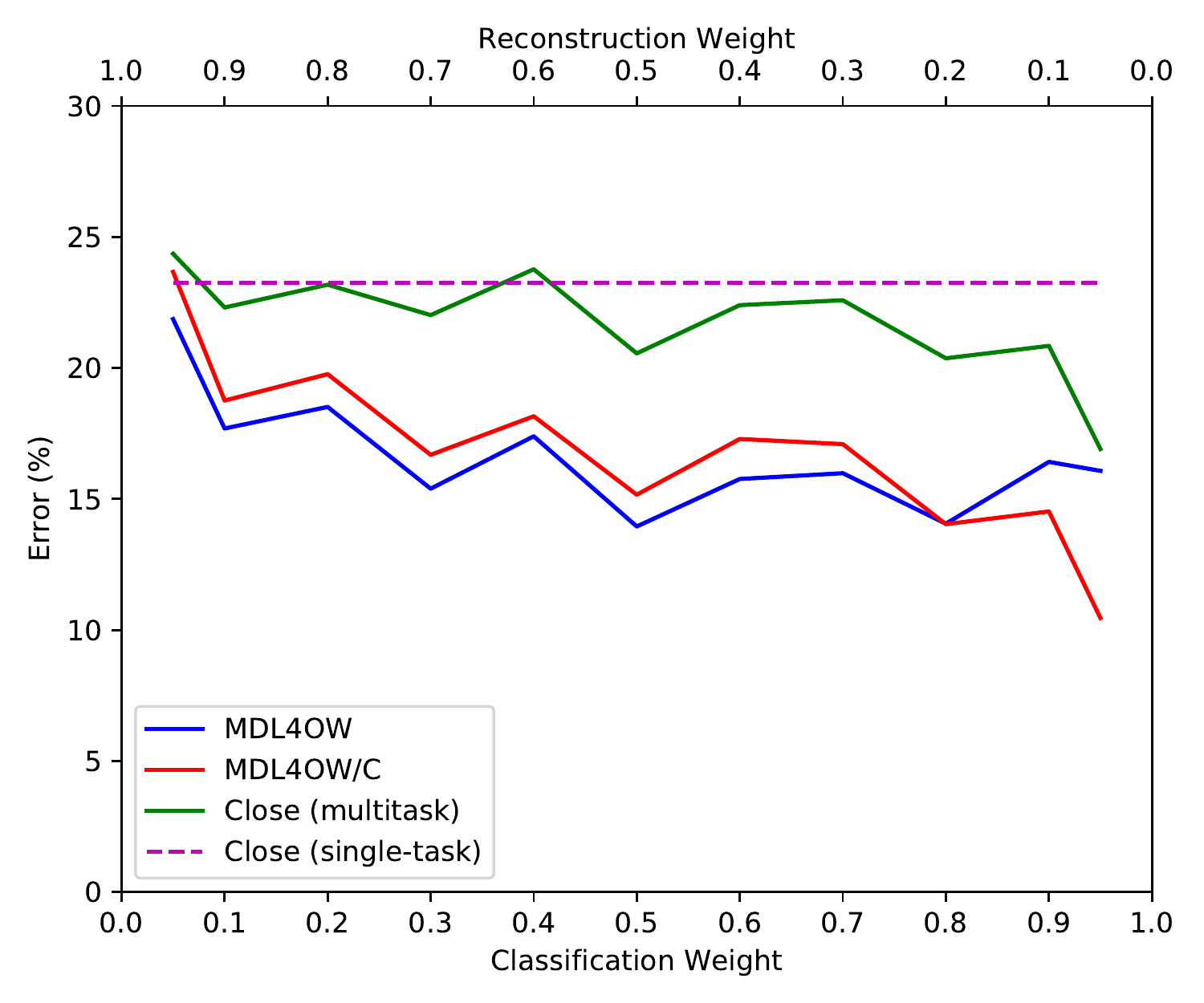}}
\subfigure[Salinas, mapping error, many-shot]{\includegraphics[width=0.24\textwidth]{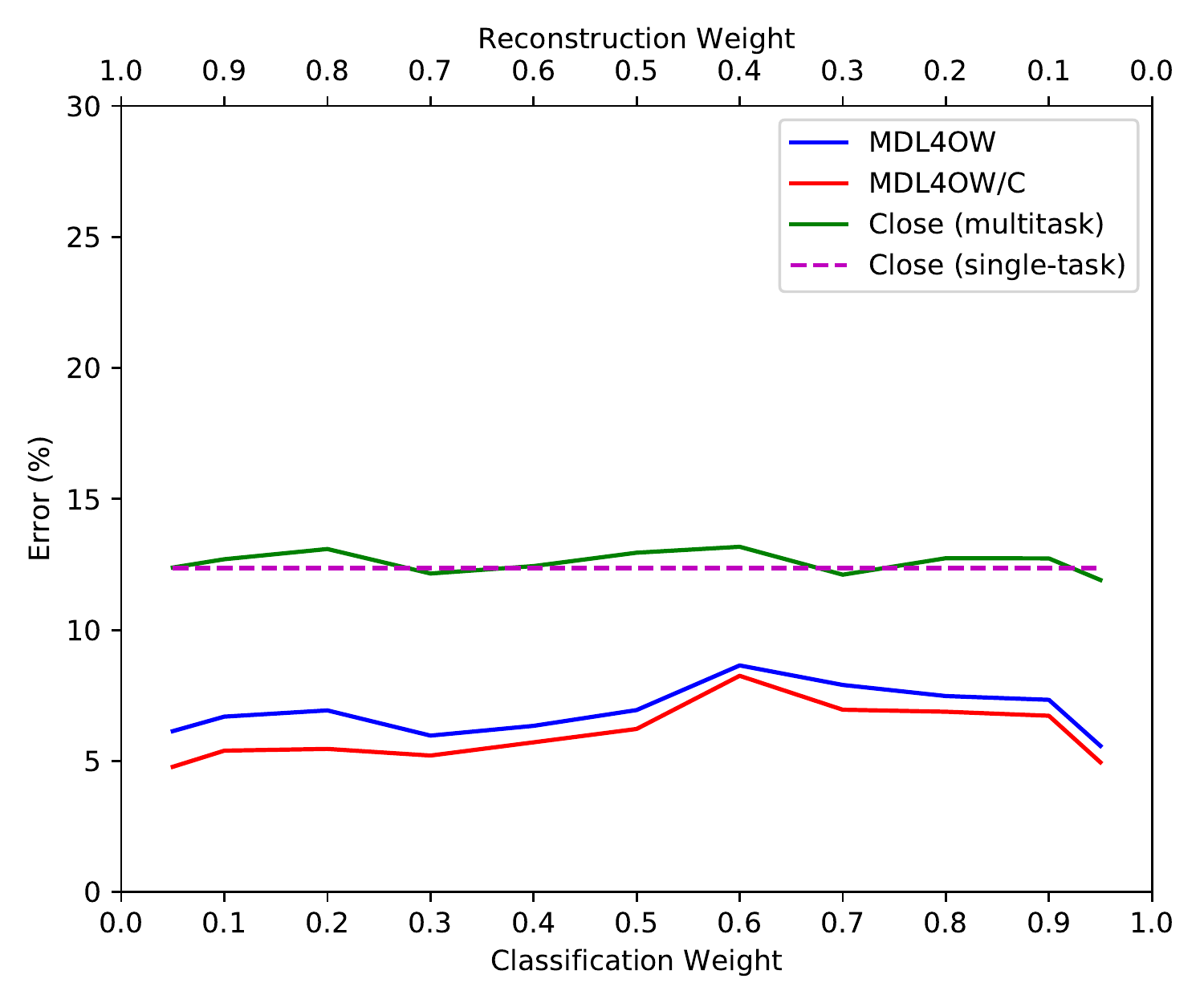}}

\subfigure[Indian, F1, few-shot]{\includegraphics[width=0.24\textwidth]{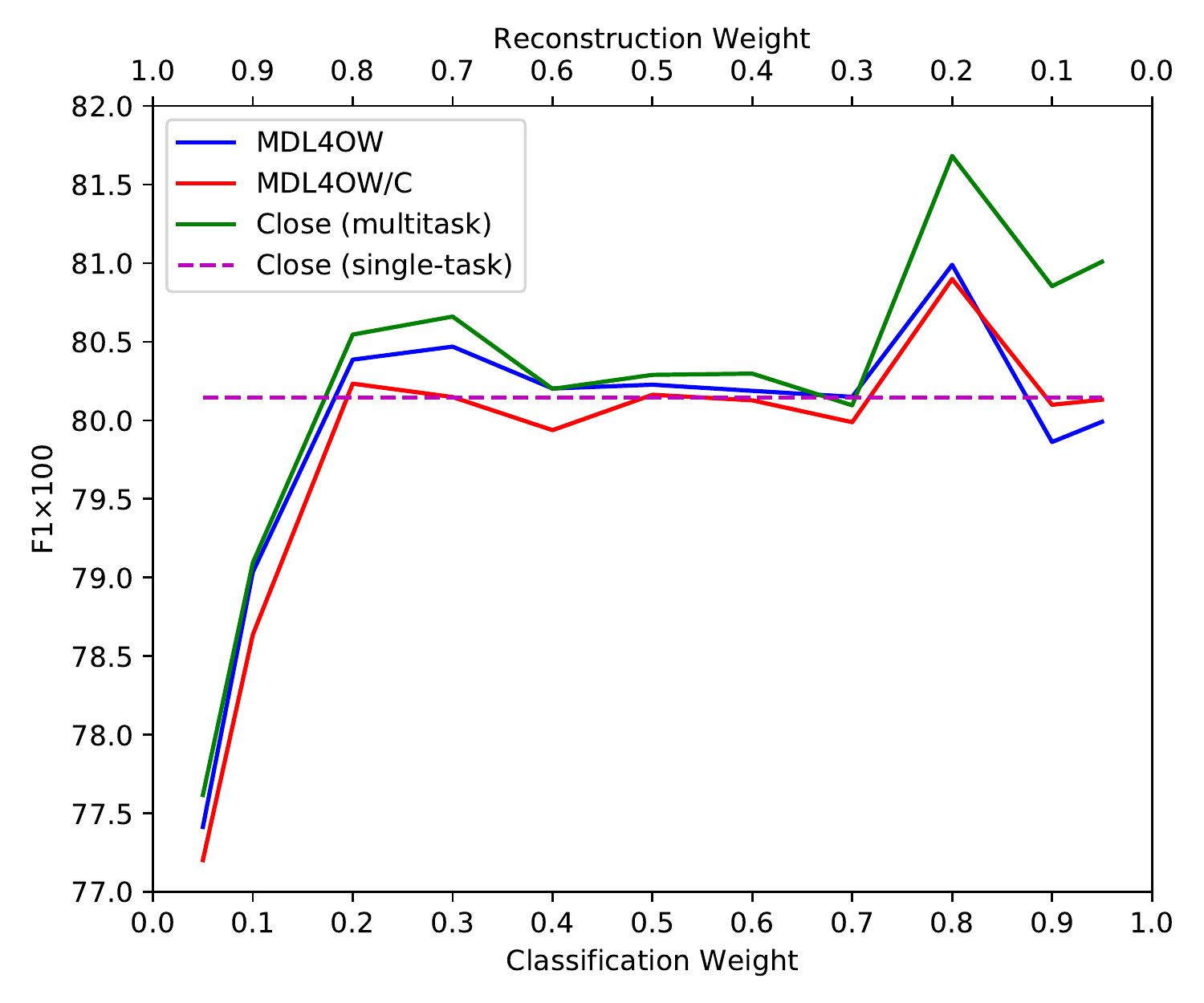}}
\subfigure[Indian, F1, many-shot]{\includegraphics[width=0.24\textwidth]{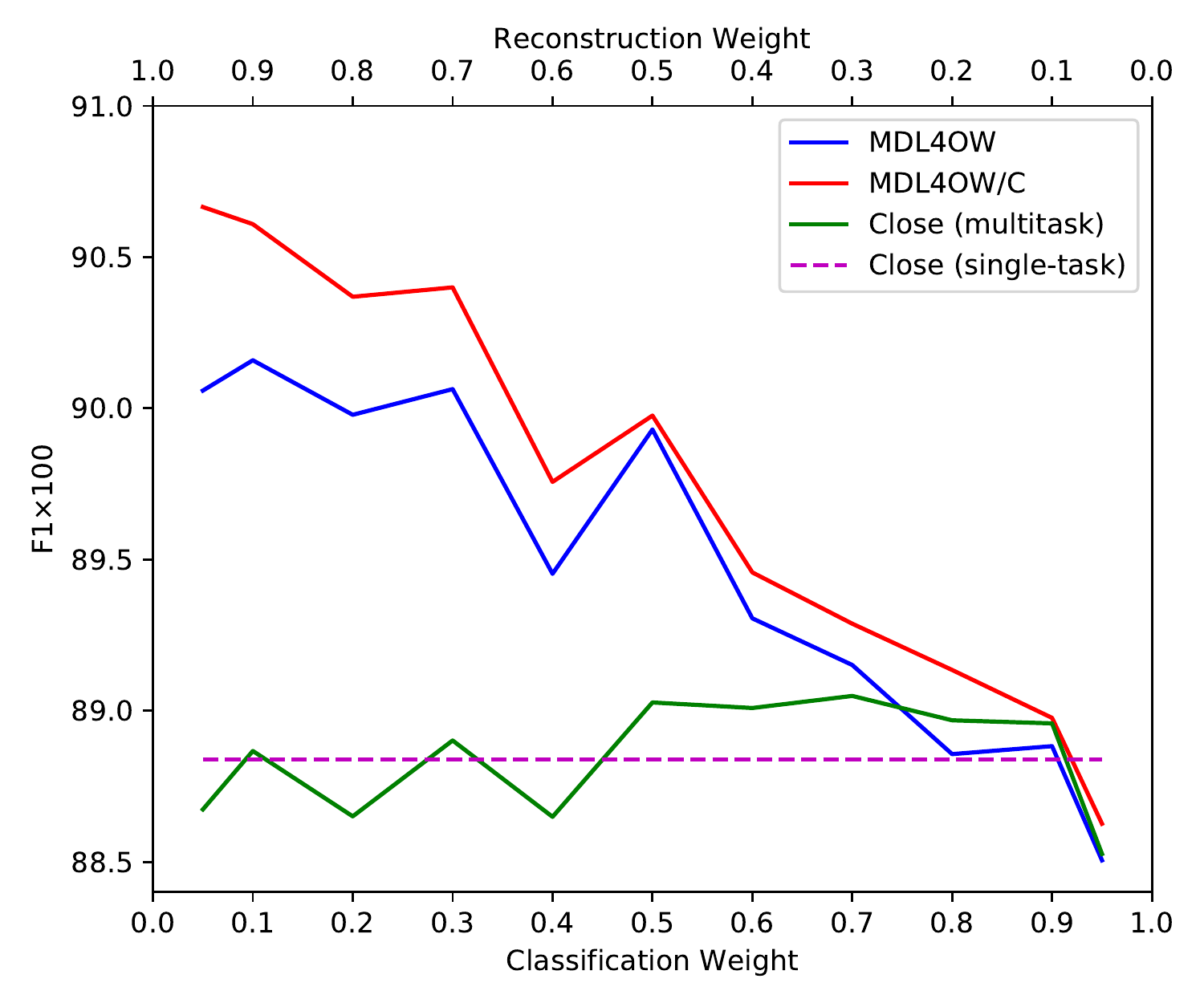}}
\subfigure[Indian, mapping error, few-shot]{\includegraphics[width=0.24\textwidth]{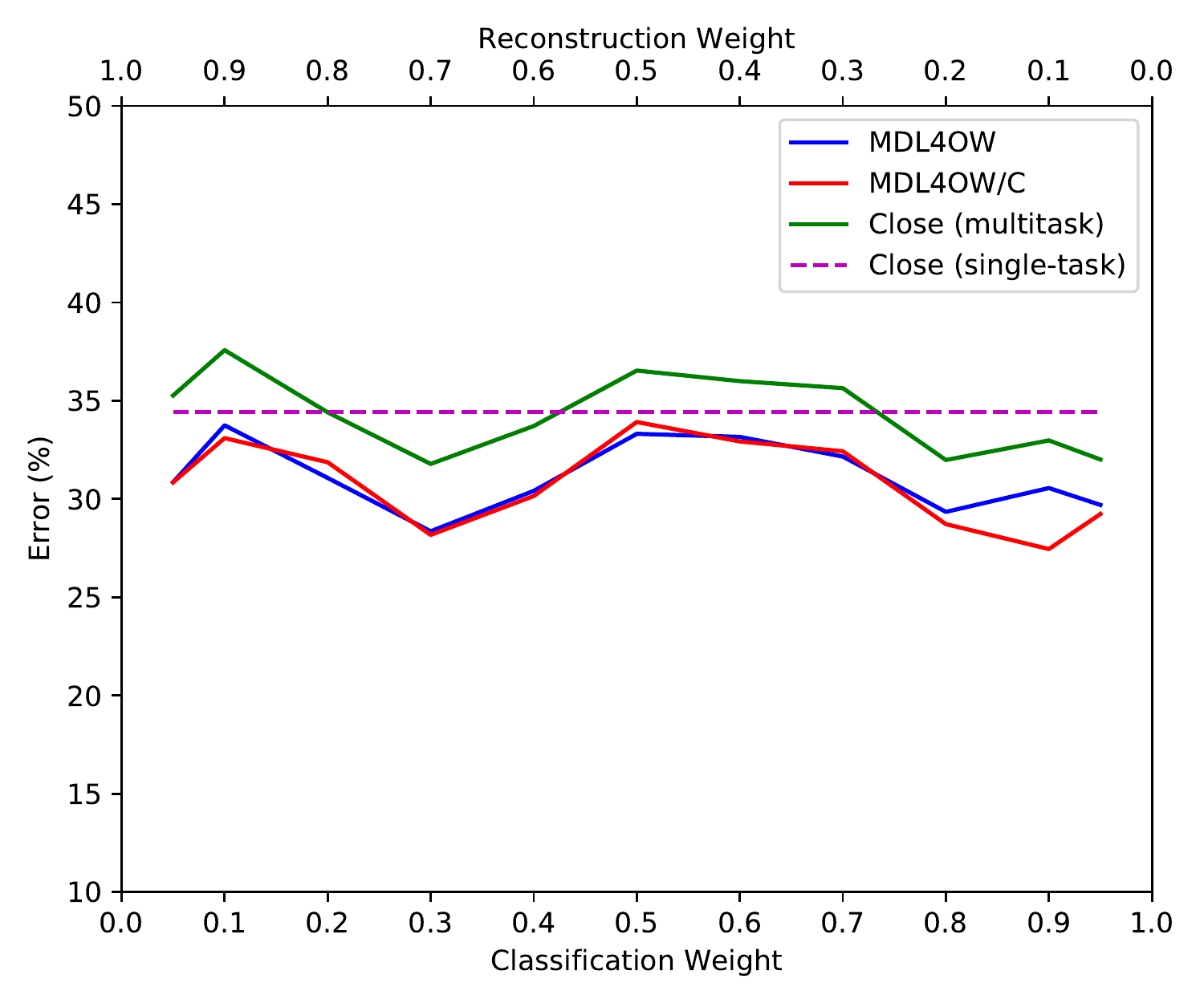}}
\subfigure[Indian, mapping error, many-shot]{\includegraphics[width=0.24\textwidth]{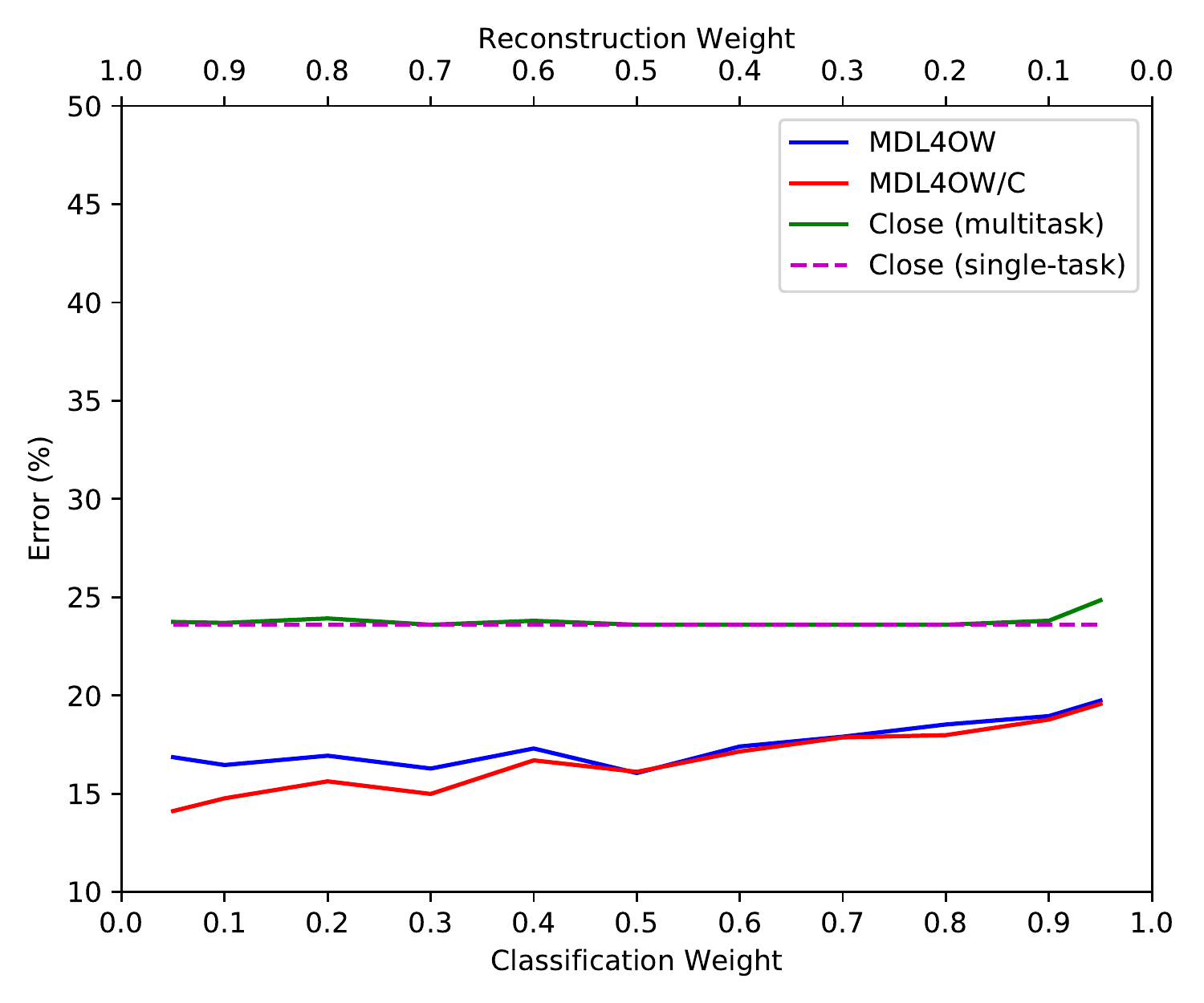}}
\caption{Sensitivity of loss weights. From first row to the third row: results on the Pavia, Salinas, and Indian datasets, respectively. From first column to the forth column: F1 score (few-shot), F1 score (many-shot), mapping error (few-shot), mapping error (many-shot). F1 score: the larger the better; mapping error: the smaller the better.}
\label{fig:loss_weight}
\end{figure*}

In this section, we show the sensitivity of loss weights in terms of F1 score and mapping error in Figure \ref{fig:loss_weight}, under both few-shot and many-shot settings. We do not show OA as the OA curve is similar to the F1 score. We set the classification weight from 0.1 to 0.9, and also examine two extreme cases with classification weight as 0.05 and 0.95.  From the first row to the third row show results on the Pavia, Salinas, and Indian datasets, respectively.  In general, both MDL4OW and MDL4OW/C can enhance the classification in terms of F1 score and mapping error, regardless of the loss weights on classification and reconstruction. One exception is Figure \ref{fig:loss_weight}a: we can see that the class-wise method MDL4OW/C would suffer from limited training samples. MDL4OW, however, successfully improved the classification in terms of F1 score.  

Although in general the methods are insensitive to the losses' weights, when under few-shot setting, a classification weight smaller than 0.2-0.3 would degrade the performance. The best results are obtained from a classification weight of 0.7-0.8. As for the many-shot setting, the performance is similar for the Pavia and Salinas data. But for the Indian data, a larger classification weight leads to a decrease on F1 score. Nevertheless, setting the classification and reconstruction weights as 0.3-0.7 can guarantee an acceptable performance of the proposed method. 

\subsection{Computation time}
\begin{table}[htbp]
\color{black}
  \centering
  \caption{Computation time in seconds}
    \begin{tabular}{|c|cc|cc|cc|}
    \hline
          & \multicolumn{2}{c|}{Pavia} & \multicolumn{2}{c|}{Salinas} & \multicolumn{2}{c|}{Indian} \\
          & Train & Predict & Train & Predict & Train & Predict \\
    \hline
    DCCNN \cite{lee2017going} & 68    & 19    & 116   & 11    & 123   & 2 \\
    WCRN \cite{liu2018wide}  & 32    & 13    & 62    & 9     & 65    & 1 \\
    HResNet \cite{liu2020multitask} & 36    & 19    & 74    & 16    & 73    & 3 \\
    CROSR \cite{yoshihashi2019classification} & 53    & 41    & 90    & 38    & 80    & 12 \\
    MDL4OW & \multirow{2}[1]{*}{61} & 33    & \multirow{2}[1]{*}{108} & 30    & \multirow{2}[1]{*}{91} & 8 \\
    MDL4OW/C &       & 34    &       & 31    &       & 8 \\
    \hline
    \end{tabular}%
  \label{tab:time}%
\end{table}%

Finally, we show the computation time on the three datasets in Table \ref{tab:time} under the few-shot context. Note MDL4OW and MDL4OW/C share the same training procedure and have the same training time. The training and predicting time of MDL4OW and MDL4OW/C is longer than its baseline, HResNet. This is because the added reconstruction task requires additional computation.  As for CROSR, although the network used is similar with MDL4OW and MDL4OW/C, it does not compute the reconstruction error and thus has a lower training time. But, since it calculates the centroid to estimate unknown scores, additional computation is required during the predicting phase.

\section{Conclusions}
\label{sec:conclusion}
In this paper, we proposed a novel multitask deep learning method for hyperspectral image classification with unknown classes in the open world, which is a first in the literature to the best of our knowledge. 
    Conventional hyperspectral image classification assumes the classification system is complete, and there will be no unknown classes in the unseen data. This assumption is risky in the real world since the Earth's surface is complex, and it is difficult to construct a perfect classification system including all the potential classes, as shown in this study using the popular hyperspectral benchmarks as examples. In this case, the number of known classes will be overestimated, resulting in an overestimation of certain land cover, e.g., crop area for food production. 
    To tackle this issue, a multitask deep learning method named MDL4OW is proposed for hyperspectral image classification with unknown classes, where a multitask network is utilized to simultaneously conduct classification and reconstruction. The classification provides the probability of known classes, whereas reconstruction is used to estimate the unknown score. Extensive experiments showed that the proposed method could significantly improve the classification accuracy in terms of OA, F1, and mapping error compared with the state-of-the-art methods currently used, especially when the training samples are scarce. 

    Further analysis showed that the completeness of a classification system plays an important  role  in hyperspectral image classification with unknown classes. If a dataset is lacking basic land cover elements (impervious, vegetation, and water), the proposed method will be extremely useful to reject the unknown classes and increase the classification accuracy. However, if the classification system includes all basic land cover elements, it will be difficult to identify the unknown classes. Future works should focus on this aspect to provide more accurate hyperspectral image classification. Additionally, the use of EVT is empirical. How to properly estimate the tail distribution is essential in classification with unknown classes. Future studies may focus on this aspect and utilize Gaussian distribution, skew normal distribution, and others to obtain better classification in the open world. Code and annotations used in this paper are available at  https://sjliu.me/MDL4OW.

\section*{Acknowledgements}
The authors would like to thank Prof D. Landgrebe for making the 1992 AVIRIS Indian Pines hyperspectral data available to the community and thank Prof P. Gamba for providing the ROSIS data over Pavia, Italy, along with the reference data. 

\tiny{
\ifCLASSOPTIONcaptionsoff
  \newpage	
\fi

\bibliographystyle{IEEEtran}
\bibliography{strings}

\begin{thebibliography}{10}
\providecommand{\url}[1]{#1}
\csname url@samestyle\endcsname
\providecommand{\newblock}{\relax}
\providecommand{\bibinfo}[2]{#2}
\providecommand{\BIBentrySTDinterwordspacing}{\spaceskip=0pt\relax}
\providecommand{\BIBentryALTinterwordstretchfactor}{4}
\providecommand{\BIBentryALTinterwordspacing}{\spaceskip=\fontdimen2\font plus
\BIBentryALTinterwordstretchfactor\fontdimen3\font minus
  \fontdimen4\font\relax}
\providecommand{\BIBforeignlanguage}[2]{{%
\expandafter\ifx\csname l@#1\endcsname\relax
\typeout{** WARNING: IEEEtran.bst: No hyphenation pattern has been}%
\typeout{** loaded for the language `#1'. Using the pattern for}%
\typeout{** the default language instead.}%
\else
\language=\csname l@#1\endcsname
\fi
#2}}
\providecommand{\BIBdecl}{\relax}
\BIBdecl

\bibitem{gebbers2010precision}
R.~Gebbers and V.~I. Adamchuk, ``Precision agriculture and food security,''
  \emph{Science}, vol. 327, no. 5967, pp. 828--831, 2010.

\bibitem{audebert2016semantic}
N.~Audebert, B.~Le~Saux, and S.~Lef{\`e}vre, ``Semantic segmentation of earth
  observation data using multimodal and multi-scale deep networks,'' in
  \emph{Asian Conference on Computer Vision}.\hskip 1em plus 0.5em minus
  0.4em\relax Springer, 2016, pp. 180--196.

\bibitem{kampffmeyer2016semantic}
M.~Kampffmeyer, A.-B. Salberg, and R.~Jenssen, ``Semantic segmentation of small
  objects and modeling of uncertainty in urban remote sensing images using deep
  convolutional neural networks,'' in \emph{Proceedings of the IEEE conference
  on Computer Vision and Pattern Recognition Workshops}, 2016, pp. 1--9.

\bibitem{ridd1995exploring}
M.~K. Ridd, ``Exploring a vis (vegetation-impervious surface-soil) model for
  urban ecosystem analysis through remote sensing: comparative anatomy for
  cities,'' \emph{International journal of remote sensing}, vol.~16, no.~12,
  pp. 2165--2185, 1995.

\bibitem{phinn2002monitoring}
S.~Phinn, M.~Stanford, P.~Scarth, A.~Murray, and P.~Shyy, ``Monitoring the
  composition of urban environments based on the vegetation-impervious
  surface-soil (vis) model by subpixel analysis techniques,''
  \emph{International Journal of Remote Sensing}, vol.~23, no.~20, pp.
  4131--4153, 2002.

\bibitem{yuan2005multi}
F.~Yuan, M.~E. Bauer, N.~J. Heinert, and G.~R. Holden, ``Multi-level land cover
  mapping of the twin cities (minnesota) metropolitan area with multi-seasonal
  landsat tm/etm+ data,'' \emph{Geocarto International}, vol.~20, no.~2, pp.
  5--13, 2005.

\bibitem{fauvel2012advances}
M.~Fauvel, Y.~Tarabalka, J.~A. Benediktsson, J.~Chanussot, and J.~C. Tilton,
  ``Advances in spectral-spatial classification of hyperspectral images,''
  \emph{Proceedings of the IEEE}, vol. 101, no.~3, pp. 652--675, 2012.

\bibitem{bioucas2013hyperspectral}
J.~M. Bioucas-Dias, A.~Plaza, G.~Camps-Valls, P.~Scheunders, N.~Nasrabadi, and
  J.~Chanussot, ``Hyperspectral remote sensing data analysis and future
  challenges,'' \emph{IEEE Geoscience and remote sensing magazine}, vol.~1,
  no.~2, pp. 6--36, 2013.

\bibitem{audebert2019deep}
N.~Audebert, B.~Le~Saux, and S.~Lef{\`e}vre, ``Deep learning for classification
  of hyperspectral data: A comparative review,'' \emph{IEEE Geoscience and
  Remote Sensing Magazine}, vol.~7, no.~2, pp. 159--173, 2019.

\bibitem{gualtieri1999support}
J.~A. Gualtieri and R.~F. Cromp, ``Support vector machines for hyperspectral
  remote sensing classification,'' in \emph{27th AIPR Workshop: Advances in
  Computer-Assisted Recognition}, vol. 3584.\hskip 1em plus 0.5em minus
  0.4em\relax International Society for Optics and Photonics, 1999, pp.
  221--232.

\bibitem{demir2007hyperspectral}
B.~Demir and S.~Erturk, ``Hyperspectral image classification using relevance
  vector machines,'' \emph{IEEE Geoscience and Remote Sensing Letters}, vol.~4,
  no.~4, pp. 586--590, 2007.

\bibitem{ghamisi2017advanced}
P.~Ghamisi, J.~Plaza, Y.~Chen, J.~Li, and A.~J. Plaza, ``Advanced spectral
  classifiers for hyperspectral images: A review,'' \emph{IEEE Geoscience and
  Remote Sensing Magazine}, vol.~5, no.~1, pp. 8--32, 2017.

\bibitem{li2011spectral}
J.~Li, J.~M. Bioucas-Dias, and A.~Plaza, ``Spectral--spatial hyperspectral
  image segmentation using subspace multinomial logistic regression and markov
  random fields,'' \emph{IEEE Transactions on Geoscience and Remote Sensing},
  vol.~50, no.~3, pp. 809--823, 2011.

\bibitem{benediktsson2005classification}
J.~A. Benediktsson, J.~A. Palmason, and J.~R. Sveinsson, ``Classification of
  hyperspectral data from urban areas based on extended morphological
  profiles,'' \emph{IEEE Transactions on Geoscience and Remote Sensing},
  vol.~43, no.~3, pp. 480--491, 2005.

\bibitem{kang2013spectral}
X.~Kang, S.~Li, and J.~A. Benediktsson, ``Spectral--spatial hyperspectral image
  classification with edge-preserving filtering,'' \emph{IEEE transactions on
  geoscience and remote sensing}, vol.~52, no.~5, pp. 2666--2677, 2013.

\bibitem{shen2011three}
L.~Shen and S.~Jia, ``Three-dimensional gabor wavelets for pixel-based
  hyperspectral imagery classification,'' \emph{IEEE Transactions on Geoscience
  and Remote Sensing}, vol.~49, no.~12, pp. 5039--5046, 2011.

\bibitem{rajan2008active}
S.~Rajan, J.~Ghosh, and M.~M. Crawford, ``An active learning approach to
  hyperspectral data classification,'' \emph{IEEE Transactions on Geoscience
  and Remote Sensing}, vol.~46, no.~4, pp. 1231--1242, 2008.

\bibitem{shi2015spatial}
Q.~Shi, B.~Du, and L.~Zhang, ``Spatial coherence-based batch-mode active
  learning for remote sensing image classification,'' \emph{IEEE Transactions
  on Image Processing}, vol.~24, no.~7, pp. 2037--2050, 2015.

\bibitem{shi2013semisupervised}
Q.~Shi, L.~Zhang, and B.~Du, ``Semisupervised discriminative locally enhanced
  alignment for hyperspectral image classification,'' \emph{IEEE Transactions
  on Geoscience and Remote Sensing}, vol.~51, no.~9, pp. 4800--4815, 2013.

\bibitem{wang2017novel}
Z.~Wang, B.~Du, L.~Zhang, L.~Zhang, and X.~Jia, ``A novel semisupervised
  active-learning algorithm for hyperspectral image classification,''
  \emph{IEEE Transactions on Geoscience and Remote Sensing}, vol.~55, no.~6,
  pp. 3071--3083, 2017.

\bibitem{ghamisi2013spectral}
P.~Ghamisi, J.~A. Benediktsson, and M.~O. Ulfarsson, ``Spectral--spatial
  classification of hyperspectral images based on hidden markov random
  fields,'' \emph{IEEE Transactions on Geoscience and Remote Sensing}, vol.~52,
  no.~5, pp. 2565--2574, 2013.

\bibitem{zhang2013nonlocal}
H.~Zhang, J.~Li, Y.~Huang, and L.~Zhang, ``A nonlocal weighted joint sparse
  representation classification method for hyperspectral imagery,'' \emph{IEEE
  Journal of Selected Topics in Applied Earth Observations and Remote Sensing},
  vol.~7, no.~6, pp. 2056--2065, 2013.

\bibitem{wang2018locality}
Q.~Wang, X.~He, and X.~Li, ``Locality and structure regularized low rank
  representation for hyperspectral image classification,'' \emph{IEEE
  Transactions on Geoscience and Remote Sensing}, vol.~57, no.~2, pp. 911--923,
  2018.

\bibitem{shi2015domain}
Q.~Shi, B.~Du, and L.~Zhang, ``Domain adaptation for remote sensing image
  classification: A low-rank reconstruction and instance weighting label
  propagation inspired algorithm,'' \emph{IEEE Transactions on Geoscience and
  Remote Sensing}, vol.~53, no.~10, pp. 5677--5689, 2015.

\bibitem{wang2016salient}
Q.~Wang, J.~Lin, and Y.~Yuan, ``Salient band selection for hyperspectral image
  classification via manifold ranking,'' \emph{IEEE transactions on neural
  networks and learning systems}, vol.~27, no.~6, pp. 1279--1289, 2016.

\bibitem{yu2018class}
C.~Yu, Y.~Wang, M.~Song, and C.-I. Chang, ``Class signature-constrained
  background-suppressed approach to band selection for classification of
  hyperspectral images,'' \emph{IEEE Transactions on Geoscience and Remote
  Sensing}, vol.~57, no.~1, pp. 14--31, 2018.

\bibitem{luo2020dimensionality}
F.~Luo, L.~Zhang, B.~Du, and L.~Zhang, ``Dimensionality reduction with enhanced
  hybrid-graph discriminant learning for hyperspectral image classification,''
  \emph{IEEE Transactions on Geoscience and Remote Sensing}, 2020.

\bibitem{lu2017remote}
X.~Lu, X.~Zheng, and Y.~Yuan, ``Remote sensing scene classification by
  unsupervised representation learning,'' \emph{IEEE Transactions on Geoscience
  and Remote Sensing}, vol.~55, no.~9, pp. 5148--5157, 2017.

\bibitem{lee2017going}
H.~Lee and H.~Kwon, ``Going deeper with contextual cnn for hyperspectral image
  classification,'' \emph{IEEE Transactions on Image Processing}, vol.~26,
  no.~10, pp. 4843--4855, 2017.

\bibitem{mou2017deep}
L.~Mou, P.~Ghamisi, and X.~X. Zhu, ``Deep recurrent neural networks for
  hyperspectral image classification,'' \emph{IEEE Transactions on Geoscience
  and Remote Sensing}, vol.~55, no.~7, pp. 3639--3655, 2017.

\bibitem{paoletti2018new}
M.~Paoletti, J.~Haut, J.~Plaza, and A.~Plaza, ``A new deep convolutional neural
  network for fast hyperspectral image classification,'' \emph{ISPRS journal of
  photogrammetry and remote sensing}, vol. 145, pp. 120--147, 2018.

\bibitem{hang2019cascaded}
R.~Hang, Q.~Liu, D.~Hong, and P.~Ghamisi, ``Cascaded recurrent neural networks
  for hyperspectral image classification,'' \emph{IEEE Transactions on
  Geoscience and Remote Sensing}, vol.~57, no.~8, pp. 5384--5394, 2019.

\bibitem{wan2019multiscale}
S.~Wan, C.~Gong, P.~Zhong, B.~Du, L.~Zhang, and J.~Yang, ``Multiscale dynamic
  graph convolutional network for hyperspectral image classification,''
  \emph{IEEE Transactions on Geoscience and Remote Sensing}, 2019.

\bibitem{liu2020local}
S.~Liu and Q.~Shi, ``Local climate zone mapping as remote sensing scene
  classification using deep learning: A case study of metropolitan china,''
  \emph{ISPRS Journal of Photogrammetry and Remote Sensing}, vol. 164, pp.
  229--242, 2020.

\bibitem{zhong2017spectral}
Z.~Zhong, J.~Li, Z.~Luo, and M.~Chapman, ``Spectral--spatial residual network
  for hyperspectral image classification: A 3-d deep learning framework,''
  \emph{IEEE Transactions on Geoscience and Remote Sensing}, vol.~56, no.~2,
  pp. 847--858, 2017.

\bibitem{liu2018deep}
B.~Liu, X.~Yu, A.~Yu, P.~Zhang, G.~Wan, and R.~Wang, ``Deep few-shot learning
  for hyperspectral image classification,'' \emph{IEEE Transactions on
  Geoscience and Remote Sensing}, vol.~57, no.~4, pp. 2290--2304, 2018.

\bibitem{rao2019spatial}
M.~Rao, P.~Tang, and Z.~Zhang, ``Spatial--spectral relation network for
  hyperspectral image classification with limited training samples,''
  \emph{IEEE Journal of Selected Topics in Applied Earth Observations and
  Remote Sensing}, vol.~12, no.~12, pp. 5086--5100, 2019.

\bibitem{gao2020deep}
K.~Gao, B.~Liu, X.~Yu, J.~Qin, P.~Zhang, and X.~Tan, ``Deep relation network
  for hyperspectral image few-shot classification,'' \emph{Remote Sensing},
  vol.~12, no.~6, p. 923, 2020.

\bibitem{zhou2019dcn}
Y.~Zhou, J.~Luo, L.~Feng, and X.~Zhou, ``Dcn-based spatial features for
  improving parcel-based crop classification using high-resolution optical
  images and multi-temporal sar data,'' \emph{Remote Sensing}, vol.~11, no.~13,
  p. 1619, 2019.

\bibitem{golipour2015integrating}
M.~Golipour, H.~Ghassemian, and F.~Mirzapour, ``Integrating hierarchical
  segmentation maps with mrf prior for classification of hyperspectral images
  in a bayesian framework,'' \emph{IEEE Transactions on Geoscience and remote
  Sensing}, vol.~54, no.~2, pp. 805--816, 2015.

\bibitem{khodadadzadeh2014subspace}
M.~Khodadadzadeh, J.~Li, A.~Plaza, and J.~M. Bioucas-Dias, ``A subspace-based
  multinomial logistic regression for hyperspectral image classification,''
  \emph{IEEE Geoscience and Remote Sensing Letters}, vol.~11, no.~12, pp.
  2105--2109, 2014.

\bibitem{makantasis2015deep}
K.~Makantasis, K.~Karantzalos, A.~Doulamis, and N.~Doulamis, ``Deep supervised
  learning for hyperspectral data classification through convolutional neural
  networks,'' in \emph{2015 IEEE International Geoscience and Remote Sensing
  Symposium (IGARSS)}.\hskip 1em plus 0.5em minus 0.4em\relax IEEE, 2015, pp.
  4959--4962.

\bibitem{liu2020multitask}
S.~Liu and Q.~Shi, ``Multitask deep learning with spectral knowledge for
  hyperspectral image classification,'' \emph{IEEE Geoscience and Remote
  Sensing Letters}, pp. 1--5, 2020.

\bibitem{jamshidpour2020ga}
N.~Jamshidpour, A.~Safari, and S.~Homayouni, ``A ga-based multi-view,
  multi-learner active learning framework for hyperspectral image
  classification,'' \emph{Remote Sensing}, vol.~12, no.~2, p. 297, 2020.

\bibitem{liu2019kernel}
Q.~Liu, Z.~Wu, L.~Sun, Y.~Xu, L.~Du, and Z.~Wei, ``Kernel low-rank
  representation based on local similarity for hyperspectral image
  classification,'' \emph{IEEE Journal of Selected Topics in Applied Earth
  Observations and Remote Sensing}, vol.~12, no.~6, pp. 1920--1932, 2019.

\bibitem{scheirer2012toward}
W.~J. Scheirer, A.~de~Rezende~Rocha, A.~Sapkota, and T.~E. Boult, ``Toward open
  set recognition,'' \emph{IEEE transactions on pattern analysis and machine
  intelligence}, vol.~35, no.~7, pp. 1757--1772, 2012.

\bibitem{bendale2015towards}
A.~Bendale and T.~Boult, ``Towards open world recognition,'' in
  \emph{Proceedings of the IEEE conference on computer vision and pattern
  recognition}, 2015, pp. 1893--1902.

\bibitem{tuia2011using}
D.~Tuia, E.~Pasolli, and W.~J. Emery, ``Using active learning to adapt remote
  sensing image classifiers,'' \emph{Remote Sensing of Environment}, vol. 115,
  no.~9, pp. 2232--2242, 2011.

\bibitem{bendale2016towards}
A.~Bendale and T.~E. Boult, ``Towards open set deep networks,'' in
  \emph{Proceedings of the IEEE conference on Computer Vision and Pattern
  Recognition}, 2016, pp. 1563--1572.

\bibitem{neal2018open}
L.~Neal, M.~Olson, X.~Fern, W.-K. Wong, and F.~Li, ``Open set learning with
  counterfactual images,'' in \emph{Proceedings of the European Conference on
  Computer Vision (ECCV)}, 2018, pp. 613--628.

\bibitem{dhamija2018reducing}
A.~R. Dhamija, M.~G{\"u}nther, and T.~Boult, ``Reducing network
  agnostophobia,'' in \emph{Advances in Neural Information Processing Systems},
  2018, pp. 9157--9168.

\bibitem{yoshihashi2019classification}
R.~Yoshihashi, W.~Shao, R.~Kawakami, S.~You, M.~Iida, and T.~Naemura,
  ``Classification-reconstruction learning for open-set recognition,'' in
  \emph{Proceedings of the IEEE Conference on Computer Vision and Pattern
  Recognition}, 2019, pp. 4016--4025.

\bibitem{krizhevsky2009cifar}
A.~Krizhevsky, V.~Nair, and G.~Hinton, ``Cifar-10 and cifar-100 datasets,''
  \emph{URl: https://www. cs. toronto. edu/kriz/cifar. html}, vol.~6, 2009.

\bibitem{xiao2017fashion}
H.~Xiao, K.~Rasul, and R.~Vollgraf, ``Fashion-mnist: a novel image dataset for
  benchmarking machine learning algorithms,'' \emph{arXiv preprint
  arXiv:1708.07747}, 2017.

\bibitem{pan2018mugnet}
B.~Pan, Z.~Shi, and X.~Xu, ``Mugnet: Deep learning for hyperspectral image
  classification using limited samples,'' \emph{ISPRS Journal of Photogrammetry
  and Remote Sensing}, vol. 145, pp. 108--119, 2018.

\bibitem{xu2019abundance}
S.~Xu, J.~Li, M.~Khodadadzadeh, A.~Marinoni, P.~Gamba, and B.~Li,
  ``Abundance-indicated subspace for hyperspectral classification with limited
  training samples,'' \emph{IEEE Journal of Selected Topics in Applied Earth
  Observations and Remote Sensing}, vol.~12, no.~4, pp. 1265--1278, 2019.

\bibitem{deng2009imagenet}
J.~Deng, W.~Dong, R.~Socher, L.-J. Li, K.~Li, and L.~Fei-Fei, ``Imagenet: A
  large-scale hierarchical image database,'' in \emph{2009 IEEE conference on
  computer vision and pattern recognition}.\hskip 1em plus 0.5em minus
  0.4em\relax Ieee, 2009, pp. 248--255.

\bibitem{geng2020collective}
C.~Geng and S.~Chen, ``Collective decision for open set recognition,''
  \emph{IEEE Transactions on Knowledge and Data Engineering}, 2020.

\bibitem{he2016deep}
K.~He, X.~Zhang, S.~Ren, and J.~Sun, ``Deep residual learning for image
  recognition,'' in \emph{Proceedings of the IEEE conference on Computer Vision
  and Pattern Recognition}, 2016, pp. 770--778.

\bibitem{lin2013network}
M.~Lin, Q.~Chen, and S.~Yan, ``Network in network,'' \emph{arXiv preprint
  arXiv:1312.4400}, 2013.

\bibitem{pan2016shallow}
J.~Pan, E.~Sayrol, X.~Giro-i Nieto, K.~McGuinness, and N.~E. O'Connor,
  ``Shallow and deep convolutional networks for saliency prediction,'' in
  \emph{Proceedings of the IEEE Conference on Computer Vision and Pattern
  Recognition}, 2016, pp. 598--606.

\bibitem{zhang2016deep}
L.~Zhang, L.~Zhang, and B.~Du, ``Deep learning for remote sensing data: A
  technical tutorial on the state of the art,'' \emph{IEEE Geoscience and
  Remote Sensing Magazine}, vol.~4, no.~2, pp. 22--40, 2016.

\bibitem{ioffe2015batch}
S.~Ioffe and C.~Szegedy, ``Batch normalization: Accelerating deep network
  training by reducing internal covariate shift,'' \emph{arXiv preprint
  arXiv:1502.03167}, 2015.

\bibitem{nair2010rectified}
V.~Nair and G.~E. Hinton, ``Rectified linear units improve restricted boltzmann
  machines,'' in \emph{Proceedings of the 27th international conference on
  machine learning (ICML-10)}, 2010, pp. 807--814.

\bibitem{pickands1975statistical}
J.~Pickands~III \emph{et~al.}, ``Statistical inference using extreme order
  statistics,'' \emph{the Annals of Statistics}, vol.~3, no.~1, pp. 119--131,
  1975.

\bibitem{grimshaw1993computing}
S.~D. Grimshaw, ``Computing maximum likelihood estimates for the generalized
  pareto distribution,'' \emph{Technometrics}, vol.~35, no.~2, pp. 185--191,
  1993.

\bibitem{oza2019c2ae}
P.~Oza and V.~M. Patel, ``C2ae: Class conditioned auto-encoder for open-set
  recognition,'' in \emph{Proceedings of the IEEE Conference on Computer Vision
  and Pattern Recognition}, 2019, pp. 2307--2316.

\bibitem{geng2020recent}
C.~Geng, S.-j. Huang, and S.~Chen, ``Recent advances in open set recognition: A
  survey,'' \emph{IEEE Transactions on Pattern Analysis and Machine
  Intelligence}, 2020.

\bibitem{liu2019large}
Z.~Liu, Z.~Miao, X.~Zhan, J.~Wang, B.~Gong, and S.~X. Yu, ``Large-scale
  long-tailed recognition in an open world,'' in \emph{Proceedings of the IEEE
  Conference on Computer Vision and Pattern Recognition}, 2019, pp. 2537--2546.

\bibitem{abadi2016tensorflow}
M.~Abadi, P.~Barham, J.~Chen, Z.~Chen, A.~Davis, J.~Dean, M.~Devin,
  S.~Ghemawat, G.~Irving, M.~Isard \emph{et~al.}, ``Tensorflow: A system for
  large-scale machine learning,'' in \emph{12th $\{$USENIX$\}$ Symposium on
  Operating Systems Design and Implementation ($\{$OSDI$\}$ 16)}, 2016, pp.
  265--283.

\bibitem{chollet2015keras}
F.~Chollet \emph{et~al.}, ``Keras,'' \url{https://github.com/fchollet/keras},
  2015.

\bibitem{zeiler2012adadelta}
M.~D. Zeiler, ``Adadelta: an adaptive learning rate method,'' \emph{arXiv
  preprint arXiv:1212.5701}, 2012.

\bibitem{liu2018wide}
S.~Liu, H.~Luo, Y.~Tu, Z.~He, and J.~Li, ``Wide contextual residual network
  with active learning for remote sensing image classification,'' in
  \emph{IGARSS 2018-2018 IEEE International Geoscience and Remote Sensing
  Symposium}.\hskip 1em plus 0.5em minus 0.4em\relax IEEE, 2018, pp.
  7145--7148.

\bibitem{hu2015deep}
W.~Hu, Y.~Huang, L.~Wei, F.~Zhang, and H.~Li, ``Deep convolutional neural
  networks for hyperspectral image classification,'' \emph{Journal of Sensors},
  vol. 2015, 2015.

\bibitem{li2016hyperspectral}
W.~Li, G.~Wu, F.~Zhang, and Q.~Du, ``Hyperspectral image classification using
  deep pixel-pair features,'' \emph{IEEE Transactions on Geoscience and Remote
  Sensing}, vol.~55, no.~2, pp. 844--853, 2016.

\bibitem{mei2017learning}
S.~Mei, J.~Ji, J.~Hou, X.~Li, and Q.~Du, ``Learning sensor-specific
  spatial-spectral features of hyperspectral images via convolutional neural
  networks,'' \emph{IEEE Transactions on Geoscience and Remote Sensing},
  vol.~55, no.~8, pp. 4520--4533, 2017.

\bibitem{lu2017hybrid}
X.~Lu, W.~Zhang, and X.~Li, ``A hybrid sparsity and distance-based
  discrimination detector for hyperspectral images,'' \emph{IEEE Transactions
  on Geoscience and Remote Sensing}, vol.~56, no.~3, pp. 1704--1717, 2017.

\end{thebibliography}
}

\end{document}